\documentclass[10pt,journal,compsoc]{IEEEtran}

\usepackage{comment}
\usepackage{balance}
\usepackage{booktabs}
\usepackage{tabularx}
\usepackage{footnote}

\usepackage{amsmath,amssymb,amsthm,amsfonts}
\usepackage{bbm,bbold,mathabx,braket,theoremref}
\usepackage{multirow}
\usepackage{pdfpages}

\newcommand{\blue}[1]{{#1}}

\usepackage{xspace}

\ifCLASSOPTIONcompsoc
	\usepackage[nocompress]{cite}
	\usepackage[caption=false,font=footnotesize,labelfont=sf,textfont=sf]{subfig}
\else
	\usepackage{cite}
	\usepackage[caption=false,font=footnotesize]{subfig}
\fi
\usepackage{graphicx}
\usepackage[hidelinks]{hyperref}

\usepackage{algpseudocode,algorithm,algorithmicx}
\usepackage{textcomp}
\usepackage{xcolor}
\def\BibTeX{{\rm B\kern-.05em{\sc i\kern-.025em b}\kern-.08em
		T\kern-.1667em\lower.7ex\hbox{E}\kern-.125emX}}

\usepackage{verbatim}
\usepackage{nicefrac}

\newtheorem{lemma}{Lemma}

\DeclareMathOperator{\mymiddle}{mid}
\DeclareMathOperator{\mean}{mean}
\DeclareMathOperator{\std}{std}
\DeclareMathOperator*{\argmax}{argmax}
\DeclareMathOperator*{\argmin}{argmin}

\newcommand{\textbbf}[1]{{\fontseries{bx}\selectfont {#1}}}
\newcommand{\rom}[1]{\uppercase\expandafter{\romannumeral #1\relax}}
\newcommand{\bs}[1]{\boldsymbol{#1}}

\newcommand{\app}{Supplementary Material\xspace}
\newcommand{\inapp}{in the Supplementary Material\xspace}
\newcommand{\figconv}{1 \inapp}
\newcommand{\tabresult}{4 \inapp}
\newcommand{\figcomplviews}{2}
\newcommand{\figcomplpoints}{3}
\newcommand{\figcompledges}{4 \inapp}
\newcommand{\tabtime}{1 \inapp}

\newcommand{\HRule}[1][\medskipamount]{\par
	\vspace*{\dimexpr-\parskip-\baselineskip+#1}
	\noindent\rule{\linewidth}{0.2mm}\par
	\vspace*{\dimexpr-\parskip-.5\baselineskip+#1}}

\usepackage{tikz}
\usetikzlibrary{shapes.multipart, shapes.arrows}
\tikzset{MyArrow/.style={single arrow, draw, inner sep=.1cm, minimum height=.4cm, single arrow head extend=0.15cm, anchor=south, rotate=-90}}
\tikzset{MyArrow2/.style={single arrow, draw, inner sep=.15cm, minimum height=.35cm, single arrow head extend=0.18cm, anchor=west}}
\usetikzlibrary{shapes.misc}
\tikzset{MyArrow3/.style={single arrow, draw, inner sep=0.1cm, minimum height=0.1cm, single arrow head extend=0.4cm, single arrow tip angle=160, anchor=north, shape border rotate=-180}}
\usetikzlibrary{shapes.misc}%
\tikzset{MyArrow4/.style={single arrow, draw, inner sep=0.1cm, minimum height=0.1cm, single arrow head extend=0.4cm, single arrow tip angle=165, anchor=north, shape border rotate=-180}}
\usetikzlibrary{shapes.misc}%
\tikzset{MyArrow5/.style={single arrow, draw, inner sep=0.1cm, minimum height=0.2cm, single arrow head extend=0.4cm, single arrow tip angle=160, anchor=north, shape border rotate=-180}}
\usetikzlibrary{shapes.misc}%
\tikzset{MyPlus/.style={cross out, draw, line width=.17cm, minimum size=.25cm, anchor=west, rotate=45, black!40}}

\title{Multi-view Graph Learning by Joint Modeling of Consistency and Inconsistency}
\author{Youwei~Liang,
	Dong~Huang,~\IEEEmembership{Member,~IEEE, }
	Chang-Dong~Wang,~\IEEEmembership{Member,~IEEE, }\protect\\
	and~Philip S. Yu,~\IEEEmembership{Fellow,~IEEE}%
	\IEEEcompsocitemizethanks{
		\IEEEcompsocthanksitem Y. Liang is with the College of Mathematics and Informatics, South China Agricultural University, Guangzhou, China, and also with Tencent, Shenzhen, China. E-mail: liangyouwei1@gmail.com.
		\IEEEcompsocthanksitem D. Huang is with the College of Mathematics and Informatics, South China Agricultural University, Guangzhou, China, and also with Pazhou Lab, Guangzhou, China. E-mail: huangdonghere@gmail.com.
		\IEEEcompsocthanksitem C.-D. Wang is with the School of Data and Computer Science,
		Sun Yat-sen University, Guangzhou, China, and also with Guangdong Key Laboratory of Information Security Technology, Guangzhou, China, and also with Key Laboratory of Machine Intelligence and Advanced Computing, Ministry of Education, China. E-mail: changdongwang@hotmail.com.
		\IEEEcompsocthanksitem P. S. Yu is with the Department of Computer Science, University of Illinois at Chicago, Chicago, IL 60607, USA. E-mail: psyu@cs.uic.edu}%
	\thanks{\normalfont{\blue{The codebase (including all comparison algorithms and all tested datasets) of this paper is available at this GitHub repository:\protect\\ \url{https://github.com/youweiliang/Multi-view_Graph_Learning}.}}}
}

\begin{document}

\IEEEtitleabstractindextext{%
\begin{abstract}
Graph learning has emerged as a promising technique for multi-view clustering due to its ability to learn a unified and robust graph from multiple views. However, existing graph learning methods mostly focus on the multi-view consistency issue, yet often neglect the inconsistency between views, which makes them vulnerable to possibly low-quality or noisy datasets. To overcome this limitation, we propose a new multi-view graph learning framework, which for the first time simultaneously and explicitly models multi-view \emph{consistency} and \emph{inconsistency} in a unified objective function, through which the consistent and inconsistent parts of each single-view graph as well as the unified graph that fuses the consistent parts can be iteratively learned. Though optimizing the objective function is NP-hard, we design a highly efficient optimization algorithm that can obtain an approximate solution with linear time complexity in the number of edges in the unified graph. Furthermore, our multi-view graph learning approach can be applied to both similarity graphs and dissimilarity graphs, which lead to two graph fusion-based variants in our framework. Experiments on twelve multi-view datasets have demonstrated the robustness and efficiency of the proposed approach.
\end{abstract}

\begin{IEEEkeywords}
	Clustering; Multi-view graph learning; Multi-view clustering; Graph fusion; Consistency; Inconsistency.
\end{IEEEkeywords}}

\maketitle
\IEEEdisplaynontitleabstractindextext
\IEEEpeerreviewmaketitle

\IEEEraisesectionheading{\section{Introduction}\label{sec:introduction}}
\IEEEPARstart{M}{ulti-view} data consist of features collected from multiple heterogeneous sources (or views). Multiple views of features can provide rich and complementary information for discovering the underlying cluster structure of data. It has been a popular research topic in recent years as to how to exploit the features effectively and jointly from multiple views and thus achieve robust clustering results for multi-view data. 

In the literature, numerous (single-view) clustering methods have been developed \cite{jain10_survey}, among which the graph-based methods are one of the most widely-studied categories \cite{shi2000normalized, ng2002spectral,von2007tutorial}. The graph-based methods typically construct a graph structure, and then partition the graph to obtain {the} clustering result. In these methods, the construction of the graph is independent of clustering, and the clustering performance heavily relies on the predefined graph. To alleviate this limitation, some graph learning methods have been presented \cite{Nie2014_kdd,nie2016constrained}, where the graph structure can be adaptively learned in {the} clustering process. Recently, inspired by single-view graph learning \cite{Nie2014_kdd,nie2016constrained}, multi-view graph learning has rapidly emerged as a powerful technique for enhancing multi-view clustering performance \cite{zhan2017graph,ZhanFusion,Zhan2018,nie2017self}. Notably, Zhan et al. \cite{zhan2017graph,ZhanFusion,Zhan2018} developed several multi-view graph learning approaches which are able to fuse multiple graphs into a unified graph with a desired number of connected components. Nie et al. \cite{nie2017self} proposed a self-weighted scheme for fusing multiple graphs with the importance of each view considered. Despite these significant progresses, a common limitation to these multi-view graph learning methods lies in that they mostly focus on the consistency of multiple views, but lack the ability to explicitly consider both multi-view consistency and inconsistency (which may be brought in by noise, corruptions, or view-specific characteristics) in their frameworks, which may degrade their performances when faced with complex or possibly noisy data. 

To deal with the potential noise or corruptions in a graph, Bojchevski et al. \cite{kdd2017robust} proposed a new graph-based clustering method based on the latent decomposition of the similarity graph into two graphs, namely, the \emph{good} graph and the \emph{corrupted} graph. Though it is able to learn a good graph by eliminating the influence of potential noise, this graph learning method \cite{kdd2017robust} is only applicable to a single graph (for a single view) and cannot be utilized in the multi-view graph learning task where multiple graphs from multiple views are involved. Thereby, how to jointly model the multi-view consistency (which can be viewed as the multi-view good graphs) and the multi-view inconsistency (which can be viewed as the multi-view corrupted graphs) in a unified graph learning model to improve multi-view clustering performance is still {an open} problem.

\begin{figure*}[!t]\vskip 0.2 in
	\centering
	\input{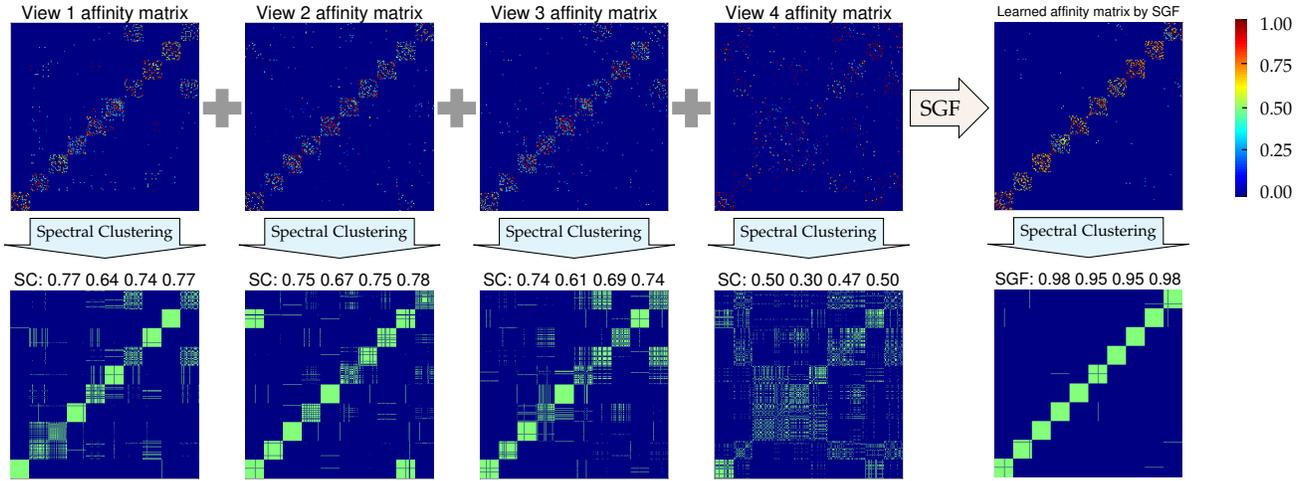}
	\caption{(Best viewed/printed in color.) Visualization of the affinity (similarity) matrices of the UCI Handwritten Digit dataset. Four views are used to learn the unified graph and the number of $k$NN is 15. The first row corresponds to the four single-view affinity matrices and the learned affinity matrix (i.e., the unified graph) by our multi-view graph learning framework SGF. The second row corresponds to the clustering results by performing spectral clustering (SC) on the single-view graph and the learned graph. The four numbers on each sub-figure are the clustering scores ACC, ARI, NMI, and purity, respectively.}
	\label{fig:visual}\vskip 0.2 in
\end{figure*}

To tackle this problem, this paper proposes a novel multi-view graph learning approach, which is further applied {to} multi-view clustering. In this paper, we argue that the simultaneous modeling of {multi-view consistency and inconsistency} can significantly benefit the multi-view graph learning process. In particular, with the graph structures of multiple views given, their consistency and inconsistency are simultaneously leveraged to learn a unified graph. It is intuitive to assume that the graph of each view can be decomposed into two parts, i.e., the consistent part and the inconsistent part, and the goal is to learn and remove the inconsistent (or noisy) parts while preserving the consistent parts. Specifically, we formulate the multi-view consistency and multi-view inconsistency as well as the graph fusion term into a new objective function. By iteratively optimizing the objective function, the multi-view graph decomposition and the multi-view graph fusion are simultaneously achieved. With the fused graph obtained, some conventional graph-based methods like spectral clustering can be performed to obtain the final multi-view clustering result. For clarity, we provide an illustration for our multi-view graph learning model in Figure~\ref{fig:visual}. As shown in the first row of Figure~\ref{fig:visual}, the four similarity (affinity) matrices from four views appear to be corrupted to different extents, and our similarity graph fusion (SGF) method can effectively remove most of these corruptions (or inconsistency) while yielding a unified and better graph with their consistent parts fused and {strengthened}. As shown in the second row in Figure~\ref{fig:visual}, by graph fusion with both consistency and inconsistency considered, the final clustering (in the fifth column) on the fused graph {is significantly} better than the clustering on the single-view graphs. It also shows that the proposed approach is able to achieve superior performance at the presence of some highly corrupted graph (e.g., the fourth single-view graph in Figure~\ref{fig:visual}). 

For clarity, the main contributions of this work are summarized as follows.

\begin{itemize}
	\item We propose a novel multi-view graph learning approach, which for the first time, to the best of our knowledge, simultaneously and explicitly models multi-view consistency and multi-view inconsistency in a unified objective function, where multi-view consistency can be iteratively learned and fused into a unified graph as the multi-view inconsistency is automatically identified and removed. 
	
	\item \blue{Though optimizing the objective function is NP-hard, we design a highly efficient algorithm to obtain an approximate solution by exploiting the structures of the quadratic programs based on eigenvalue analysis and constraint simplification. Our algorithm has roughly linear time complexity in the number of effective edges in an affinity graph or the number of nodes in a $k$NN graph.}
	
	\item A novel multi-view clustering framework based on multi-view graph learning is presented, which is further extended into two graph fusion variants, corresponding to distance (dissimilarity) graph fusion (DGF) and similarity graph fusion (SGF), respectively. {The proposed DGF and SGF have shown superior performance over the state-of-the-art multi-view clustering algorithms in extensive experiments. }
	
	\item %
	\blue{%
	Even without dataset-specific hyper-parameter tuning, the proposed algorithms can still achieve highly competitive clustering results on various multi-view datasets when comparing to the best results obtained by the some other state-of-the-art multi-view clustering approaches.} %
\end{itemize}

A preliminary version of this paper was reported in\cite{liang2019consistency}. In this paper, we have made significant {revisions} and {added} substantial analysis for the proposed framework. Firstly, we have almost completely revised the optimization algorithm, which improves the efficiency and stability of the proposed framework. \blue{In particular, we have gained deeper understandings of the optimization problem (e.g., in Lemma~\ref{lemma:nonneg}, Lemma~\ref{lemma:pos eig}, and Sections~\ref{sec:optim}, \ref{sec:dense}, \ref{sec:sub1}, \ref{subproblem 3}), and developed a new optimization algorithm for it, which is more efficient and stable than the projection method in \cite{liang2019consistency} (as demonstrated by extensive comparison experiments)}. Specifically, our theoretical analysis reveals that one of the optimization subproblems can be rewritten as the sum of many quadratic functions that share a same Hessian matrix, which allows us to modify the d.c. algorithm \cite{tao1998branch} to optimize these quadratic functions all at once. Secondly, as a consequence of improved stability, {the proposed framework with the new optimization method can learn a good graph even without dataset-specific parameter tuning}, which {will be} reported in the \app. Thirdly, we theoretically show that the time complexity of the new optimization approach is basically linear in the number of edges in affinity graph or the number of nodes in a $k$NN graph, even though exactly solving the problem is NP-hard. Fourthly, we introduce multi-view dense representation to replace the sparse matrix used in \cite{liang2019consistency} that consumes three times larger memory and is less efficient. And a graph normalization method is added to the two graph fusion algorithms (SGF and DGF), which improves the performance of clustering. Fifthly, we introduce view-specific weights for every view in our framework so that users can ensure the unified graph is closer to the important views specific to their applications. Last but not least, the experimental section is substantially extended, where more benchmark datasets are used and more experimental comparison and analysis are provided, which further demonstrate the efficiency, effectiveness, and robustness of the proposed framework. 

The rest of this paper is organized as follows. In Section~\ref{sec:related}, we review the related work in multi-view clustering, especially in multi-view graph learning. In Section~\ref{sec:motivation}, we propose the novel multi-view graph learning framework. {In Section~\ref{sec:optimization}, we} theoretically analyze the optimization problem in our framework and present a highly efficient algorithm to solve it. {In Section~\ref{spectral clustering}, two} {specific} graph fusion versions for multi-view spectral clustering are presented {based on the proposed framework}. {Finally, we} report the experimental results in Section~\ref{sec:experiment} and conclude this paper in Section~\ref{conclusion}. More experimental results are reported in the \app.

\section{Related work} \label{sec:related}
In recent years, multi-view clustering has been a popular topic and many multi-view clustering algorithms have been developed from different technical perspectives.

Bickel and Scheffer \cite{bickel2004multi} extended the semi-supervised co-training approaches \cite{blum1998combining} to multi-view clustering. The basic idea of co-training is to iterate over all views and optimize an objective function in next view using result obtained from last view.
However, its limitation is that co-training based multi-view clustering algorithms may not converge \cite{bickel2004multi} and thus it is difficult to decide when to stop.

Kumar et al. \cite{kumar2011co} proposed Co-regularized Spectral Clustering based on maximizing clustering agreement among all views. They presented an alternative regularization scheme that regularizes each view-specific set of eigenvectors towards a common centroid and used the common centroid to obtain the clustering result. The basic idea of their algorithm is that all views should yield a consensus clustering result. 
The idea of maximizing clustering agreement of all views is exploited by many other multi-view clustering approaches\cite{wang2016iterative, zong2018weighted, wang2018multiview}.
For example, Zong et al. \cite{zong2018weighted} introduced Weighted Multi-View Spectral Clustering. They found that the similarity between the clustering results of different views can be measured by the largest canonical angle between the subspaces spanned by the eigenvectors of the normalized Laplacian matrices for different views. Therefore, minimizing the canonical angles leads to maximizing the clustering agreement of all views.
To eliminate the potential noise in data, Xia et al. \cite{xia2014robust} proposed Robust Multi-View Spectral Clustering, which aims to learn an intrinsic transition matrix from multiple views by restricting the transition matrix to be low-rank.

Nie et al. \cite{nie2018multiview} use Procrustes Analysis technique to obtain a consensus cluster indicator matrix from the spectral embedding of multi-view kernels. 
One possible limitation to this approach is that some information of multi-view graphs is lost before obtaining the clustering result. 
To elaborate, we know that the eigenvectors corresponding to the largest eigenvalues of Laplacian matrix contain information to partition the graph\cite{shi2000normalized}. The AWP method \cite{nie2018multiview} use only information from the eigenvectors corresponding to some largest eigenvalues, and the information of the eigenvectors corresponding to other large eigenvalues is lost. To utilize the information from the entire spectrum of the Laplacian matrices of all views, it seems more reasonable to perform multi-view learning before obtaining spectral embedding. Our framework is likely to make the most of the spectral information in all views since the unified graph is learned before computing the eigenvectors. 

Huang et al. \cite{huang2012affinity} proposed a method for aggregating affinity matrices for spectral clustering, which attempts to reduce the influence of unreliable and irrelevant features in data. 
Nie et al. \cite{nie2017self} proposed a parameter-free self-weighted scheme to fuse multiple graphs with the importance of each view considered. 
Zhan et al. \cite{zhan2017graph,Zhan2018,ZhanFusion} proposed to learn an intrinsic similarity graph from multiple similarity graphs. Their approaches learn the consensus graph by tuning the fused graph structure until it contains exactly the desired number of connected components $n_c$. During the iterative learning of graph structure, the $n_c$ smallest eigenvalues of the graph Laplacian need to be computed. If there are exactly $n_c$ smallest eigenvalues being $0$, then the number of connected components is exactly $n_c$ \cite{von2007tutorial} and the learning is finished. However, with $n_c$ eigenvalues being the same, which is termed eigenvalues cluster in the literature, the eigen-decomposition algorithm {may have difficulty in converging} \cite{li2015convergence}. {As pointed out in \cite{li2015convergence}, the} closer the eigenvalues lie in the cluster, the slower the algorithm convergences \cite{li2015convergence}. Indeed, we have found in our experiments that {the convergence of the MCGC \cite{Zhan2018} and MVGL \cite{zhan2017graph} algorithms could be quite slow, which results in a heavy computational burden for larger datasets. Moreover, they mostly focus on the consistent properties across multiple views, but often lack the ability to simultaneously and explicitly model the multi-view {consistency and} inconsistency information, which may degrade their graph learning and clustering performances when faced with complex multi-view datasets. }

\section{Learning a Consistent Graph with Inconsistency Considered} \label{sec:motivation}
In this section, we propose a new multi-view graph learning method which is capable of joint modeling of multi-view consistency and multi-view inconsistency in a unified optimization framework. The idea is to decompose the adjacency matrix of each graph (for each view) into two parts, the consistent part and the inconsistent part. By the definition of inconsistency, we design a novel objective function which can automatically identify the inconsistent parts and fuse the consistent parts into a unified adjacency matrix. By iteratively optimizing the objective function, the inconsistent and consistent parts of each view as well as the unified adjacency matrix are iteratively learned. 

\blue{Let $\mathbf W^{(i)} \in \mathbb{R}_{\geq 0}^{n\times n}$ denote the similarity matrix for the $i$-th view, with $n$ being the number of instances (data points). 
We assume that the similarity matrix is scale invariant in the applications, i.e., a similarity matrix $\mathbf W^{(i)}$ is equivalent to $k \mathbf W^{(i)}$ for all $k > 0$. This is a reasonable assumption because in practice we typically only care the relative similarity of two nodes instead of their absolute similarity. For instance, we typically care whether the similarity between nodes $x$ and $y$ is higher than that of nodes $y$ and $z$. }

\blue{Specifically, this assumption holds well for spectral clustering since scaling does not affect the eigenvectors of a matrix. To elaborate, suppose we have two similarity matrices $\mathbf W^{(i)}$ and $\mathbf W^{(j)}$ with $\mathbf W^{(i)} = k \mathbf W^{(j)}$.} Then their symmetrically normalized Laplacian matrices $\mathbf L^{(i)} = \mathbf L^{(j)}$, which give exactly the same clustering results in normalized cut\cite{ng2002spectral}.
Under the scale invariant assumption, we need to scale the similarity matrices before fusing them into a unified similarity matrix, i.e., multiplying $\mathbf W^{(i)}$ with a learnable scaling coefficient $\alpha_i $. To make the scaling result unique, we restrict the sum of the coefficients to $1$, i.e., $\boldsymbol{\alpha}^{\top} \mathbf {1} = 1$. %
All the scaled similarity matrices should be close to the unified similarity matrix $\mathbf {S}$. Hence we have the following constrained optimization problem:
\begin{align} \label{func:obj1}
\min _{\boldsymbol \alpha,\mathbf {S}} \quad &\sum_{i=1}^{v} \big\|\alpha_i \mathbf W^{(i)}-\mathbf {S} \big\|_F^2 \\
\text { s.t. } \quad &\boldsymbol \alpha^{\top} \mathbf {1} = 1, \alpha \geq 0, \mathbf {S} \geq 0. \notag
\end{align}
Here, $v$ is the number of views.

To jointly model multi-view consistency and multi-view inconsistency, we decompose the similarity matrix $\mathbf W^{(i)}$ for the $i$-th view into two parts: the consistent part $\mathbf A^{(i)}$ and the inconsistent part $\mathbf E^{(i)}$: 
\begin{equation}
\mathbf W^{(i)} = \mathbf A^{(i)} + \mathbf E ^{(i)}
\end{equation}
with $\mathbf A^{(i)}, \mathbf E^{(i)} \in \mathbb{R}_{\geq 0}^{n\times n}$.
The core question is how to find matrices $\mathbf A^{(i)}$ and $\mathbf E ^{(i)}$ for $i=1,\dots,v$. 

Different from previous decomposition works \cite{xia2014robust,kdd2017robust} that mainly focus on modeling the noise in data, in this paper, the inconsistency in multi-view data is a much broader concept, which involves not only noise, but also the difference in view-specific characteristics. \blue{While noise is typically considered sparse on a similarity graph\cite{xia2014robust, kdd2017robust}, the inconsistency may not. Since the relationship of nodes on a graph may be intrinsically different across views,  inconsistency can appear everywhere on the similarity graphs. Thus, the sparsity of noise \emph{within a single similarity matrix} is not suitable for identifying the inconsistency on multi-view similarity graphs. Instead, we assume that the inconsistency is sparse \emph{across views}.} For example, suppose we have five views and the similarities between nodes $x$ and $y$ on each view are $3.16, 3.19, 3.22, 3.17,$ and $3.95$, respectively (assuming we have properly scaled the similarity matrices). We tend to believe that a good similarity measure between $x$ and $y$ is $3.20$ (i.e., the consistent part). The similarity on the $5$-th view has a deviation of $0.75$ (i.e., the inconsistent part) from the consistent part. We say the inconsistency is sparse across views because only the $5$-th view has a relatively large inconsistent part. To ensure the inconsistency is sparse across views, it is natural to decrease the sum of the products of the inconsistent parts, i.e.,
\begin{equation} \label{eq:inconsistency penalty}
\sum_{\substack{i,j=1 \\ i \neq j}}^v \operatorname{sum}\left( \big(\alpha_i \mathbf E^{(i)}\big) \circ \big(\alpha_i \mathbf E^{(j)}\big)\right),
\end{equation}
where $\circ$ denotes the Hadamard product (element-wise multiplication) of two matrices and $\mathrm{sum}(\cdot)$ is the operator of summing all elements in a matrix. We scale the inconsistent part of each view to make them have fair contributions to the sum. 
Furthermore, we generally do not want the inconsistent parts to be too large, which leads to preventing the following value from becoming too large during learning:
\begin{equation} \label{eq:inconsistency magnitude}
\sum_{i=1}^v \operatorname{sum}\left(\big(\alpha_i \mathbf E^{(i)}\big) \circ \big(\alpha_i \mathbf E^{(i)}\big)\right).
\end{equation}

To jointly model multi-view consistency and inconsistency in a unified optimization framework, we combine the three terms \eqref{func:obj1}, \eqref{eq:inconsistency penalty} and \eqref{eq:inconsistency magnitude} into a unified objective function:
\begin{align} \label{func:obj2}
\min_{\substack{\boldsymbol \alpha,\mathbf A^{(1)},\dots,\mathbf A^{(v)},\\\mathbf E^{(1)},\dots,\mathbf E^{(v)},\mathbf {S}}} \quad &\sum_{i=1}^{v} \big\|\alpha_i \mathbf A^{(i)}-\mathbf {S} \big\|_F^2 \\&
+ \beta \sum_{i=1}^v \operatorname{sum}\left(\big(\alpha_i \mathbf E^{(i)}\big) \circ \big(\alpha_i \mathbf E^{(i)}\big)\right) \label{2nd term}\\[5pt] &+
\gamma \sum_{\substack{i,j=1 \\ i \neq j}}^v \operatorname{sum}\left(\big(\alpha_i \mathbf E^{(i)}\big) \circ \big(\alpha_j \mathbf E^{(j)}\big)\right) \label{3rd term}\\[10pt]
\text { s.t. } \quad &\boldsymbol \alpha^\top \mathbf {1}=1, \boldsymbol \alpha \geq 0, \mathbf {S} \geq 0, \notag\\
&\mathbf W^{(i)}=\mathbf A^{(i)}+\mathbf E^{(i)}, \\
&\mathbf A^{(i)} \geq 0, \mathbf E^{(i)} \geq 0, \quad i=1, \dots, {v}, \notag
\end{align}
where $\beta, \gamma$ are parameters, and $\beta$ controls the magnitude of inconsistent parts and $\gamma$ prevents the consistent parts being incorrectly moved to inconsistent parts (i.e., ensuring sparsity of multi-view inconsistency) as we explain next. %

\blue{To see how this objective can remove multi-view inconsistency, let us initialize $\mathbf A^{(i)}$ as $\mathbf W^{(i)}$, and $\mathbf E^{(i)}$ as $\mathbf 0$. During an iterative learning process, if the inconsistent parts are correctly moved to $\mathbf E^{(i)}$ from $\mathbf A^{(i)}$, $\alpha_{i} \mathbf A^{(i)}$ will gets closer to the consistent component of all views (i.e., $\mathbf S$) and thus the first term in our objective will decrease. Although the second and third terms will increase in this case, their increases will be small because of the sparsity of cross-view inconsistency and the small value of $\beta$, thus canceled out by the considerable amount of reduction of the first term in the overall objective, and the net result is the reduction of the overall objective.} Therefore, \emph{the optimization process is actually moving the inconsistent parts from the original similarity matrix $\mathbf W^{(i)}$ to the matrix $\mathbf E^{(i)}$ by minimizing the overall objective}, which is the core principle of how we simultaneously model multi-view consistency and multi-view inconsistency in a unified optimization framework.

We shall rewrite the objective function in order to better apply optimization techniques to solve it.
Let $\mathbf B$ be a ${v}$-by-${v}$ matrix with its diagonal elements being $\beta$ and off-diagonal elements being $\gamma$.
Then our objective function can be written in a more compact form
\begin{align} \label{func:obj3}
\min _{\substack{\boldsymbol \alpha,\mathbf {S},\\ \mathbf A^{(1)},\dots,\mathbf A^{(v)}}} \quad &\sum_{i=1}^{v} \lambda_{i} \big\|\alpha_i \mathbf A^{(i)}-\mathbf {S}\big\|_F^2 + \\
\sum_{i,j=1}^{v} B_{ij} \lambda_{i} \lambda_{j} &\alpha_i \alpha_j \operatorname{sum}\left(\big(\mathbf W^{(i)}-\mathbf A^{(i)}\big) \circ \big(\mathbf W^{(j)}-\mathbf A^{(j)}\big)\right)\notag\\
\text{s.t.} \quad &\boldsymbol \alpha^\top \mathbf {1}=1,\ \boldsymbol \alpha \geq 0,\ \mathbf {S} \geq 0,\notag\\ &{\mathbf W^{(i)} \geq \mathbf A^{(i)} \geq 0}, \; i=1, \dots, {v} \notag
\end{align}
\blue{where $\lambda_i$ is a parameter to incorporate the importance of the $i$-th view, and a higher value indicates greater importance.  Particularly, the parameter $\lambda_i$ is introduced to provide the users with an opportunity to incorporate prior knowledge through it. Yet in this paper, we mainly focus on the unsupervised multi-view clustering task, and the parameter $\lambda_i=1$ can be simply used for all views when no prior knowledge is involved. %
When prior knowledge is involved, the requirement for $\lambda_{i}$ is $\lambda_{i} > 0$, because negative value does not make sense and if one wants to set $\lambda_{i}=0$ then the $i$-th view can be simply removed from the objective. %
In the next section, the optimization problem will be theoretically analyzed, and a highly efficient algorithm will be developed to approximately solve it.
}

\section{Optimization} \label{sec:optimization}
\blue{Though the proposed framework, if optimized properly, can identify multi-view inconsistency and fuse consistent parts into a unified graph, the optimization of objective~\eqref{func:obj3} is difficult for three reasons.
Firstly, the objective function~\eqref{func:obj3} is not jointly convex on all variables, which rules out the possibility of \emph{directly} applying convex optimization techniques. 
Secondly, there are a lot of coupling between the variables, i.e., different variables are multiplied together, which causes the exponent of variables as large as 4. }
Thirdly and unfortunately, the optimization of objective~\eqref{func:obj3} turns out to be NP-hard, which will be explained in Section~\ref{sec:convergence}.
\blue{In the preliminary version of this paper \cite{liang2019consistency}, a projection based method is applied to optimize the objective. Although the projection based method is a good heuristic, we observe that the objective increases in some optimization iterations, and it may yield unsatisfactory clustering performance on some datasets, as we will show in the \app. }

\blue{Based on better understandings of the problem, we develop the following approaches to tackle these problems. 
Firstly, we simplify the constraints by proving one constraint can be automatically satisfied in Lemma~\ref{lemma:nonneg}. 
Secondly, objective~\eqref{func:obj3} can be rewritten as two forms of quadratic functions, corresponding to Subproblem~(1) and Subproblem~(3), respectively, so that we can optimize them alternately. }
Finally, Subproblem~(3) consists of at least $n$ nonconvex quadratic programs (QPs) with box constraints, which are difficult to solve since they are NP-hard \cite{burer2009nonconvex}. By proving that these QPs share a same Hessian that has a desired property (Lemma~\ref{lemma:pos eig}), we are able to modify the d.c. algorithm \cite{tao1998branch} to solve them all at once, which is much more efficient than solving them sequentially. 

\blue{To facilitate these approaches, the multi-view dense representation is {further} proposed, where the nonzero elements of adjacency matrices of all views are arranged into a dense matrix. 
A significant advantage of the proposed approaches is that they mainly involve matrix-vector and matrix-matrix multiplications, which contribute to their high efficiency for large-scale problems.}
\subsection{Constraint Simplification}
We first show that the constraint $\mathbf{S} \geq 0$ in Problem~\eqref{func:obj3} can be removed while the global minimizer(s) remains the same. 
Define the following sets:
\begin{align}
	&\mathcal{G}_0 = \{\boldsymbol \alpha \geq 0 \mid  \boldsymbol \alpha^\top \mathbf {1}=1 \},\\
	&\mathcal{G}_i = \{\mathbf A^{(i)} \mid \mathbf W^{(i)} \geq \mathbf A^{(i)} \geq 0 \}, \quad i=1, \dots, {v}, \\
	&\mathcal{G} = \mathcal{G}_0 \times \mathcal{G}_1 \times \dots \times \mathcal{G}_{v} \times \mathbb{R}^{n\times n}, \\
	&\mathcal{G}_{+} = \mathcal{G}_0 \times \mathcal{G}_1 \times \dots \times \mathcal{G}_{v} \times \mathbb{R}_{\geq 0}^{n\times n},  \\
	&\mathcal{G}_{-} = \mathcal{G}_0 \times \mathcal{G}_1 \times \dots \times \mathcal{G}_{v} \times (\mathbb{R}^{n\times n} \setminus \mathbb{R}_{\geq 0}^{n\times n}),
\end{align}
where $\times$ denotes the Cartesian product. Clearly, $\mathcal{G}_{+}, \mathcal{G}_{-} \subset \mathcal{G}$ and $\mathcal{G}_{+} \cup \mathcal{G}_{-} = \mathcal{G}, \mathcal{G}_{+} \cap \mathcal{G}_{-} = \varnothing$.
Then objective function \eqref{func:obj3} is denoted by $f \colon \mathcal{G} \to \mathbf R$.
The following lemma shows that with the constraint $\mathbf{S} \geq 0$ removed, the minimizer of the resulting problem still satisfies $\mathbf{S} \geq 0$.
\begin{lemma} \label{lemma:nonneg}
For every minimizer $x^*$ of the problem
\begin{equation} \label{func:obj4}
\min_{x} \; f(x), \quad\quad  \mathrm{s.t.} \;\ x \in \mathcal{G},
\end{equation}
we have
$x^* \in \mathcal{G}_{+}$. 
\end{lemma}
\begin{IEEEproof}
	Suppose $x^* \notin \mathcal{G}_{+}$. Then $x^* \in \mathcal{G}_{-}$. Suppose $x^* = (\boldsymbol \alpha, \mathbf A^{(1)}, \dots, \mathbf A^{(v)}, \mathbf{S})$, where $S_{pq} < 0$ for some $p,q\in\{1,\dots,n\}$. Let $\mathcal{V}=\{(p,q)\mid S_{pq} < 0\}$.
	Let $\tilde{x} = (\boldsymbol \alpha, \mathbf A^{(1)}, \dots, \mathbf A^{(v)}, \tilde{\mathbf{S}}) \in \mathcal{G}_{+}$, where $\tilde{\mathbf{S}} \in \mathbb{R}_{\geq 0}^{n\times n}$ such that $\tilde{S}_{pq} = 0$ for all $(p,q)\in \mathcal{V}$ and $\tilde{S}_{pq} = S_{pq}$ for all $(p,q)\notin \mathcal{V}$. 
	Let $c = \sum_{k=1}^{v} \lambda_{k} \sum_{(p,q)\notin\mathcal{V}}(\alpha_k A_{pq}^{(k)} - S_{pq})^2+$ {\small$\sum_{i,j=1}^{v} B_{ij} \lambda_{i} \lambda_{j} \alpha_i \alpha_j \operatorname{sum}((\mathbf W^{(i)}-\mathbf A^{(i)}) \circ (\mathbf W^{(j)}-\mathbf A^{(j)}))$}. Then
	\begin{align}
	f({x^*}) &= f(\boldsymbol \alpha, \mathbf A^{(1)}, \dots, \mathbf A^{(v)}, \mathbf{S}) \notag \\
	&= \sum_{k=1}^{v} \lambda_{k} \sum_{(p,q)\in \mathcal{V}} \Big(\alpha_k A_{pq}^{(k)} - S_{pq}\Big)^2 + c \notag \\
	&= \sum_{k=1}^{v} \lambda_{k} \sum_{(p,q)\in \mathcal{V}}  \Big( \big(\alpha_k A_{pq}^{(k)}\big)^2 + S_{pq}^2 - 2 S_{pq} \alpha_k A_{pq}^{(k)}\Big)\! +\! c \notag \\
	&> \sum_{k=1}^{v} \lambda_{k} \sum_{(p,q)\in \mathcal{V}} \Big(\alpha_k A_{pq}^{(k)}\Big)^2 + c \notag \\
	&= f(\tilde{x}). \notag
	\end{align}
	This contradicts that $x^*$ is a minimizer of the problem \eqref{func:obj4}. Therefore, we conclude that $x^* \in \mathcal{G}_{+}$.
\end{IEEEproof}
Since the minimizer of Problem~\eqref{func:obj4} cannot appear outside the region $\mathcal{G}_{+}$, we do not need the constraint $\mathbf{S} \geq 0$, which largely simplifies the problem. 

\subsection{Optimization Scheme} \label{sec:optim}
Due to the complex coupling of the variables, we adopt an alternating optimization scheme as follows. We first optimize the objective function over $\boldsymbol \alpha$ with $\mathbf {S}, \mathbf{A}^{(1)},\dots,\mathbf{A}^{(v)}$ fixed, and then over $\mathbf{S}$ with $\boldsymbol \alpha, \mathbf{A}^{(1)},\dots,\mathbf{A}^{(v)}$ fixed, and then over $\mathbf{A}^{(1)},\dots,\mathbf{A}^{(v)}$ with $\boldsymbol \alpha$ and $\mathbf {S}$ fixed. That is, we divide the problem into three subproblems and optimize one at a time. We repeat the procedures until the objective value converges. 
The objective~\eqref{func:obj3} can be rewritten in two different forms, as shown in \eqref{F:alpha} and \eqref{F:A}, which show that \eqref{func:obj3} is actually a quadratic form of variables $\{\alpha_{i}\}_{i=1}^{v}$, as in \eqref{F:alpha}, or quadratic forms of variables $\{A_{jk}^{(i)}\}_{i=1}^{v}$, as in \eqref{F:A}.
{%
\begin{align}
\begin{split}
&f(\boldsymbol \alpha, \mathbf A^{(1)}, \dots, \mathbf A^{(v)}, \mathbf{S}) = \sum_{i=1}^{v} \lambda_{i} \sum_{j,k} \big(\alpha_{i} A_{jk}^{(i)} - S_{jk} \big)^2 + \\
&\sum_{i, l = 1}^{v} B_{il} \lambda_{i} \lambda_{l} \alpha_{i} \alpha_{l} \sum_{j,k} \big(W_{jk}^{(i)} - A_{jk}^{(i)} \big) \big(W_{jk}^{(l)} - A_{jk}^{(l)} \big)
\end{split}
\\[5pt]
\begin{split} \label{F:alpha}
&\!=\!\sum_{i=1}^{v}\! \Big( \lambda_{i} \sum_{j,k} (A_{jk}^{(i)})^2 \Big) \alpha_i^2 - 2 \sum_{i=1}^{v} \Big( \lambda_{i} \sum_{j,k} A_{jk}^{(i)} S_{jk} \Big) \alpha_i + \\ 
&\!\sum_{i,l= 1}^{v}\!\! \Big( B_{il} \lambda_{i} \lambda_{l} \sum_{j,k} \big(W_{jk}^{(i)}\! -\! A_{jk}^{(i)} \big) \big(W_{jk}^{(l)} \!-\! A_{jk}^{(l)} \big) \Big)  \alpha_{i} \alpha_{l}\! +\! C_1
\end{split}
\\[5pt]
\begin{split} \label{F:A}
&=\sum_{j,k} \left[ \sum_{i=1}^{v} \lambda_{i} \alpha_{i}^2 (A_{jk}^{(i)})^2  +\!\! \sum_{i, l = 1}^{v} B_{il} \lambda_{i} \lambda_{l} \alpha_{i} \alpha_{l} A_{jk}^{(i)} A_{jk}^{(l)} \right. -  \\ 
&\left. 2\sum_{i=1}^{v} \Big( \lambda_{i} \alpha_{i} S_{jk} + \sum_{l=1}^{v} B_{il} \lambda_{i} \lambda_{l} \alpha_{i} \alpha_{l} W_{jk}^{(i)} \Big) A_{jk}^{(i)} \right] + C_2
\end{split}
\end{align}
}%
where $C_1$ is a quantity which does not depend on $\boldsymbol{\alpha}$, and $C_2$ is a quantity which does not depend on $\mathbf{A}^{(1)}, \dots, \mathbf{A}^{(v)}$. Let $q(\boldsymbol{\alpha}) = f(\boldsymbol \alpha, \mathbf A^{(1)}, \dots, \mathbf A^{(v)}, \mathbf{S}) - C_1$, i.e., the first two terms of \eqref{F:alpha}, and $p(\mathbf{A}^{(1)}, \dots, \mathbf{A}^{(v)} ) = f(\boldsymbol \alpha, \mathbf A^{(1)}, \dots, \mathbf A^{(v)}, \mathbf{S}) - C_2$, i.e., the first terms in \eqref{F:A}. Then Subproblem~(1) is to optimize $q(\boldsymbol{\alpha})$ w.r.t. $\boldsymbol{\alpha}$ and Subproblem~(3) is to optimize $p(\mathbf{A}^{(1)}, \dots, \mathbf{A}^{(v)})$.

\subsection{Multi-view Dense Representation} \label{sec:dense}
Since there are typically a lot of zero elements in the multi-view (sparse) adjacency matrices $\mathbf{W}^{(i)}$, many terms within the summation in \eqref{F:alpha} and \eqref{F:A} vanish. We can exploit this to make our algorithm more efficient. 
In our preliminary version \cite{liang2019consistency}, the optimization is performed on sparse matrices, but we propose a better approach by constructing a dense matrix with \emph{only} the nonzero elements from the sparse matrices of all views. %

Let $\mathcal{F}$ be the \emph{common} index set of the nonzero elements in the adjacency matrices of all views. %
Let $\mathbf{w}^{(i)}$ be a row vector by taking the non-zero elements from $\mathbf{W}^{(i)}$, corresponding to the indices in $\mathcal{F}$. Then we stack the row vectors $\mathbf{w}^{(1)}, \dots, \mathbf{w}^{(v)}$ to form a $v$-by-$n_e$ matrix $\mathbf{W}$ (note that the notation differs from $\mathbf{W}^{(i)}$), where $n_e$ is the number of elements in the non-zero index set $\mathcal{F}$. Similarly, we form a $v$-by-$n_e$ matrix $\mathbf{A}$ from $\mathbf{A}^{(1)}, \dots, \mathbf{A}^{(v)}$, according to $\mathcal{F}$. %
Then the inconsistent part for all views can be represented by the matrix $\mathbf{E} = \mathbf{W} - \mathbf{A}$ and the fused graph is represented by a row vector $\boldsymbol{s}$ of length $n_e$. Multi-view dense representation (MVDR) is more efficient than sparse matrix representation, since the dense matrix requires only \nicefrac{1}{3} memory of the sparse matrix and it is faster to access elements in a dense matrix than in a sparse one. For these reasons we use only MVDR in our optimization algorithm. 

We also normalize the multi-view adjacency matrices before performing optimization because normalization aids the optimization process and can reduce the total number of iterations. %
A typical normalization method is to divide the (non-negative) adjacency matrices by its sum. For the multi-view dense representation $\mathbf{W}$, we normalize each row separately.

\subsection{Subproblem~(1)} \label{sec:sub1}
With $\mathbf{A}$ and $\boldsymbol{s}$ fixed, we optimize $q(\boldsymbol{\alpha})$. Note that $q(\boldsymbol{\alpha})$ is a quadratic function of $\boldsymbol{\alpha}$. Formally, Subproblem~(1) is formulated as a standard quadratic program (StQP)~\cite{bomze1998standard}:
\begin{align} %
\min_{\boldsymbol \alpha} \quad & q(\boldsymbol{\alpha}) = \nicefrac{1}{2}\; \boldsymbol{\alpha}^\top \mathbf{H} \boldsymbol{\alpha} - \boldsymbol{\alpha}^\top \boldsymbol{c} \label{eq:stqp} \\
\text{s.t.} \quad &\boldsymbol{\alpha}^\top \mathbf {1}=1, \ \boldsymbol{\alpha} \geq 0\label{eq:alpha1}
\end{align}
where $\mathbf{H}$ and $\boldsymbol c$ are computed by \eqref{F:alpha} as we now explain. %
Let $\boldsymbol{h}$ be a vector where $h_i = \sum_{j=1}^{n_e} A^2_{ij}$, and let $\mathbf{T}$ be a $v$-by-$v$ diagonal matrix where $T_{ii} = \lambda_{i} h_i$.
Let $\mathbf{Z}$ and $\mathbf{P}$ be $v$-by-$v$ matrices defined as $Z_{ij} = B_{ij} \lambda_{i} \lambda_{j}$ and $\mathbf{P} = \mathbf{Z} \circ (\mathbf{E}\mathbf{E}^\top)$ where $\mathbf{E}=\mathbf{W}-\mathbf{A} \in \mathbb{R}^{v \times n_e}$ is the inconsistent parts for all views.
According to \eqref{F:alpha}, the Hessian of $q(\boldsymbol{\alpha})$ is $\mathbf{H} = 2( \mathbf{T} + \mathbf{P})$. 
By \eqref{F:alpha}, the linear coefficient of $q(\boldsymbol{\alpha})$ can be defined as $c_{i} = 2\lambda_{i} \mathbf{A}_{i} \boldsymbol{s}^\top$ where $\mathbf{A}_{i}$ is the $i$-th row of $\mathbf{A}$.

\blue{
We exploit the away-step Frank-Wolfe (AFW) algorithm \cite{nips15-frankwolfe, siam20-afw} (Algorithm~\ref{alg:afw}) to solve Subproblem~(1), which is able to identify the active set (i.e., the set $\{i : \alpha^*_i = 0\}$ for the minimizer $\bs{\alpha}^*$) in a natural way~\cite{siam19-active-set, siam20-afw}. Besides, the AFW algorithm has low computation at each iteration and thus adds little overheads to the whole graph learning algorithm. Maybe the biggest advantage of using the AFW algorithm is that it is an iterative algorithm, i.e., it accepts an initial point and iteratively updates the point by stepping towards a descent direction. As we will see next, the whole graph learning algorithm (Algorithm~\ref{alg:graph}) is itself iterative. Thus, every time we start AFW in Algorithm~\ref{alg:graph}, we can simply initialize $\bs \alpha$ in AFW as the latest $\bs \alpha$ instead of random initialization. %
The benefit of this initialization is that, if Algorithm~\ref{alg:graph} is close to convergence, the latest $\bs{\alpha}$ is also close to the minimizer \emph{of} AFW, which reduces the number of iterations \emph{within} AFW.}

\blue{We now elaborate the exact line search method at line~15 of Algorithm~\ref{alg:afw}. 
The purpose of exact line search is to determine the best step size $\eta$ such that the objective $f(\eta)$ in Eq.~\eqref{eq:stepsize} decreases the most. We substitute the new point $\bs \alpha + \eta \bs d$ into the objective $\nicefrac{1}{2}\; \bs{\alpha}^\top \mathbf{H} \bs{\alpha} - \bs{\alpha}^\top \boldsymbol{c}$ and get a quadratic function of the step size $\eta$:
\begin{align}
f(\eta) =& \nicefrac{1}{2}\; (\bs \alpha + \eta \bs d)^\top \mathbf{H} (\bs \alpha + \eta \bs d) - (\bs \alpha + \eta \bs d)^\top \boldsymbol{c} \label{eq:stepsize}\\
=& \nicefrac{1}{2}\; \eta^2 \bs{d}^\top \mathbf{H} \bs d + \eta (\bs{\alpha}^\top \mathbf{H} \bs d - \bs{d}^\top \boldsymbol{c}) + \textrm{constant}
\end{align}
The minimum of this quadratic function in the interval $(0, \eta^{\textrm{max}}]$ can be obtained by a simple analysis of its axis of symmetry. 
}

\begin{algorithm}%
	\begin{algorithmic}[1]
		\renewcommand{\algorithmicrequire}{\textbf{Input:}}
		\renewcommand{\algorithmicensure}{\textbf{Output:}}
		\newcommand{\fw}{{\mathcal{F}\mathcal{W}}}
		\Require $\mathbf H$ (Hessian), $\boldsymbol{c}$ (linear term), $\bs \alpha$ (initial point), $\epsilon$ (tolerance), $N$ (\# maximum iterations)
		\Ensure  $\bs{\alpha}$
		\For {$k= 1, \dots, N$}
		\State $\boldsymbol{g} \gets \mathbf H \bs{\alpha} - \boldsymbol{c}$ \Comment{derivative of the objective}
		\State $\boldsymbol{s} \gets \boldsymbol{e}_{\hat\imath}$, with $\hat\imath = \argmin_i \boldsymbol{g}_i $ \Comment{$\boldsymbol{e}_{\hat\imath}$: $\hat\imath$-th unit vector}
		\State $\bs{d}^\fw \gets \bs s - \bs \alpha$  \Comment{the Frank-Wolfe direction}
		\State $\bs v \gets \bs{e}_{\hat\jmath}$, with $\hat\jmath = \argmax_{j \in S} \bs{g}_j, \, S := \{j: \bs{\alpha}_j > 0\}$
		\State $\bs{d}^{\mathcal{A}} \gets \bs \alpha - \bs v$  \Comment{the away direction}
		\If{$-\bs{g}^\top \bs{d}^\fw \ge -\bs{g}^\top \bs{d}^{\mathcal{A}}$}
		\State $\bs d \gets \bs{d}^\fw$, and $\eta^{\textrm{max}} := 1$
		\Else
		\State $\bs d \gets \bs{d}^{\mathcal{A}}$, and $\eta^{\textrm{max}} := \bs{\alpha}_{\hat\jmath} / (1 - \bs{\alpha}_{\hat\jmath})$
		\EndIf
		\If{$- \bs{g}^\top \bs d \le \epsilon$}
		\State \Return{$\bs \alpha$} \Comment{$\bs \alpha$ is a stationary point}
		\EndIf
		\State Use exact line search to get step size $\eta \in (0, \eta^{\textrm{max}}]$
		\State $\bs \alpha \gets \bs \alpha + \eta \bs d$
		\EndFor
	\end{algorithmic}
	\caption{\blue{Away-step Frank-Wolfe algorithm for standard Quadratic Programming \eqref{eq:stqp}-\eqref{eq:alpha1}}}
	\label{alg:afw}
\end{algorithm}

\subsection{Subproblem~(2)} \label{subproblem 2}
In this section we update $\boldsymbol{s}$ with $\boldsymbol{\alpha}$ and $\mathbf{A}$ fixed.
Taking the derivative of objective function \eqref{func:obj3} with respect to $\mathbf{S}$ gives
\begin{equation*}
	2 (\sum_{i=1}^{v} \lambda_{i}) \mathbf {S} -  2 \sum_{i=1}^{v} \lambda_{i} \alpha_{i} \mathbf A^{(i)}= \mathbf{0}.
\end{equation*}
Thus, 
\begin{equation} \label{E:S0}
\mathbf {S}= \sum_{i=1}^{v} (\lambda_{i} \alpha_{i} \mathbf A^{(i)} )/ (\sum_{i=1}^{v} \lambda_{i}).
\end{equation}
In multi-view dense representation, \eqref{E:S0} becomes
\begin{equation} \label{E:S}
	\boldsymbol{s} = \boldsymbol{t}^\top \mathbf A
\end{equation}
where $\boldsymbol{t} = \boldsymbol{\lambda} \circ \boldsymbol{\alpha} / (\sum_{i=1}^{v} \lambda_{i})$.

\subsection{Subproblem~(3)} \label{subproblem 3}
In this section we update $\mathbf{A}$ with $\boldsymbol{\alpha}$ and $\boldsymbol{s}$ fixed. For a pair of fixed $(j,k)$, let $\boldsymbol{x} = [A_{jk}^{(1)}, \dots, A_{jk}^{(v)}]^\top$ (note that this is $\boldsymbol{x}$, not $\boldsymbol{\alpha}$) be the corresponding column of $\mathbf A$. Then the three terms within the brackets in \eqref{F:A} is a quadratic function of $\boldsymbol{x}$, for which we can formulate a quadratic program (QP) with upper and lower bound:
\begin{align}
\min_{\boldsymbol{x}} & \quad \nicefrac{1}{2}\; \boldsymbol{x}^\top \mathbf{D} \boldsymbol{x} - \boldsymbol{l}^\top \boldsymbol{x} \label{eq:sub3} \\
\text{s.t.} & \quad \mathbf{0} \leq \boldsymbol{x} \leq \boldsymbol{u}
\end{align}
where $\boldsymbol{u} = [W_{jk}^{(1)}, \dots, W_{jk}^{(v)}]^\top$ is the corresponding column of $\mathbf W$, and $\mathbf D$ and $\boldsymbol{l}$ is computed by \eqref{F:A} as we soon explain. For now, we need to solve a quadratic program for every pair of $(j,k)$. Fortunately, note that in \eqref{F:A}, summing over $(j, k)$ is the same as summing over the indices of nonzero elements in $\mathbf{A}^{(i)}$. Since the adjacency matrices are usually sparse, the actual number of quadratic programs (QPs) we need to solve is $n_e$ instead of $n^2$, where $n_e$ is the number of elements in $\mathcal{F}$ (see Section~\ref{sec:dense}). For example, when $k$NN graphs are used to construct similarity graphs, the number of QPs is $n_e = k n$, where $k$ is the number of nearest neighbors and $n$ is the number of data points. In multi-view dense representation, every column of $\mathbf A$ corresponds to a QP. Moreover, these QPs share the same Hessian $\mathbf D$ as we can see in \eqref{F:A}. With this observation, we find that Subproblem~(3) can be efficiently solved via the d.c. (difference of convex functions) optimization algorithms \cite{tao1998branch}, which requires only matrix multiplication and is simple to implement. Since the original d.c. algorithm (DCA) solves only one QP at a time, we modify it so that it can efficiently solve multiple QPs sharing the same Hessian all at once, which is presented in Algorithm~\ref{alg:dca}. To do so, we need to combine the linear coefficients of these QPs by stacking all $\boldsymbol{l}$ (column vectors) of each QP horizontally to form a $v$-by-$n_e$ matrix $\mathbf L$, and the \emph{combined} lower bound and upper bound of the QPs are $\mathbf 0$ and $\mathbf W$, respectively.

To compute $\mathbf D$ and $\mathbf L$, let $\mathbf K$ be a $v$-by-$v$ matrix defined as $K_{ij} = B_{ij} \lambda_{i} \lambda_{j} \alpha_{i} \alpha_{j}$. Let $\mathbf Q$ be a $v$-by-$v$ diagonal matrix defined as $Q_{ii} = \lambda_{i} \alpha_{i}^2$. According to \eqref{F:A}, the Hessian of all QPs is $\mathbf D = 2( \mathbf Q + \mathbf K)$. 
Let $\boldsymbol t = \boldsymbol{\lambda} \circ \boldsymbol{\alpha}$ and $\mathbf P = \boldsymbol t \circ \boldsymbol s$ where the second $\circ$ is broadcasting element-wise multiplication, i.e., $P_{ij} = t_{i} s_{j}$. The combined linear coefficient of all QPs is $\mathbf L = 2 (\mathbf P + \mathbf K \mathbf W)$. In Algorithm~\ref{alg:dca}, we need to compute $\rho$, the largest eigenvalue of $\mathbf D$, which is required to be positive\cite{tao1998branch}. Fortunately, this requirement is satisfied as shown in the following lemma.
\begin{lemma} \label{lemma:pos eig}
	The matrix $\mathbf D$ has at least one positive eigenvalue.
\end{lemma}
\begin{IEEEproof}
	By the definition of $\mathbf D$, for all $1 \le i \le v, \ D_{ii} = 2 ( Q_{ii} + K_{ii})$ where $Q_{ii} = \lambda_{i} \alpha_{i}^2$ and $K_{ii} = B_{ii} \lambda_{i}^2 \alpha_{i}^2$. Since $B_{ii} \ge 0$ and $\lambda_{i} > 0$ by our definitions, then $Q_{ii} \ge 0$ and $ K_{ii} \ge 0$. Note that for all $i \in \{1,\dots,v\}, \ \alpha_{i} \ge 0$ and $\sum_{i=1}^{v} \alpha_{i} = 1$ by constraints \eqref{eq:alpha1}. Thus, 
	there exists $j \in \{1,\dots,v\}$ such that $\alpha_{j} > 0$. %
	Then $Q_{jj} > 0$ and thus $D_{jj} > 0$. 
	Suppose $\mathbf D$ is negative semi-definite. Then for any vector $\boldsymbol{v} \in \mathbb R^v$, $\boldsymbol{v}^\top \mathbf D \boldsymbol{v} \le 0$. If we let $\boldsymbol{v} = \boldsymbol{e}_j$, where $\boldsymbol{e}_j \in \mathbb R^v$ is the $j$-th unit vector (i.e., its $j$-th element is $1$ and all other elements is $0$), then $\boldsymbol{e}_j^\top \mathbf D \boldsymbol{e}_j = D_{jj} > 0$. This leads to a contradiction and thus $\mathbf D$ is not negative semi-definite. Hence, $\mathbf D$ has at least one positive eigenvalue.
\end{IEEEproof}

Another reason to use DCA is that $\mathbf D$ can be indefinite, which can be seen by adjusting $\beta$, $\gamma$ and $\boldsymbol{\lambda}$. Consequently, the QPs in Subproblem~(3) are nonconvex and thus solving them is NP-hard\cite{burer2009nonconvex}, while DCA is designed for such nonconvex QPs\cite{tao1998branch}. 
\begin{algorithm}%
	\begin{algorithmic}[1]
		\renewcommand{\algorithmicrequire}{\textbf{Input:}}
		\renewcommand{\algorithmicensure}{\textbf{Output:}}
		\Require $\mathbf D$ (Hessian), $\mathbf L$ (combined linear coefficient), $\mathbf W$ (upper bound), $\mathbf A$ (initial point), $N$ (number of iterations)
		\Ensure  $\mathbf A$
		\State Compute $\rho$, the largest eigenvalue of $\mathbf D$. \footnotemark
		\State $\mathbf H \gets \rho \mathbf I - \mathbf D$ \Comment{$\mathbf I \in \mathbb R^{v \times v}$ is an identity matrix}
		\For {$i= 1, \dots, N$}
		\State $\mathbf Y \gets \mathbf H \mathbf A$
		\State $\mathbf A \gets (\mathbf Y + \mathbf L) / \rho$
		\State $\mathbf A \gets \mymiddle(0, \mathbf A, \mathbf W)$
		\EndFor
	\end{algorithmic}
	\caption{Parallelized DCA for Quadratic Programs with box constraints (modified Algorithm 2a in \cite{tao1998branch})}
	\label{alg:dca}
\end{algorithm}

In Algorithm~\ref{alg:dca}, the $\rho$ in line 1 can be obtained via the Implicit Restarted Lanczos Method \cite{sorensen1997implicitly}, and $\mymiddle(\cdot,\cdot,\cdot)$ denotes element-wise median operator, i.e., the $(i,j)$-th element of $\mymiddle(0, \mathbf A, \mathbf W) = \min(\max(0, A_{ij}), W_{ij})$. We found that Algorithm~\ref{alg:dca} converges very fast, usually in 3 iterations, as pointed out in \cite{tao1998branch}. Thus, $N=3$ is our default.

\subsection{Consistent Graph Learning Algorithm}
By alternatively solving the three subproblems, objective~\eqref{func:obj3} is optimized and the inconsistent part of each view is removed and the unified adjacency matrix of all view is iteratively learned. The complete consistent graph learning algorithm is presented in Algorithm~\ref{alg:graph}. We next analyze the convergence and complexity of the proposed algorithm. 

\subsubsection{Convergence Analysis} \label{sec:convergence}
\blue{While Subproblem~(3) is NP-hard and thus finding a global minimizer of objective~\eqref{func:obj3} is also NP-hard, we adopt an alternating optimization approach to iteratively update the local variables $\bs \alpha$, $\mathbf S$, and $\mathbf A$.  
In Subproblems (1), since both the Frank-Wolfe direction and the away direction are descent directions, the objective value always decreases via the AFW algorithm \cite{nips15-frankwolfe, siam20-afw}. In Subproblem~(2), the objective function \eqref{func:obj3} is a quadratic form of $\mathbf S$, and its Hessian is $(\sum_{i=1}^{v} \lambda_{i}) \mathbf I$ and thus positive definite. Hence, the stationary point obtained by Eq.~\eqref{E:S0} is a global minimizer of Subproblem~(2). 
In Subproblem~(3), the iteration converges to a Karush–Kuhn–Tucker point and it is a local minimum in most cases~\cite{tao1998branch}. 
Since objective \eqref{func:obj3} is bounded below by $0$, the alternatively solving of the 3 subproblems can converge to a local minimizer of the total optimization problem~\eqref{func:obj3} in most cases. }
\begin{algorithm}%
	\begin{algorithmic}[1]
		\renewcommand{\algorithmicrequire}{\textbf{Input:}}
		\renewcommand{\algorithmicensure}{\textbf{Output:}}
		\Require Adjacency matrices $\{\mathbf {W}^{(1)}, \dots, \mathbf {W}^{(v)}\}$, $\beta$, $\gamma$, $\boldsymbol{\lambda}$, $M$ (max iteration)
		\Ensure  Adjacency matrix of the unified graph $\mathbf {S}$
		\State Construct multi-view dense representation $\mathbf W$ according to the index set $\mathcal F$
		\State Normalization: $\mathbf W_{ij} = \mathbf W_{ij} / (\sum_{k=1}^{n_e} \mathbf W_{ik})$
		\State Initialization: $\mathbf A = \mathbf W$, $\boldsymbol{\alpha} = \mathbf 1 / v$; set $\boldsymbol{s}$ by Eq.~\eqref{E:S}
		\Repeat
		\State \blue{Use AFW~(Algorithm~\ref{alg:afw}) to update $\boldsymbol{\alpha}$}
		\State Update $\boldsymbol{s}$ by Eq.~\eqref{E:S}
		\State Use DCA (Algorithm~\ref{alg:dca}) to update $\mathbf A$
		\Until {convergence or max iteration is reached}
		\State Construct $\mathbf S$ with $\boldsymbol{s}$ according to the index set $\mathcal F$ %
	\end{algorithmic}
	\caption{Consistent Graph Learning}
	\label{alg:graph}
\end{algorithm}

\subsubsection{Complexity Analysis} \label{sec:complexity}
\blue{In Subproblem~(1), computing the Hessian $\mathbf H$ and linear coefficient $\boldsymbol{c}$ of the StQP requires $O(v^2 n_e)$ time and $O(v (v + n_e))$ space. Solving the StQP via the AFW algorithm requires $O(v^2)$ space and $O(v^2)$ time at each iteration and $O(m_1 v^2)$ time in total, where $m_1$ is the number of iterations in AFW. Then, solving Subproblem~(1) requires $O(v^2 (m_1 + n_e))$ time and $O(v (v + n_e))$ space. }
Solving Subproblem~(2) requires $O(v n_e)$ time and $O(v n_e)$ space. 
In Subproblem~(3), computing $\mathbf D$ and $\mathbf L$ requires $O(v^2 n_e)$ time and $O(v (v + n_e))$ space. In Algorithm~\ref{alg:dca}, computing the largest eigenvalue $\rho$ of $\mathbf D$ requires $O(v^2 \kappa(\mathbf D))$ time and $O(v^2)$ space \cite{hou2019eigen}, where $\kappa(\mathbf D)$ is the condition number of $\mathbf D$. \blue{The other steps in Algorithm~\ref{alg:dca} requires $O(v^2 n_e)$ time and $O(v (v + n_e))$ space.
Thus, the total time complexity of Algorithm~\ref{alg:graph} is $O(m_2 v^2 (m_1 + \kappa(\mathbf D) + n_e))$, where $m_2$ is the number of iterations of the loop in Algorithm~\ref{alg:graph}, and the total space complexity is $O(v (v + n_e))$.} 

\blue{
We have found that the iterations $m_2$ is typically less than 20 in our experiments (see Figure~\figconv) and thus $m_2$ can be considered as a constant. Since $n_e$ is typically very large in practice, it is reasonable to assume $n_e \gg m_1$ and $n_e \gg \kappa(\mathbf D)$. Then, the time complexity of Algorithm~\ref{alg:graph} is $O(v^2 n_e)$, which means the running time is linear in the number of edges $n_e$ and quadratic in the number of views $v$. Since $n_e = k n$ in a $k$NN graph where $k$ is the number of nearest neighbors and $n$ is the number of nodes (data points), the time complexity further reduces to $O(k n v^2)$. %
In the \app, we will report experimental results that show the empirical complexity aligns very well with the theoretical results here.}

\section{Two Graph Fusion Versions} \label{spectral clustering}
The proposed multi-view graph learning method is applicable to both similarity graphs and distance (dissimilarity) graphs. Thus, we extend our multi-view clustering framework into two graph fusion versions, Similarity Graph Fusion (SGF) and Distance Graph Fusion (DGF). The reason to fuse distance matrix is that distance may better reflect the relationship between data points than similarity as we soon explain in Section~\ref{sec:dgf}. 

\subsection{Similarity Graph Fusion}
\label{sec:sgf} %
Similarity Graph Fusion (SGF) fuses multiple similarity graphs into one. If full similarity graphs are available, we first construct $k$NN similarity graphs, which means only the edges connecting a node and its $k$-nearest neighbors are kept on the similarity graph \cite{von2007tutorial}. %
It is worth noting that the $k$NN graphs used in our algorithms are slightly different from the usual ones. The difference is that, when constructing the $k$NN graph for a view, we keep the edge connecting nodes $x_i$ and $x_j$ if $x_i$ is among the $k$-nearest neighbors of $x_j$ in \textit{any} view.
Then the positions of nonzero elements in the similarity matrices for all views will be the same (i.e., the index set $\mathcal{F}$), and we use them to construct the multi-view dense representation. Note that this will cause the number of nonzero elements in each row of the learned unified matrix greater than $k$. 
Hence, we select the $k$NNs for each node after learning the unified graph. %

As $k$NNs capture the local structure by preserving the edges of the nearest neighbors, we further normalize these edges in $k$NN distance graphs and strengthen a small portion of strong edges (associated with low distance) from a global perspective \emph{before} learning the unified graph. Let $\mathbf{D}$ be the edges in a $k$NN distance graph, and let $\mu = \mean(\mathbf{D}), \sigma=\std(\mathbf{D})$ be the mean and standard deviation of these edges, respectively. Without loss of generality, the edges that are $1\ \sigma$ lower than the mean distance are considered the strong edges and will be set to zero. Mathematically, the normalization for $k$NN graph is 
\begin{equation} \label{eq:norm}
	\mathbf{D} = \max((\mathbf{D}-\mu+\sigma)/\sigma, 0) = \max((\mathbf{D}-\mu)/\sigma+1, 0),
\end{equation}
where all operations are element-wise. 
Note that the distance of 0 will be transformed to the maximum similarity of 1 by means of the Gaussian kernel mapping and hence the strong edges that are set to 0 are ``strengthened''. More importantly, the normalization of $k$NN graphs provides a baseline for the graph learning algorithm to compare edges from multiple views with different mean and different standard deviations, which aids the process of fusing them into one unified graph. 

Lastly, we perform spectral clustering on the final unified graph to obtain the clustering results. The Similarity Graph Fusion algorithm for spectral clustering is summarized in Algorithm~\ref{alg:sgf}. 	
\begin{algorithm}%
	\begin{algorithmic}[1]
		\renewcommand{\algorithmicrequire}{\textbf{Input:}}
		\renewcommand{\algorithmicensure}{\textbf{Output:}}
		\Require {Dataset with $v$ views $\mathcal{X}=\left\{\mathbf{X}^{(1)}, \dots, \mathbf{X}^{(v)}\right\}$, number of clusters $n_c$, $\beta, \gamma$ (for Algorithm~\ref{alg:graph}), $k$ (number of $k$NN)}
		\Ensure {Cluster indicator vector $\mathbf{c}$}
		\State Construct $k$NN distance graphs $\{\mathbf{W}^{(1)}, \dots, \mathbf{W}^{(v)}\}$ that share neighbors across views
		\State Apply Eq.~\eqref{eq:norm} to $\mathbf{W}^{(i)}$ for $i=1,\dots,v$
		\State Apply Gaussian kernel function to transform $k$NN distance graphs to $k$NN similarity graphs
		\State Use Algorithm~\ref{alg:graph} to obtain the unified similarity matrix $\mathbf{S}$ from the $k$NN similarity graphs
		\State Keep the $k$ largest elements at each row of $\mathbf{S}$ and set other elements to $0$
		\State $\mathbf{S}\gets (\mathbf{S}+\mathbf{S}^\top) / 2$
		\State Perform spectral clustering on $\mathbf{S}$ to obtain cluster indicator vector $\mathbf{c}$
	\end{algorithmic}
	\caption{Similarity Graph Fusion for Spectral Clustering}
	\label{alg:sgf}
\end{algorithm}

\subsection{Distance (Dissimilarity) Graph Fusion}
\label{sec:dgf}
Distance (Dissimilarity) Graph Fusion (DGF) learns the unified graph directly from multiple distance (dissimilarity) graphs, since fusion of distance may better preserve the relationship between nodes than the fusion of similarity. We know that distance is transformed to similarity with a kernel (similarity) function, typically with the Gaussian kernel 
$k(\mathbf x_{i}, \mathbf x_{j})=\exp \big(-\big(d(\mathbf x_{i}, \mathbf x_{j})\big)^{2} /\left(2 \rho^{2}\right)\big)$
where $d(\mathbf x_{i}, \mathbf x_{j})$ is the distance between $\mathbf x_{i}$ and $\mathbf x_{j}$ under some metric. 
The kernel function may bias the intrinsic relationship between nodes in the original graphs and exert a negative influence on the consistent graph learning process. 
Therefore, we suggest directly learning a unified graph from the distance (dissimilarity) graphs of all views. %
Then we apply a kernel (similarity) function to the learned unified graph to transform distance to similarity. That is, line 3 and line 4 in Algorithm~\ref{alg:sgf} become\HRule[6pt] \noindent
3: Use Algorithm~\ref{alg:graph} to obtain the unified distance matrix from the $k$NN distance graphs\\ \noindent
4: Apply Gaussian kernel function to transform the unified distance graph to unified similarity graph $\mathbf{S}$ \HRule[6pt] \noindent

A natural question in practice is how to measure the distance (dissimilarity) between data points (nodes). A widely used metric is Euclidean distance while the choice largely depends on the applications. If the features are words frequency, the cosine distance is more suitable than Euclidean distance and it is calculated as
$d(\mathbf x_{i}, \mathbf x_{j})= 1 - {\mathbf x_{i}^\top \mathbf x_{j}} / (\|\mathbf x_{i}\| \|\mathbf x_{j}\|)$.

\section{{Experiments}}
\label{sec:experiment}
We perform extensive experiments to compare the two proposed graph learning-based multi-view spectral clustering algorithms, namely {SGF} and {DGF}, against seven state-of-the-art multi-view spectral clustering algorithms, namely,
Co-Regularized Spectral Clustering ({CoReg})\cite{kumar2011co},
Robust Multi-view Spectral Clustering ({RMSC})\cite{xia2014robust},
Affinity Aggregation for Spectral Clustering ({AASC})\cite{huang2012affinity},
Weighted Multi-view Spectral Clustering based on spectral perturbation ({WMSC})\cite{zong2018weighted},
multi-view clustering via Adaptively Weighted Procrustes ({AWP})\cite{nie2018multiview},
Graph Learning for Multi-view Clustering ({MVGL})\cite{zhan2017graph},
and Multi-view Consensus Graph Clustering ({MCGC})\cite{Zhan2018}.
In addition, the conventional Spectral Clustering ({SC})\cite{ng2002spectral} is also performed on each view of the datasets, and the best single-view SC performance is reported.

\begin{table}%
	\centering
	\caption{Statistics of the real-world datasets}
	\begin{tabular}{lccc}
		\toprule
		Dataset & \# views & \# classes & \# instances \\
		\midrule
		ORL   & 3     & 40    & 400 \\
		Yale  & 3     & 15    & 165 \\
		Reuters & 5     & 6     & 1200 \\
		BBCSport & 2     & 5     & 544 \\
		NUS-WIDE & 5     & 31    & 2000 \\
		Reuters-21578 & 5     & 6     & 1500 \\
		MSRC-v1 & 5     & 7     & 210 \\
		CiteSeer & 2     & 6     & 3312 \\
		ALOI & 4     & 100   & 10800 \\
		Flower17 & 7     & 17    & 1360 \\
		Caltech101 & 6     & 102   & 9144 \\
		UCI Digits & 6     & 10    & 2000 \\
		\bottomrule
	\end{tabular}
	\label{tab:stat}%
\end{table}

\subsection{Datasets and Evaluation Metrics}
Except the datasets used in the preliminary version of this paper, we conduct experiments on more datasets in this paper, with a total of 12 datasets as we now introduce. 
The ORL dataset contains 400 face images of 40 distinct subjects.\footnote{\url{http://cam-orl.co.uk/facedatabase.html}} 
The Yale dataset contains 165 gray-scale images of 15 individuals.\footnote{\url{http://cvc.cs.yale.edu/cvc/projects/yalefaces/yalefaces.html}} 
The Reuters dataset contains 1200 documents, where each document is in 5 languages (views).\footnote{\url{http://lig-membres.imag.fr/grimal/data.html}} 
The BBCSport dataset consists of 544 documents from the BBC Sport website.\footnote{\url{http://mlg.ucd.ie/datasets/segment.html}} 
The NUS-WIDE dataset %
contains multi-view features extracted from images of the NUS-WIDE-OBJECT dataset reported in \cite{nus-wide-civr09}. 
The Reuters-21578 dataset is a collection of documents that appeared on Reuters news in 1987.\footnote{\url{https://archive.ics.uci.edu/ml/datasets/reuters-21578+text+categorization+collection}} 
The MSRC-v1 dataset\cite{winn2005locus} contains 240 pixel-wise labeled images. %
The CiteSeer dataset contains 3312 documents.\footnote{\url{http://lig-membres.imag.fr/grimal/data.html}} 
The ALOI dataset\footnote{\url{https://elki-project.github.io/datasets/multi_view}} is a collection of 110250 images of 1000 small objects\cite{geusebroek2005amsterdam}. Since it is too large for some algorithms such as CoReg and RMSC, we follow Houle et al.\cite{houle2011knowledge} to use a subset. 
The Flower17 dataset consists of images of 17 categories of flower\footnote{\url{http://www.robots.ox.ac.uk/~vgg/data/flowers/17/}}, and the multi-view features are extracted by Nilsback and Zisserman\cite{nilsback2006visual, nilsback2008automated}. 
The Caltech101 dataset contains pictures of objects belonging to 101 categories.\footnote{\url{http://www.vision.caltech.edu/Image_Datasets/Caltech101/}} 
The UCI Digits dataset contains 2000 images of handwritten digits.\footnote{\url{https://archive.ics.uci.edu/ml/datasets/Multiple+Features}} 
The statistics of the datasets are summarized in Table~\ref{tab:stat} and more details of these datasets can be found in the \app.

\newcommand{\tib}[1]{\textit{\textbbf{#1}}}
\newcommand{\tbb}[1]{\textbbf{#1}}

\begin{table*}[!t]%
	\centering
	\caption{Average clustering scores and standard deviation (\%) over 10 runs by different multi-view spectral clustering methods; the best score and the second best score in each row are highlighted in bold and italic bold, respectively; the last two methods are our algorithms.}
	\resizebox{\textwidth}{!}{%
		\setlength{\tabcolsep}{3pt}
		\renewcommand{\arraystretch}{1.15}
		\begin{tabular}{clcccccccccc}
			\toprule
			{Metric} & {Dataset} & AASC  & AWP   & CoReg & MCGC  & MVGL  & RMSC  & WMSC  & SC (best) & SGF   & DGF \\
			\midrule
			\multirow{12}[0]{*}{NMI} 
			& ORL & 86.74$_{\pm.87}$ & 85.60$_{\pm.00}$ & 90.49$_{\pm.71}$ & 89.39$_{\pm.00}$ & 83.79$_{\pm.00}$ & 90.67$_{\pm.62}$ & 90.33$_{\pm.52}$ & 90.78$_{\pm.50}$ &\tib{91.81}$_{\pm.47}$ & \tbb{91.82}$_{\pm.30}$ \\
			& Yale & 66.39$_{\pm2.1}$ & 69.42$_{\pm.00}$ & 71.57$_{\pm.90}$ & 67.17$_{\pm.00}$ & 65.95$_{\pm.00}$ & 70.25$_{\pm1.4}$ & 71.89$_{\pm.76}$ & 71.14$_{\pm.76}$ &\tbb{74.04}$_{\pm.00}$ & \tib{73.90}$_{\pm.00}$ \\
			& Reuters & $\,\;$7.89$_{\pm.00}$ & 10.78$_{\pm.00}$ & 10.80$_{\pm.14}$ & $\,\;$9.66$_{\pm.00}$ & $\,\;$7.89$_{\pm.00}$ & $\,\;$9.38$_{\pm.88}$ & $\,\;$7.53$_{\pm.03}$ & 14.19$_{\pm.15}$ &\tib{14.99}$_{\pm.00}$ & \tbb{16.31}$_{\pm.08}$ \\
			& BBCSport & 64.20$_{\pm.00}$ & 78.13$_{\pm.00}$ & 91.87$_{\pm.00}$ & 79.62$_{\pm.00}$ & 68.05$_{\pm.00}$ & 71.77$_{\pm.00}$ & 67.72$_{\pm.00}$ & 87.11$_{\pm.00}$ &\tbb{93.96}$_{\pm.00}$ & \tib{92.68}$_{\pm.00}$ \\
			& NUS-WIDE & 17.83$_{\pm.43}$ & 15.96$_{\pm.00}$ & 18.95$_{\pm.12}$ & 14.55$_{\pm.00}$ & $\,\;$5.50$_{\pm.00}$ & 18.95$_{\pm.29}$ & 19.03$_{\pm.22}$ & 17.33$_{\pm.33}$ &\tib{19.52}$_{\pm.21}$ & \tbb{19.96}$_{\pm.22}$ \\
			& Reuters-21578 & 11.06$_{\pm.00}$ & 10.54$_{\pm.00}$ & 29.80$_{\pm.25}$ & 11.43$_{\pm.00}$ & $\,\;$8.63$_{\pm.00}$ & 13.28$_{\pm.31}$ & 25.81$_{\pm.88}$ & 27.27$_{\pm.13}$ &\tbb{31.40}$_{\pm.11}$ & \tib{31.03}$_{\pm.01}$ \\
			& MSRC-v1 & 69.78$_{\pm.14}$ & 67.71$_{\pm.00}$ & 74.57$_{\pm.09}$ & 71.80$_{\pm.00}$ & 65.58$_{\pm.00}$ & 68.87$_{\pm1.9}$ & 72.11$_{\pm.40}$ & 64.78$_{\pm1.0}$ &\tib{78.91}$_{\pm.00}$ & \tbb{80.98}$_{\pm.00}$ \\
			& CiteSeer & 15.74$_{\pm1.6}$ & $\,\;$8.90$_{\pm.00}$ & 34.16$_{\pm.02}$ & 17.99$_{\pm.00}$ & $\,\;$1.48$_{\pm.00}$ & 33.39$_{\pm.00}$ & 32.22$_{\pm2.1}$ & 17.63$_{\pm1.1}$ &\tib{37.55}$_{\pm.00}$ & \tbb{38.47}$_{\pm.05}$ \\
			& ALOI & 35.68$_{\pm.50}$ & 69.90$_{\pm.00}$ & 85.36$_{\pm.57}$ & 69.75$_{\pm.01}$ & 46.94$_{\pm.00}$ & 82.45$_{\pm.68}$ & 84.22$_{\pm.17}$ & 80.18$_{\pm.45}$ &\tbb{91.08}$_{\pm.34}$ & \tib{90.98}$_{\pm.28}$ \\
			& Flower17 & 52.34$_{\pm1.1}$ & 46.58$_{\pm.00}$ & 55.75$_{\pm1.1}$ & 44.38$_{\pm.00}$ & 22.51$_{\pm.00}$ & 53.39$_{\pm.87}$ & 56.25$_{\pm.74}$ & 47.34$_{\pm.29}$ &\tbb{66.00}$_{\pm.70}$ & \tib{65.34}$_{\pm.78}$ \\
			& Caltech101 & 37.88$_{\pm.69}$ & 44.52$_{\pm.00}$ & 45.85$_{\pm.24}$ & 41.97$_{\pm.00}$ & 14.13$_{\pm.00}$ & 41.52$_{\pm.33}$ & 45.81$_{\pm.25}$ & \textbbf{48.41}$_{\pm.20}$ &{46.09}$_{\pm.36}$ & \tib{46.44}$_{\pm.21}$ \\
			& UCI Digits & 87.07$_{\pm.00}$ & 92.67$_{\pm.00}$ & 94.74$_{\pm.00}$ & 83.70$_{\pm.00}$ & 89.24$_{\pm.00}$ & 78.08$_{\pm1.3}$ & 86.91$_{\pm.03}$ & 92.50$_{\pm.04}$ &\tib{95.63}$_{\pm.00}$ & \tbb{95.77}$_{\pm.00}$ \\
			\midrule
			\multirow{12}[0]{*}{ACC} 
			& ORL & 76.22$_{\pm1.4}$ & 71.50$_{\pm.00}$ & 82.15$_{\pm2.0}$ & 78.25$_{\pm.00}$ & 71.25$_{\pm.00}$ & 80.70$_{\pm1.4}$ & 81.42$_{\pm1.4}$ & 80.88$_{\pm.99}$ &\tbb{85.05}$_{\pm1.0}$ & \tib{84.50}$_{\pm.53}$ \\
			& Yale & 65.88$_{\pm1.5}$ & 67.27$_{\pm.00}$ & 68.42$_{\pm.34}$ & 61.82$_{\pm.00}$ & 64.85$_{\pm.00}$ & 68.79$_{\pm1.4}$ & 69.70$_{\pm.57}$ & 69.21$_{\pm.75}$ &\tbb{70.91}$_{\pm.00}$ & \tbb{70.91}$_{\pm.00}$ \\
			& Reuters & 19.75$_{\pm.00}$ & 25.17$_{\pm.00}$ & 24.41$_{\pm.26}$ & 23.92$_{\pm.00}$ & 19.67$_{\pm.00}$ & 23.50$_{\pm1.7}$ & 21.00$_{\pm.00}$ & 29.38$_{\pm.26}$ &\tib{29.97}$_{\pm.07}$ & \tbb{31.90}$_{\pm.14}$ \\
			& BBCSport & 67.46$_{\pm.00}$ & 89.15$_{\pm.00}$ & 97.61$_{\pm.00}$ & 90.44$_{\pm.00}$ & 73.16$_{\pm.00}$ & 81.80$_{\pm.00}$ & 67.32$_{\pm.08}$ & 95.96$_{\pm.00}$ &\tbb{98.35}$_{\pm.00}$ & \tib{97.98}$_{\pm.00}$ \\
			& NUS-WIDE & 15.70$_{\pm.19}$ & 14.60$_{\pm.00}$ & 14.95$_{\pm.09}$ & 12.75$_{\pm.00}$ & 13.85$_{\pm.00}$ & 15.49$_{\pm.62}$ & 15.02$_{\pm.10}$ & 13.86$_{\pm.47}$ &\tib{15.99}$_{\pm.64}$ & \tbb{16.39}$_{\pm.51}$ \\
			& Reuters-21578 & 36.00$_{\pm.00}$ & 35.47$_{\pm.00}$ & 50.33$_{\pm.67}$ & 32.80$_{\pm.00}$ & 28.93$_{\pm.00}$ & 33.87$_{\pm.51}$ & 47.09$_{\pm.26}$ & 44.66$_{\pm.34}$ &\tbb{51.64}$_{\pm.28}$ & \tib{50.59}$_{\pm.03}$ \\
			& MSRC-v1 & 77.33$_{\pm.25}$ & 76.19$_{\pm.00}$ & \tib{85.08}$_{\pm.27}$ & 84.76$_{\pm.00}$ & 68.10$_{\pm.00}$ & 71.05$_{\pm1.7}$ & 76.52$_{\pm.45}$ & 67.29$_{\pm.84}$ &{80.48}$_{\pm.00}$ & \tbb{87.14}$_{\pm.00}$ \\
			& CiteSeer & 36.32$_{\pm2.7}$ & 30.89$_{\pm.00}$ & 59.09$_{\pm.02}$ & 43.72$_{\pm.00}$ & 21.50$_{\pm.00}$ & 57.85$_{\pm.00}$ & 56.26$_{\pm3.4}$ & 40.41$_{\pm1.1}$ &\tib{63.40}$_{\pm.01}$ & \tbb{63.64}$_{\pm.09}$ \\
			& ALOI & 15.90$_{\pm.41}$ & 59.04$_{\pm.00}$ & 77.46$_{\pm1.5}$ & 56.62$_{\pm.00}$ & 42.47$_{\pm.00}$ & 77.04$_{\pm2.6}$ & 78.22$_{\pm.59}$ & 68.65$_{\pm1.3}$ &\tib{84.17}$_{\pm1.6}$ & \tbb{84.18}$_{\pm1.3}$ \\
			& Flower17 & 51.62$_{\pm1.4}$ & 44.85$_{\pm.00}$ & 55.96$_{\pm2.0}$ & 43.90$_{\pm.00}$ & 25.00$_{\pm.00}$ & 54.00$_{\pm2.1}$ & 55.88$_{\pm1.4}$ & 43.47$_{\pm.97}$ &\tbb{68.06}$_{\pm1.4}$ & \tib{67.88}$_{\pm1.6}$ \\
			& Caltech101 & 23.80$_{\pm.77}$ & \tib{26.22}$_{\pm.00}$ & 25.34$_{\pm.93}$ & 23.00$_{\pm.00}$ & 13.44$_{\pm.00}$ & 22.77$_{\pm.93}$ & 23.29$_{\pm.67}$ & \textbbf{26.74}$_{\pm.54}$ &{23.45}$_{\pm.44}$ & {23.34}$_{\pm.89}$ \\
			& UCI Digits & 84.55$_{\pm.00}$ & 96.85$_{\pm.00}$ & 97.65$_{\pm.00}$ & 82.40$_{\pm.00}$ & 86.05$_{\pm.00}$ & 78.94$_{\pm2.0}$ & 87.02$_{\pm.04}$ & 96.59$_{\pm.03}$ &\tib{98.10}$_{\pm.00}$ & \tbb{98.20}$_{\pm.00}$ \\
			\midrule
			\multirow{12}[0]{*}{ARI} 
			& ORL & 62.89$_{\pm2.4}$ & 66.34$_{\pm.00}$ & 75.38$_{\pm1.8}$ & 70.76$_{\pm.00}$ & 46.00$_{\pm.00}$ & 75.19$_{\pm1.6}$ & 74.44$_{\pm1.4}$ & 74.63$_{\pm1.3}$ &\tbb{78.80}$_{\pm1.0}$ & \tib{77.95}$_{\pm.71}$ \\
			& Yale & 42.36$_{\pm4.0}$ & 49.31$_{\pm.00}$ & 51.42$_{\pm1.6}$ & 47.35$_{\pm.00}$ & 43.81$_{\pm.00}$ & 51.43$_{\pm2.1}$ & 51.95$_{\pm1.2}$ & 51.82$_{\pm1.2}$ &\tbb{55.07}$_{\pm.00}$ & \tib{54.83}$_{\pm.00}$ \\
			& Reuters & $\,\;$1.26$_{\pm.00}$ & $\,\;$2.16$_{\pm.00}$ & $\,\;$2.24$_{\pm.04}$ & $\,\;$1.71$_{\pm.00}$ & $\,\;$1.25$_{\pm.00}$ & $\,\;$2.32$_{\pm.41}$ & $\,\;$1.64$_{\pm.00}$ & $\,\;$6.00$_{\pm.07}$ &\tib{6.52}$_{\pm.02}$ & \tbb{8.90}$_{\pm.16}$ \\
			& BBCSport & 52.33$_{\pm.00}$ & 80.45$_{\pm.00}$ & 93.92$_{\pm.00}$ & 79.83$_{\pm.00}$ & 58.35$_{\pm.00}$ & 70.78$_{\pm.00}$ & 55.43$_{\pm.02}$ & 89.75$_{\pm.00}$ &\tbb{95.53}$_{\pm.00}$ & \tib{94.76}$_{\pm.00}$ \\
			& NUS-WIDE & $\,\;$4.13$_{\pm.18}$ & $\,\;$3.75$_{\pm.00}$ & $\,\;$4.85$_{\pm.11}$ & $\,\;$2.43$_{\pm.00}$ & $\,\;$0.16$_{\pm.00}$ & $\,\;$4.53$_{\pm.30}$ & $\,\;$4.71$_{\pm.12}$ & $\,\;$4.38$_{\pm.23}$ &\tib{4.92}$_{\pm.33}$ & \tbb{5.78}$_{\pm.27}$ \\
			& Reuters-21578 & $\,\;$2.44$_{\pm.00}$ & $\,\;$3.02$_{\pm.00}$ & 19.25$_{\pm.27}$ & $\,\;$2.87$_{\pm.00}$ & $\,\;$0.24$_{\pm.00}$ & $\,\;$3.22$_{\pm.50}$ & 17.23$_{\pm.30}$ & \tib{23.09}$_{\pm.44}$ &\tbb{23.54}$_{\pm.05}$ & {21.31}$_{\pm.00}$ \\
			& MSRC-v1 & 59.90$_{\pm.18}$ & 62.25$_{\pm.00}$ & 69.47$_{\pm.66}$ & 68.09$_{\pm.00}$ & 49.67$_{\pm.00}$ & 55.03$_{\pm2.4}$ & 65.02$_{\pm.55}$ & 54.40$_{\pm1.4}$ &\tib{72.31}$_{\pm.00}$ & \tbb{75.35}$_{\pm.00}$ \\
			& CiteSeer & 12.08$_{\pm1.2}$ & $\,\;$2.99$_{\pm.00}$ & 31.92$_{\pm.02}$ & 12.11$_{\pm.00}$ & $\,\;$-0.02$_{\pm.00}$ & 24.97$_{\pm.00}$ & 25.45$_{\pm.53}$ & 10.12$_{\pm.54}$ &\tib{37.63}$_{\pm.01}$ & \tbb{38.15}$_{\pm.08}$ \\
			& ALOI & $\,\;$6.39$_{\pm.34}$ & 47.42$_{\pm.00}$ & 69.15$_{\pm1.6}$ & 41.61$_{\pm.01}$ & $\,\;$2.48$_{\pm.00}$ & 65.61$_{\pm1.5}$ & 68.03$_{\pm.52}$ & 56.13$_{\pm1.8}$ &\tbb{79.15}$_{\pm1.3}$ & \tib{78.27}$_{\pm.40}$ \\
			& Flower17 & 28.82$_{\pm2.2}$ & 30.36$_{\pm.00}$ & 39.13$_{\pm2.0}$ & 27.51$_{\pm.00}$ & $\,\;$3.02$_{\pm.00}$ & 36.77$_{\pm1.6}$ & 40.24$_{\pm1.1}$ & 26.89$_{\pm.60}$ &\tbb{52.27}$_{\pm1.2}$ & \tib{51.60}$_{\pm1.3}$ \\
			& Caltech101 & $\,\;$7.18$_{\pm1.5}$ & 15.28$_{\pm.00}$ & \tib{17.33}$_{\pm1.1}$ & 13.84$_{\pm.00}$ & $\,\;$-0.55$_{\pm.00}$ & \textbbf{21.57}$_{\pm1.7}$ & 15.39$_{\pm.92}$ & 16.45$_{\pm.53}$ &{14.60}$_{\pm.46}$ & {14.07}$_{\pm.84}$ \\
			& UCI Digits & 81.26$_{\pm.00}$ & 93.14$_{\pm.00}$ & 94.86$_{\pm.00}$ & 76.81$_{\pm.00}$ & 83.78$_{\pm.00}$ & 71.37$_{\pm2.2}$ & 82.22$_{\pm.04}$ & 92.60$_{\pm.06}$ &\tib{95.82}$_{\pm.00}$ & \tbb{96.04}$_{\pm.00}$ \\
			\midrule
			\multirow{12}[0]{*}{purity} 
			& ORL & 80.20$_{\pm1.2}$ & 72.75$_{\pm.00}$ & 84.82$_{\pm1.2}$ & 83.00$_{\pm.00}$ & 77.00$_{\pm.00}$ & 84.60$_{\pm1.1}$ & 84.40$_{\pm1.1}$ & 83.80$_{\pm.93}$ &\tbb{87.65}$_{\pm.68}$ & \tib{86.85}$_{\pm.68}$ \\
			& Yale & 66.00$_{\pm1.5}$ & 67.88$_{\pm.00}$ & 68.48$_{\pm.29}$ & 63.03$_{\pm.00}$ & 64.85$_{\pm.00}$ & 69.27$_{\pm1.3}$ & 69.70$_{\pm.57}$ & 70.24$_{\pm.78}$ &\tbb{70.91}$_{\pm.00}$ & \tbb{70.91}$_{\pm.00}$ \\
			& Reuters & 24.00$_{\pm.00}$ & 28.33$_{\pm.00}$ & 28.09$_{\pm.21}$ & 28.08$_{\pm.00}$ & 24.00$_{\pm.00}$ & 27.23$_{\pm1.6}$ & 25.08$_{\pm.00}$ & 34.09$_{\pm.05}$ &\tib{34.42}$_{\pm.00}$ & \tbb{35.97}$_{\pm.05}$ \\
			& BBCSport & 74.26$_{\pm.00}$ & 89.15$_{\pm.00}$ & 97.61$_{\pm.00}$ & 90.44$_{\pm.00}$ & 75.55$_{\pm.00}$ & 82.17$_{\pm.00}$ & 74.82$_{\pm.00}$ & 95.96$_{\pm.00}$ &\tbb{98.35}$_{\pm.00}$ & \tib{97.98}$_{\pm.00}$ \\
			& NUS-WIDE & 23.92$_{\pm.40}$ & 22.85$_{\pm.00}$ & 24.92$_{\pm.14}$ & 22.40$_{\pm.00}$ & 15.70$_{\pm.00}$ & 24.48$_{\pm.30}$ & 25.68$_{\pm.53}$ & 25.75$_{\pm.34}$ &\tib{25.79}$_{\pm.37}$ & \tbb{26.92}$_{\pm.54}$ \\
			& Reuters-21578 & 38.07$_{\pm.00}$ & 37.67$_{\pm.00}$ & 56.97$_{\pm.25}$ & 43.47$_{\pm.00}$ & 33.00$_{\pm.00}$ & 44.05$_{\pm.36}$ & 51.95$_{\pm1.4}$ & 52.11$_{\pm.05}$ &\tbb{57.97}$_{\pm.28}$ & \tib{56.99}$_{\pm.03}$ \\
			& MSRC-v1 & 77.33$_{\pm.25}$ & 79.52$_{\pm.00}$ & \tib{85.08}$_{\pm.27}$ & 84.76$_{\pm.00}$ & 72.86$_{\pm.00}$ & 76.14$_{\pm1.8}$ & 81.14$_{\pm.25}$ & 73.19$_{\pm.81}$ &{83.81}$_{\pm.00}$ & \tbb{87.14}$_{\pm.00}$ \\
			& CiteSeer & 36.99$_{\pm2.7}$ & 31.31$_{\pm.00}$ & 62.30$_{\pm.02}$ & 46.32$_{\pm.00}$ & 22.22$_{\pm.00}$ & 59.87$_{\pm.00}$ & 58.00$_{\pm3.2}$ & 41.51$_{\pm.98}$ &\tib{65.95}$_{\pm.01}$ & \tbb{66.58}$_{\pm.04}$ \\
			& ALOI & 18.40$_{\pm.36}$ & 60.25$_{\pm.00}$ & 78.49$_{\pm1.2}$ & 60.39$_{\pm.00}$ & 44.98$_{\pm.00}$ & 78.44$_{\pm2.3}$ & 79.78$_{\pm.47}$ & 70.41$_{\pm1.1}$ &\tib{85.94}$_{\pm1.1}$ & \tbb{86.21}$_{\pm.00}$ \\
			& Flower17 & 54.56$_{\pm1.2}$ & 47.87$_{\pm.00}$ & 59.85$_{\pm1.4}$ & 47.72$_{\pm.00}$ & 26.47$_{\pm.00}$ & 56.36$_{\pm1.3}$ & 59.73$_{\pm1.2}$ & 47.96$_{\pm.60}$ &\tbb{69.88}$_{\pm1.4}$ & \tib{69.81}$_{\pm1.5}$ \\
			& Caltech101 & 40.11$_{\pm.39}$ & 42.79$_{\pm.00}$ & 45.83$_{\pm.44}$ & 43.12$_{\pm.00}$ & 21.46$_{\pm.00}$ & 38.98$_{\pm.34}$ & 45.94$_{\pm.37}$ & \textbbf{48.34}$_{\pm.18}$ &\tib{45.97}$_{\pm.47}$ & {45.96}$_{\pm.39}$ \\
			& UCI Digits & 87.00$_{\pm.00}$ & 96.85$_{\pm.00}$ & 97.65$_{\pm.00}$ & 84.75$_{\pm.00}$ & 88.10$_{\pm.00}$ & 81.41$_{\pm1.4}$ & 87.02$_{\pm.04}$ & 96.59$_{\pm.03}$ &\tib{98.10}$_{\pm.00}$ & \tbb{98.20}$_{\pm.00}$ \\
			\bottomrule
	\end{tabular}}
	\label{tab:result}%
\end{table*}

The normalized mutual information (NMI)\cite{strehl2002cluster}, adjusted rand index (ARI)\cite{vinh2010information}, clustering accuracy (ACC)\cite{Zhan2018} and purity\cite{Zhan2018} are used to measure the clustering performance.

\subsection{Experimental Setup}
We downloaded the source code of {AWP, AASC, MVGL} and {MCGC} from the authors' websites, and implement other algorithms following the instruction in the original papers. 
We conduct all experiments with \textsc{Matlab} R2019b on a machine with an Intel Core i9-9960X 16-core CPU and 128GB RAM. 
For the algorithms which use the Laplacian matrices of graphs, we adopt the symmetrically normalized Laplacian $\mathbf{L}_{sym}=\mathbf{I}-\mathbf{D}^{\nicefrac{-1}{2}}\mathbf{S}\mathbf{D}^{\nicefrac{-1}{2}}$, where $\mathbf{S}$ and $\mathbf{D}$ are the adjacency matrix and degree matrix of the graph, respectively \cite{ng2002spectral, von2007tutorial}. 
We fix the number of the nearest neighbors in $k$NN to 6 in all experiments.  
For the algorithms with parameter(s), which include {CoReg} (1 parameter), {RMSC} (1 parameter), {MCGC} (1 parameter), {WMSC} (2 parameters), DGF and SGF (2 parameters), we use grid search to test the parameter(s) of these algorithms on the grids $\{10^{-5},10^{-4},\dots,10^5\}^m$, where $m$ is the number of parameters of the algorithm, and we report the scores with the best parameter(s) (i.e., the parameter(s) achieving the highest NMI) found on each dataset. Note that the grids contain values that are very close to the parameters suggested by the authors. Thus, all algorithms should exhibit their best performance in the experiments. 
We set the weights $\lambda_{i}$ for each view to $1$ in DGF and SGF, without considering the importance of different views.

We run all algorithms 10 times and report the average scores and standard deviation. If $k$-means clustering is used in any algorithm, we run $k$-means 10 times and set its maximum number of iterations to 1000 to reduce the effect of random initialization. %
We use cosine distance to construct distance matrices for the text datasets Reuters, Reuters-21578, BBCSport and CiteSeer, and use Euclidean distance for other datasets. %
All distance matrices are transformed to similarity matrices with the Gaussian kernel. The parameter $\rho$ in Gaussian kernel is set to the mean value of the edges in the $k$NN distance graph. %

\subsection{Clustering Result}
The clustering performance is shown in Table~\ref{tab:result}. We can see that multi-view clustering methods generally achieve better performance than single-view clustering. The two proposed graph learning methods, DGF and SGF, achieve better performance than other state-of-the-art methods on most datasets in the experiments, which demonstrates the effectiveness and robustness of the proposed algorithms. Our algorithms outperform RMSC, which is another multi-view clustering method that is robust to noise, as we model not only noise but also the multi-view inconsistency in our framework. Besides, the graph learning methods {AASC, MVGL} and {MCGC} do not outperform our methods probably because they neglect the inconsistency across views in graph learning. 
We also note that DGF achieves better scores than SGF on a few datasets, which may result from the hypothesis in Section~\ref{sec:dgf} that fusion of distance matrices may better preserve the relationship between data points. 

It is noteworthy that, even without dataset-specific parameter tuning, the proposed algorithms can still achieve stable clustering results in the benchmark datasets as indicated by the parameter analysis in Section~\ref{sec:param}. Thereby, we fix the parameters of DGF and SGF by setting $\beta=1$ and $\gamma=10^4$ and run the two algorithms on the 12 datasets. \blue{The result is that DGF and SGF with fixed parameters still outperform the state-of-the-art multi-view clustering methods on many datasets, or achieve results that are at least comparable to the state-of-the-art. Please see the \app for the experimental results.} 

\subsection{Comparisons with the Preliminary Version}
\begin{figure*}[!t]%
	\centering
	\subfloat{\includegraphics[width=.166\linewidth]{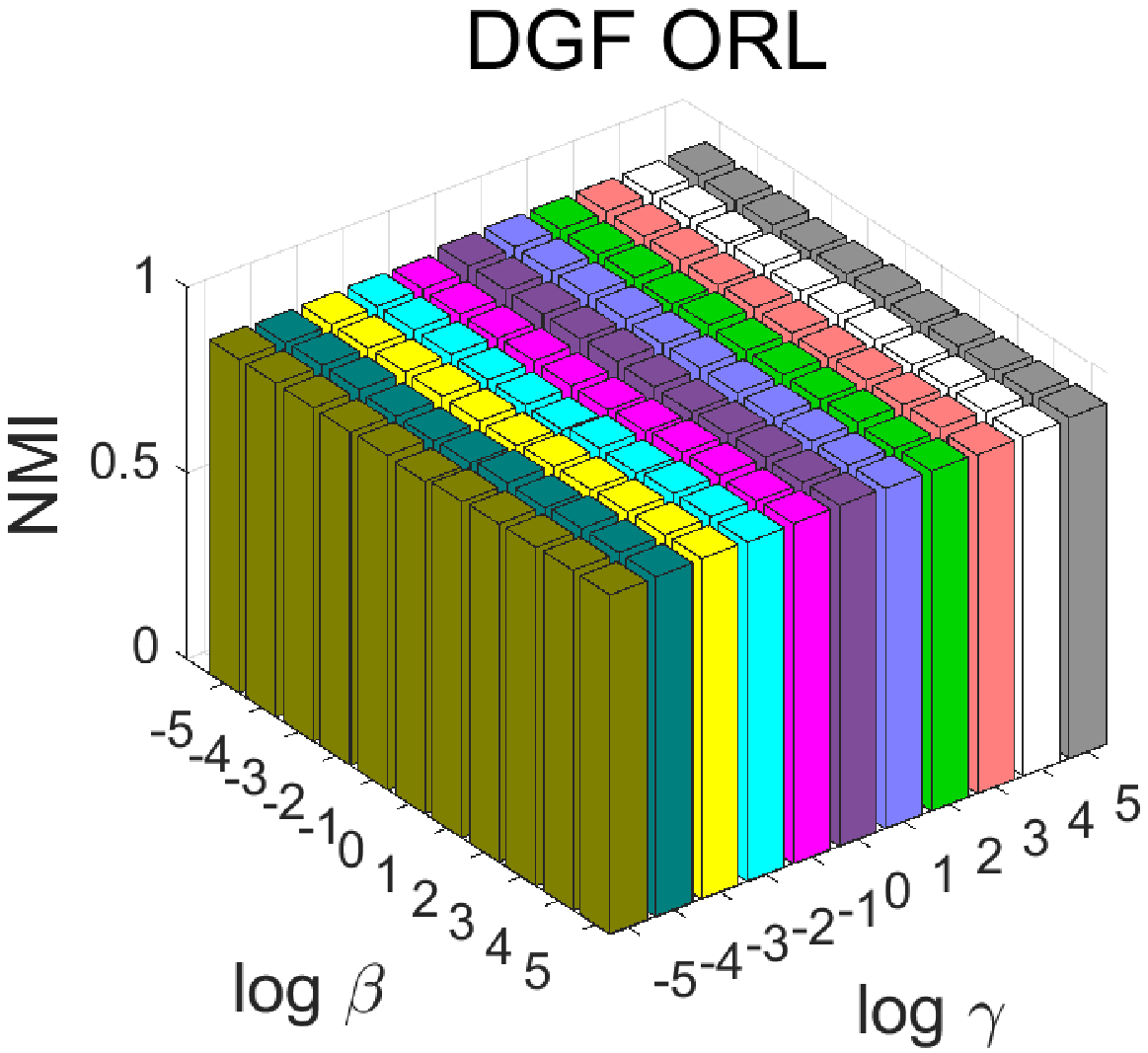}}\hfil
	\subfloat{\includegraphics[width=.166\linewidth]{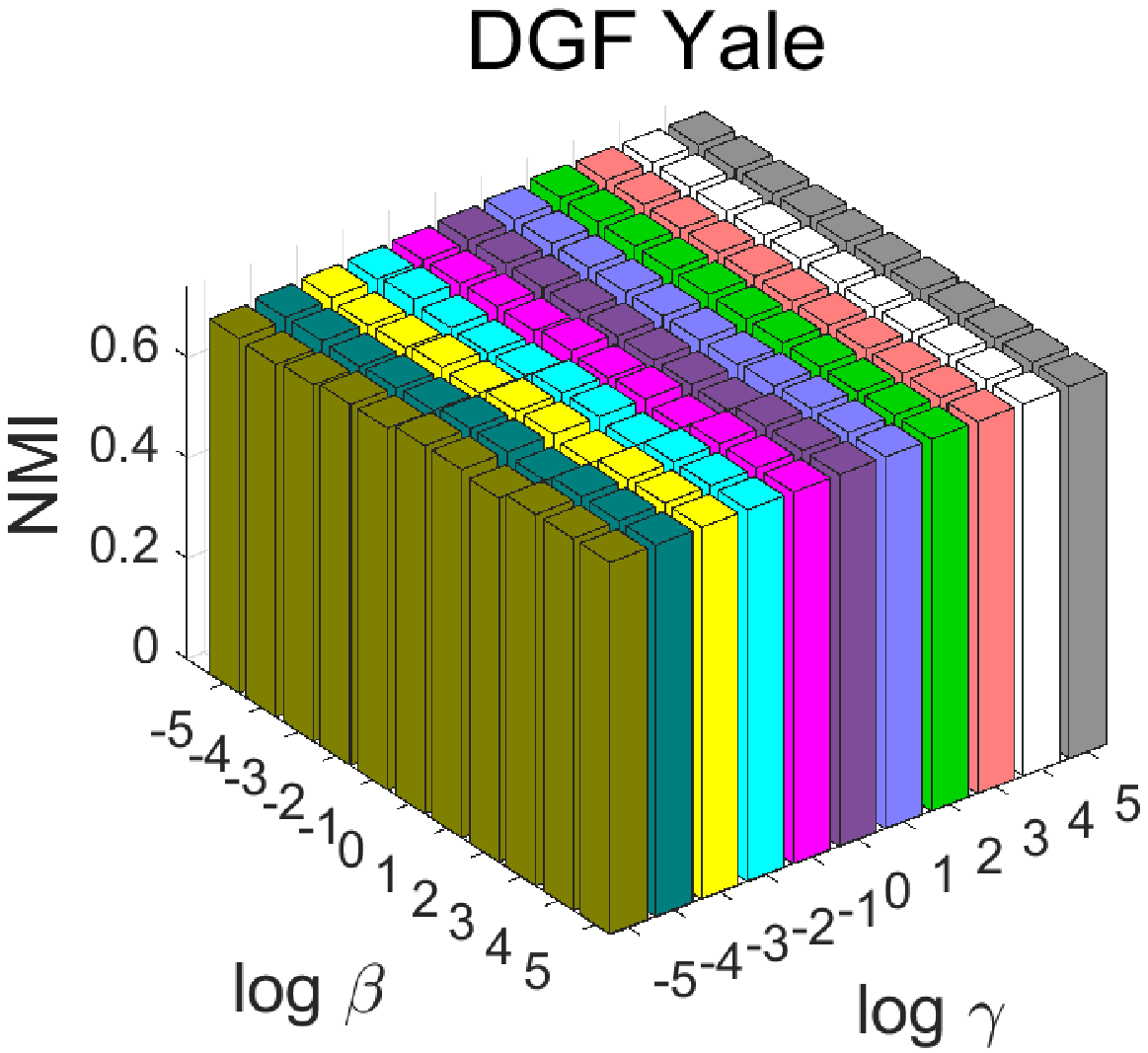}}\hfil
	\subfloat{\includegraphics[width=.166\linewidth]{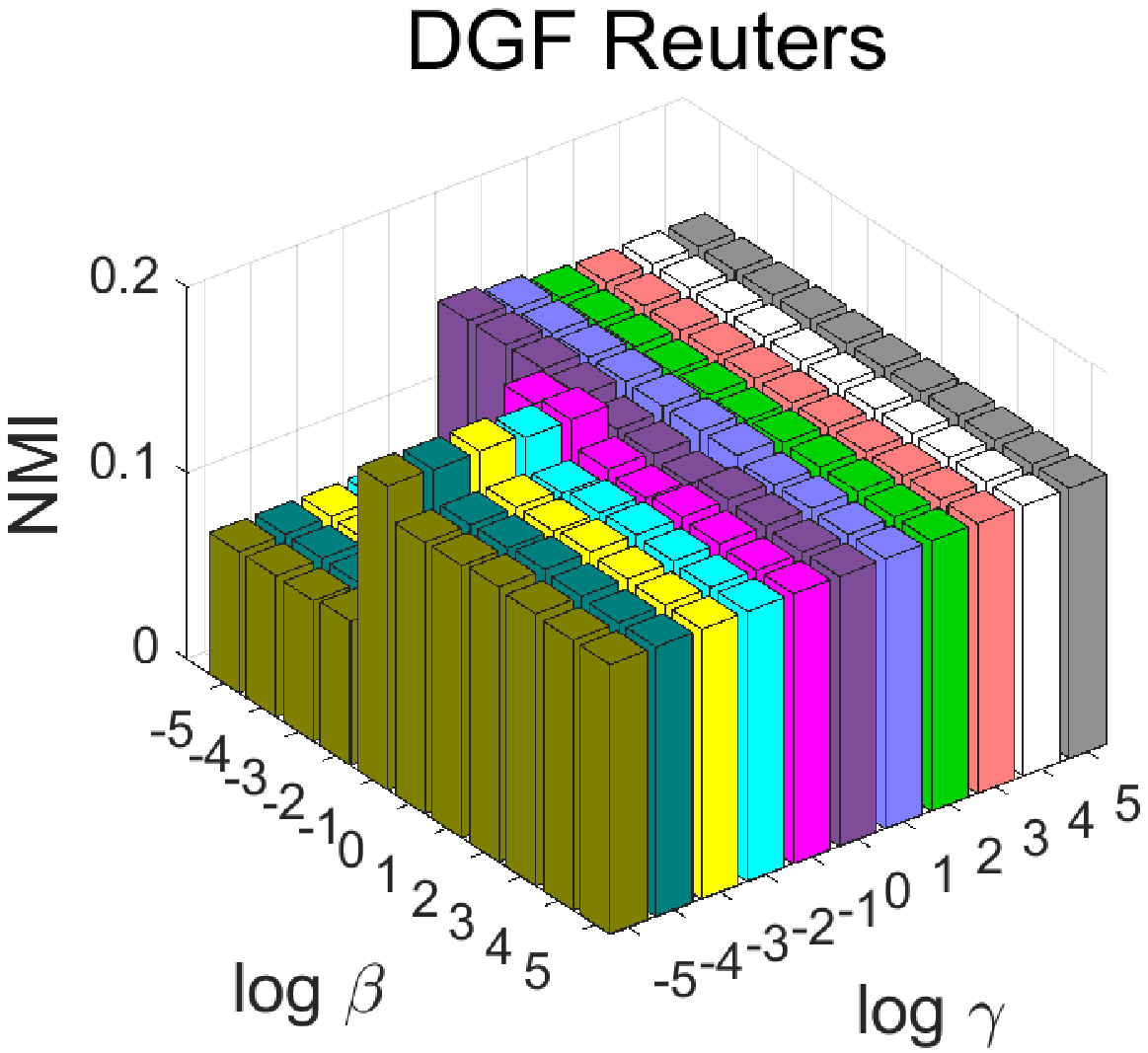}}\hfil
	\subfloat{\includegraphics[width=.166\linewidth]{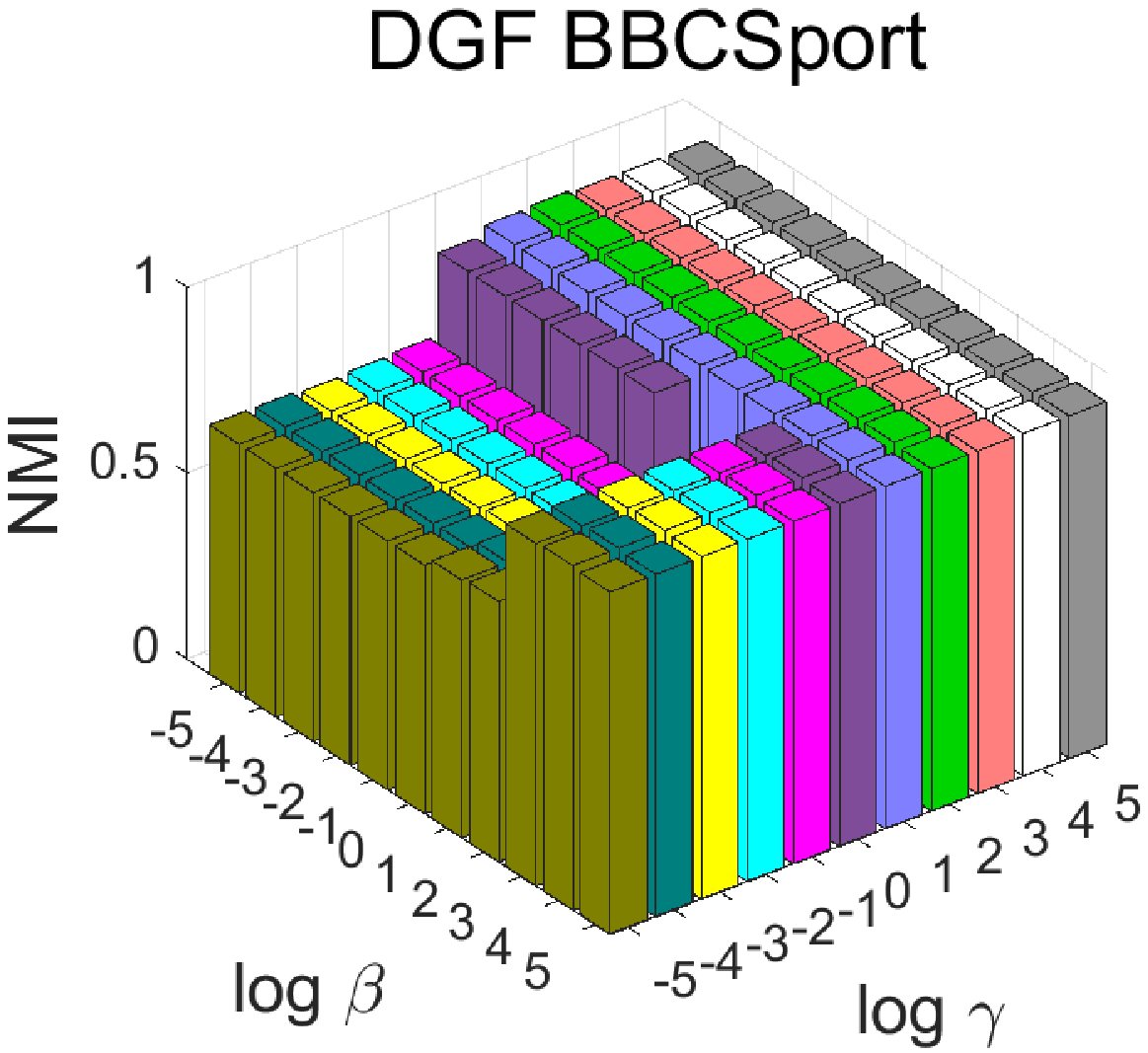}}\hfil
	\subfloat{\includegraphics[width=.166\linewidth]{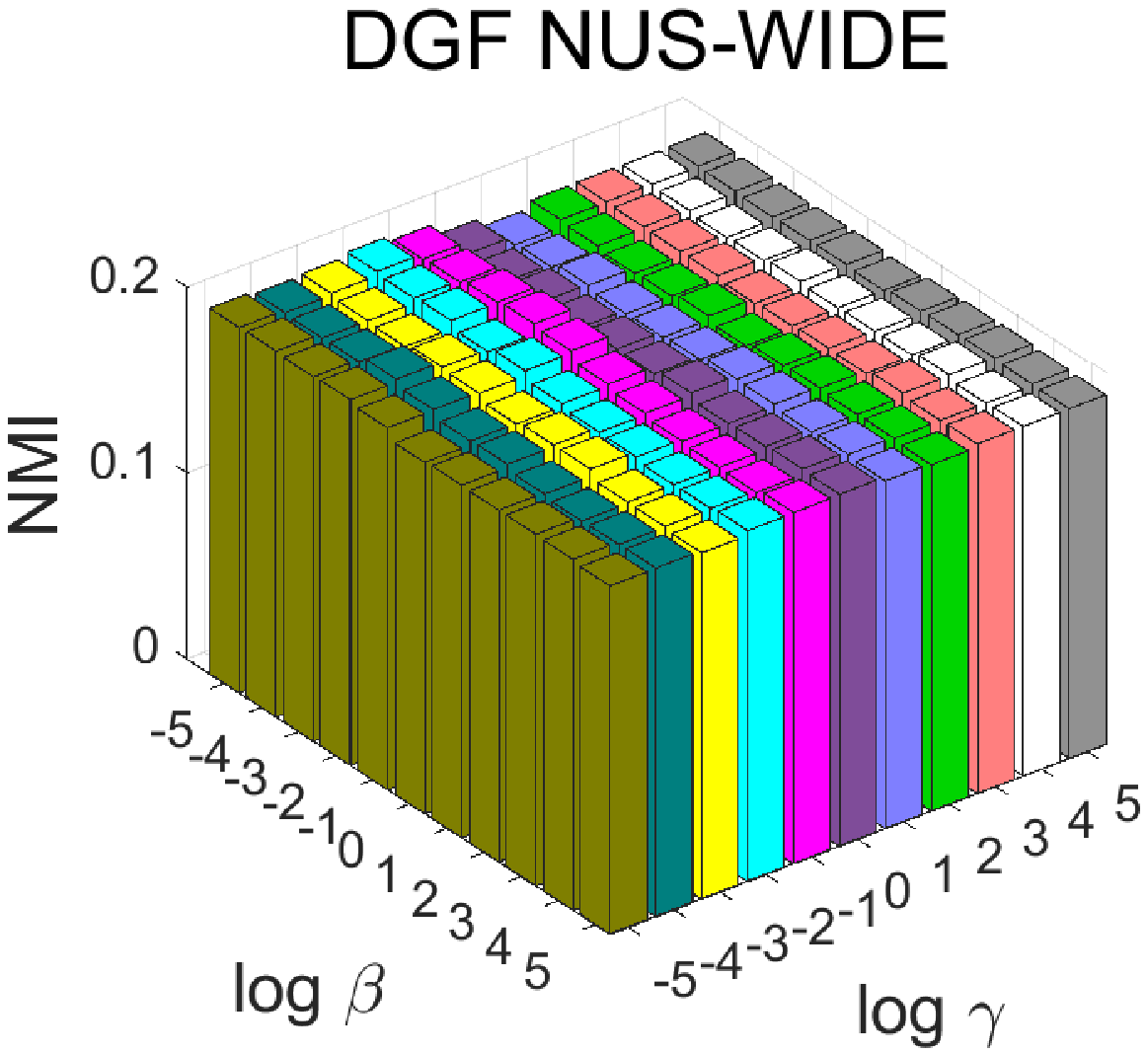}}\hfil
	\subfloat{\includegraphics[width=.166\linewidth]{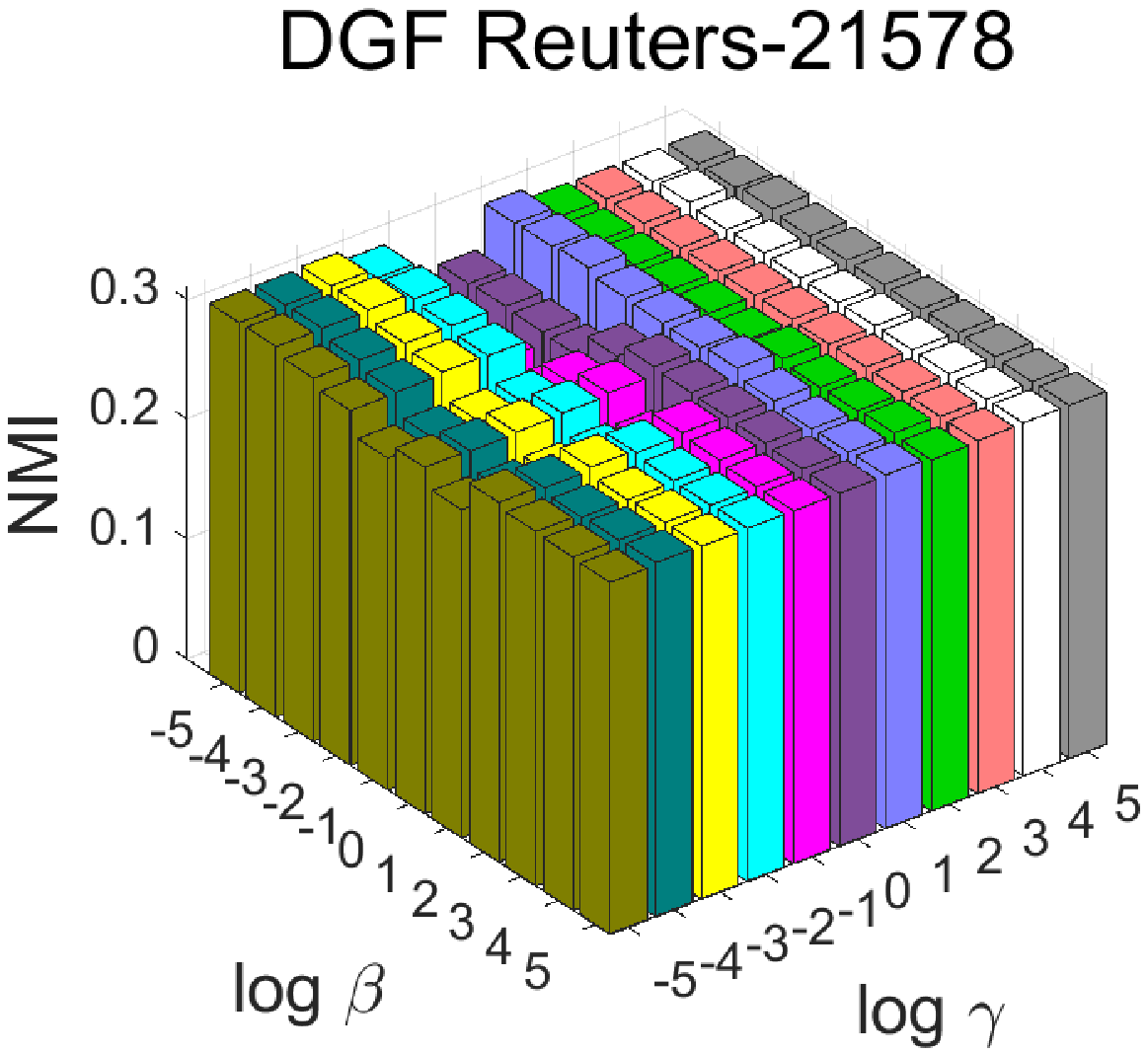}}\\
	\subfloat{\includegraphics[width=.166\linewidth]{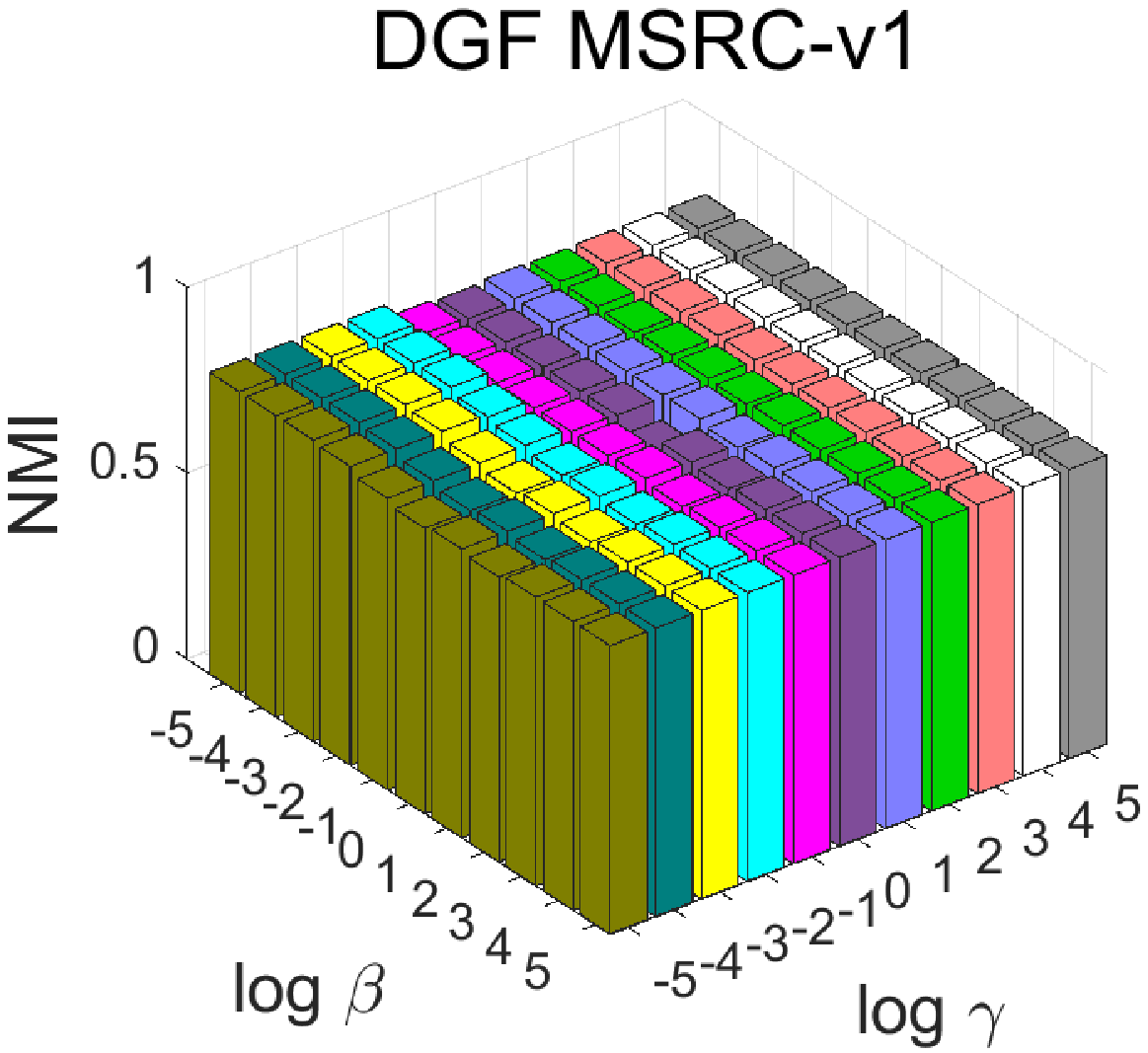}}\hfil
	\subfloat{\includegraphics[width=.166\linewidth]{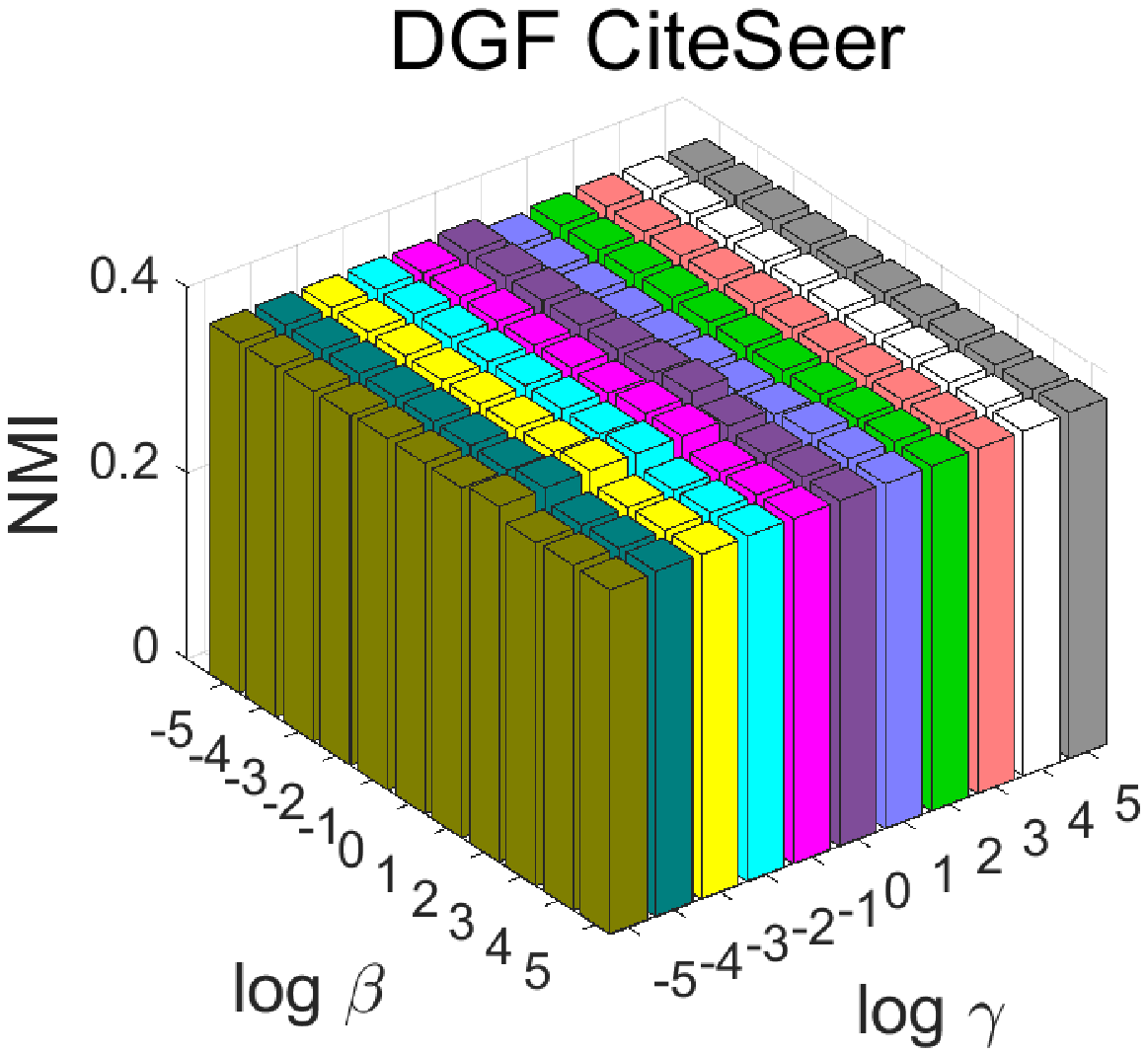}}\hfil
	\subfloat{\includegraphics[width=.166\linewidth]{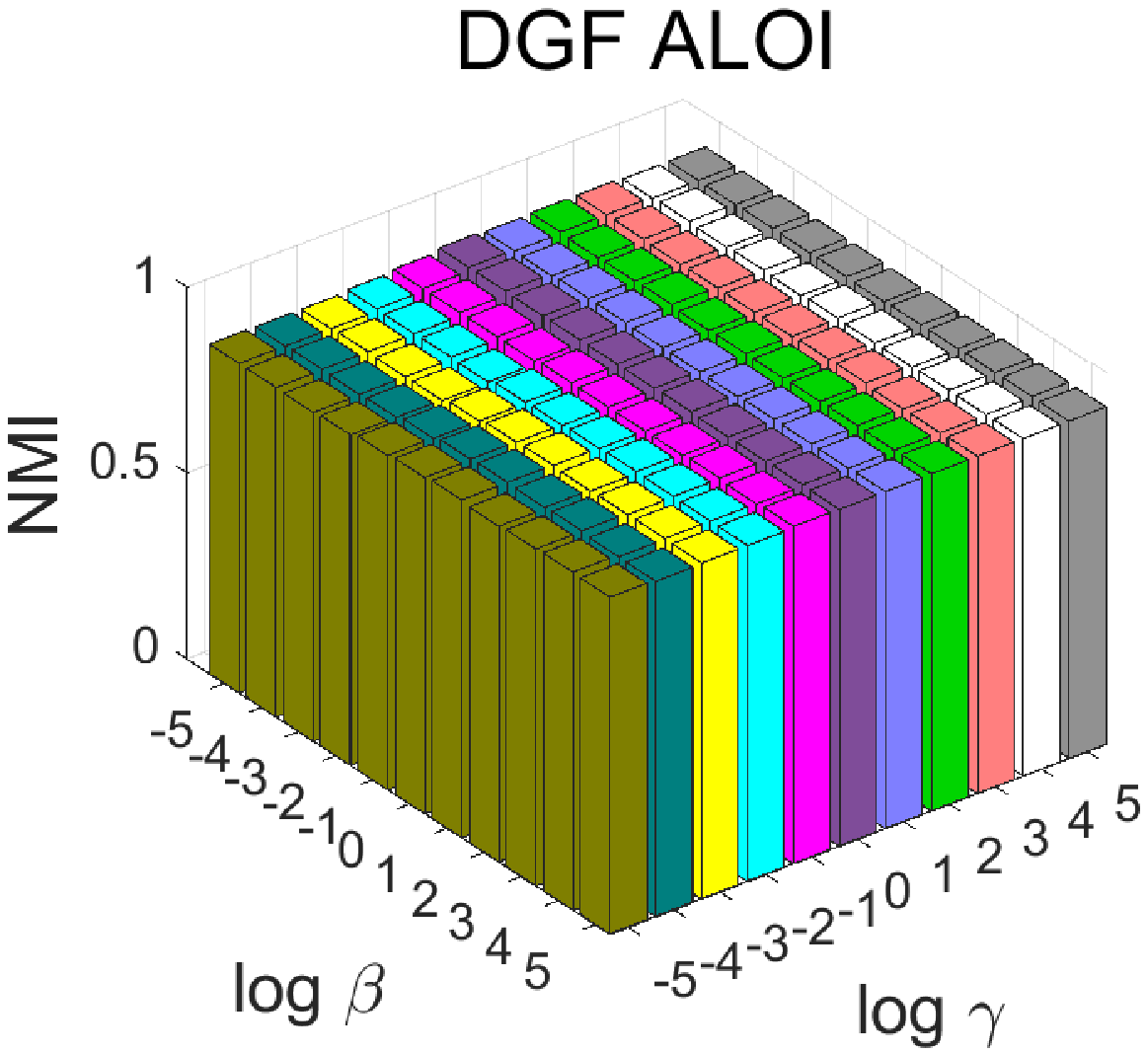}}\hfil
	\subfloat{\includegraphics[width=.166\linewidth]{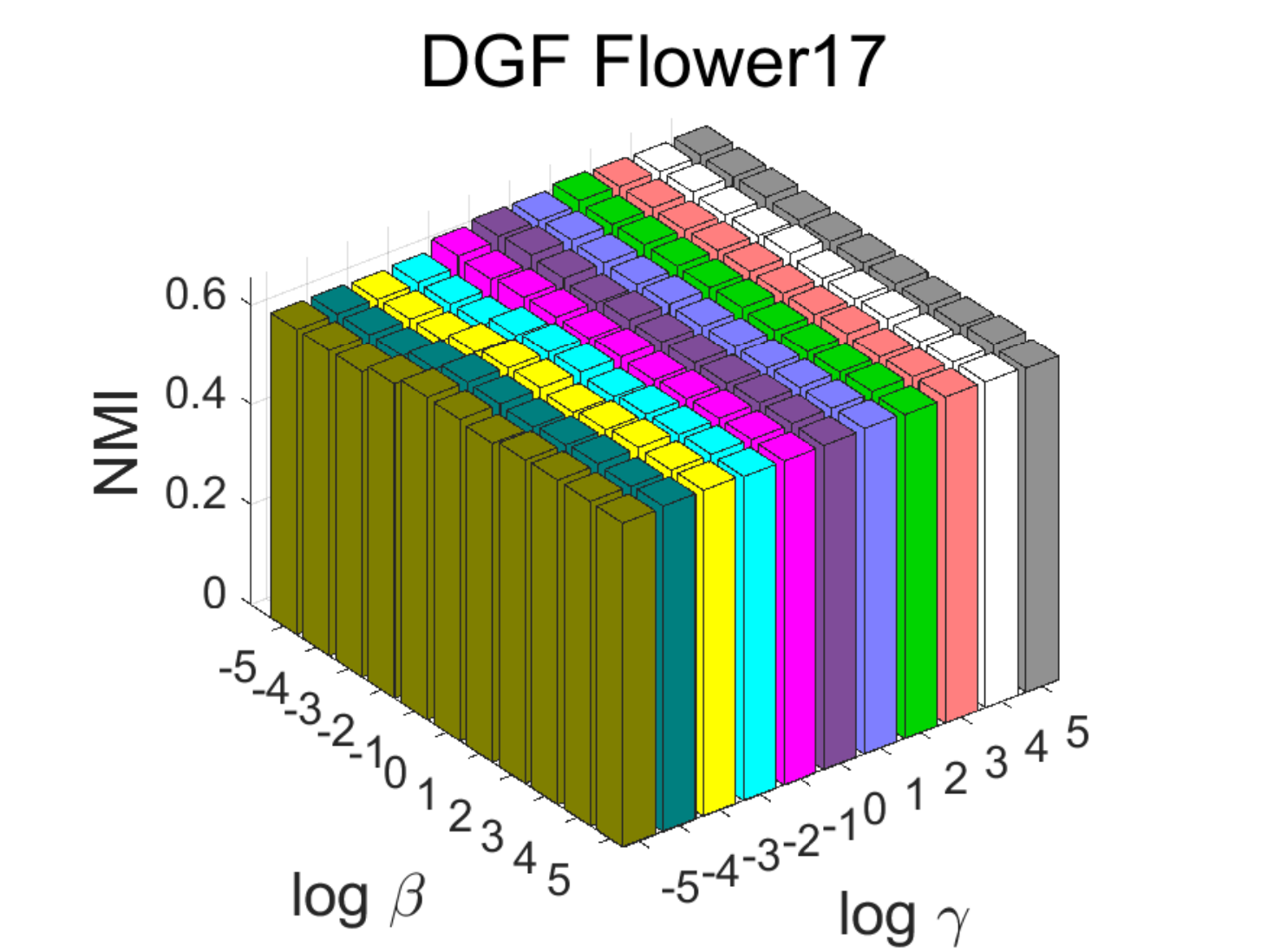}}\hfil
	\subfloat{\includegraphics[width=.166\linewidth]{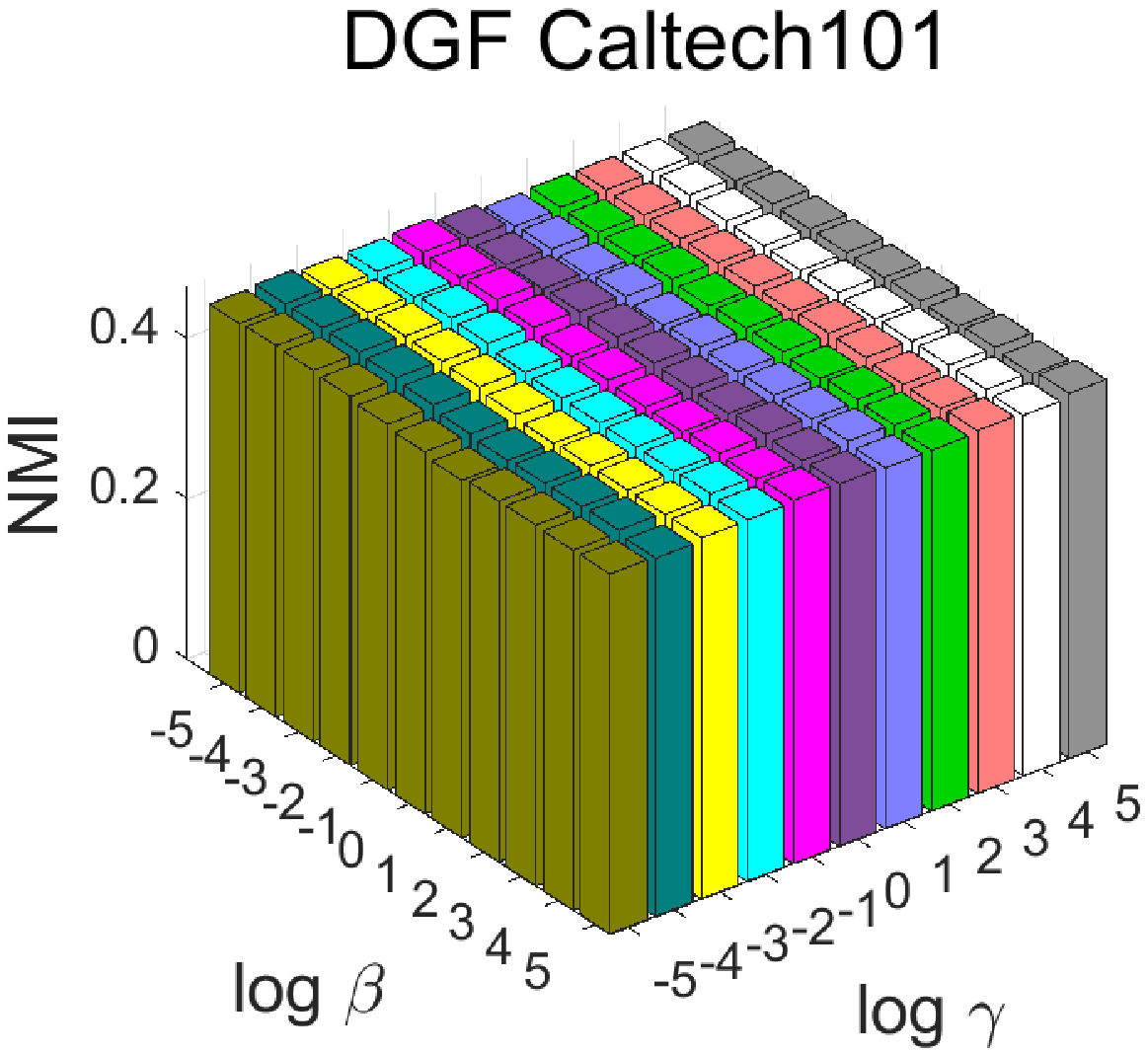}}\hfil
	\subfloat{\includegraphics[width=.166\linewidth]{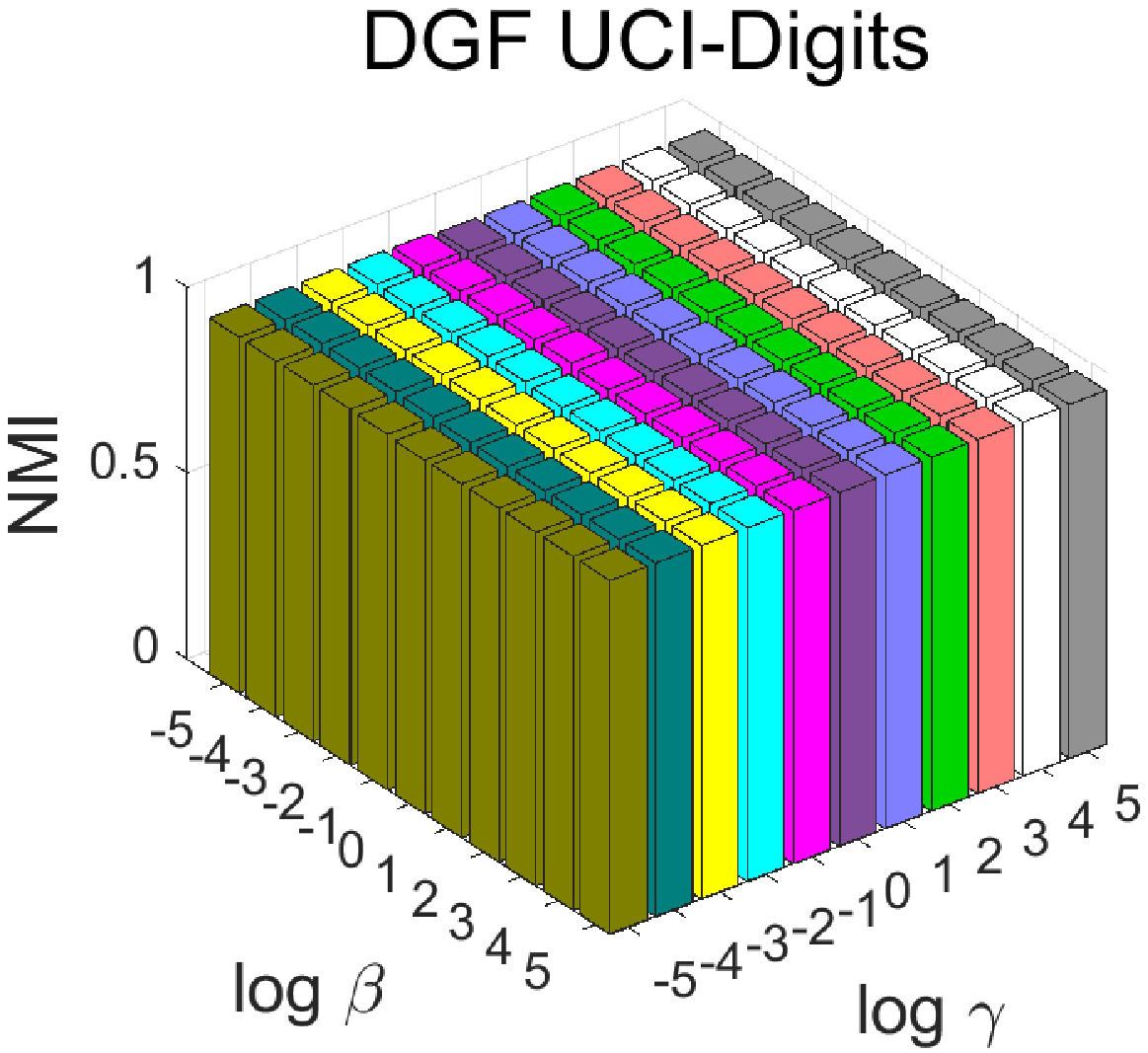}}\\
	\subfloat{\includegraphics[width=.166\linewidth]{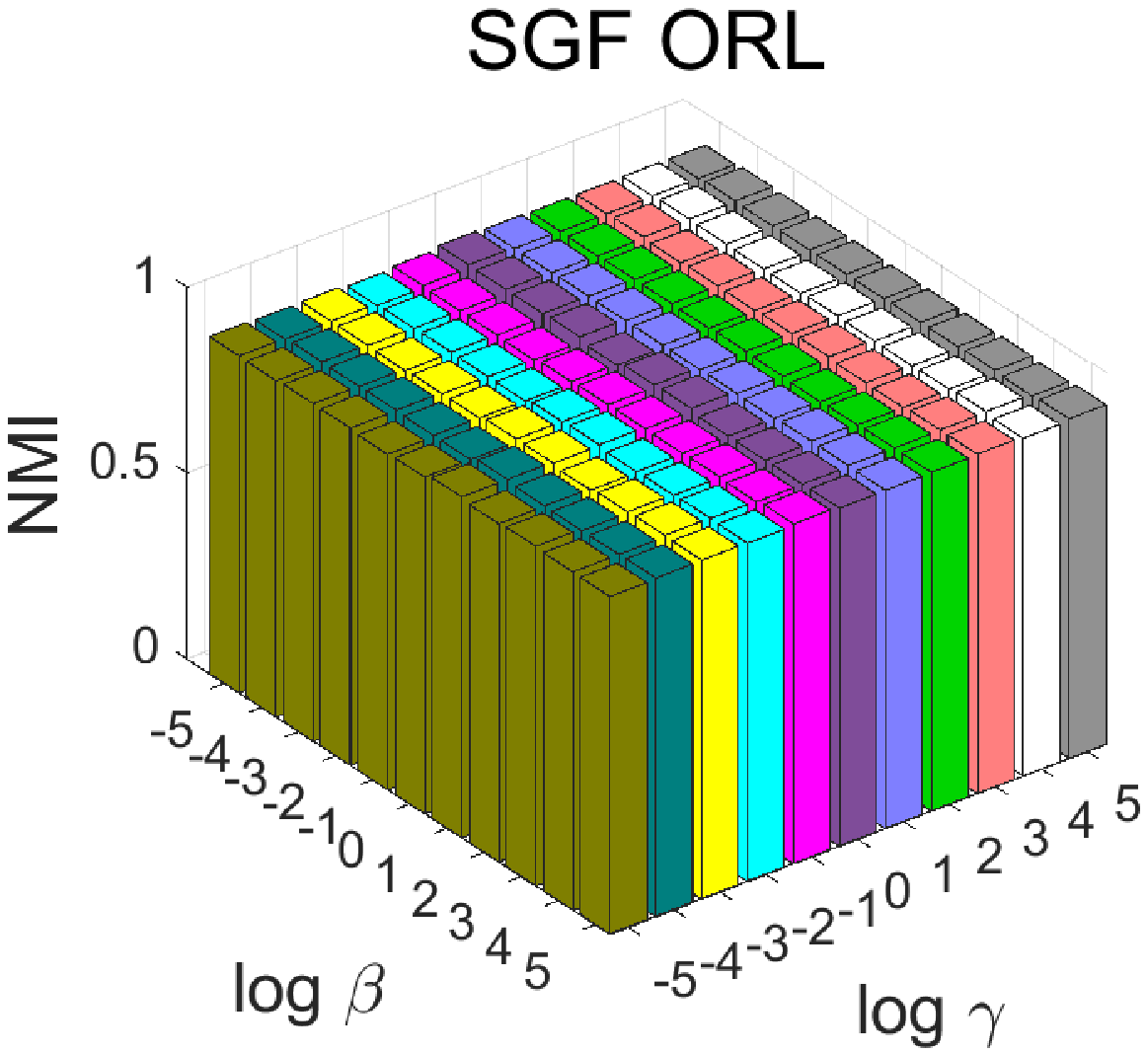}}\hfil
	\subfloat{\includegraphics[width=.166\linewidth]{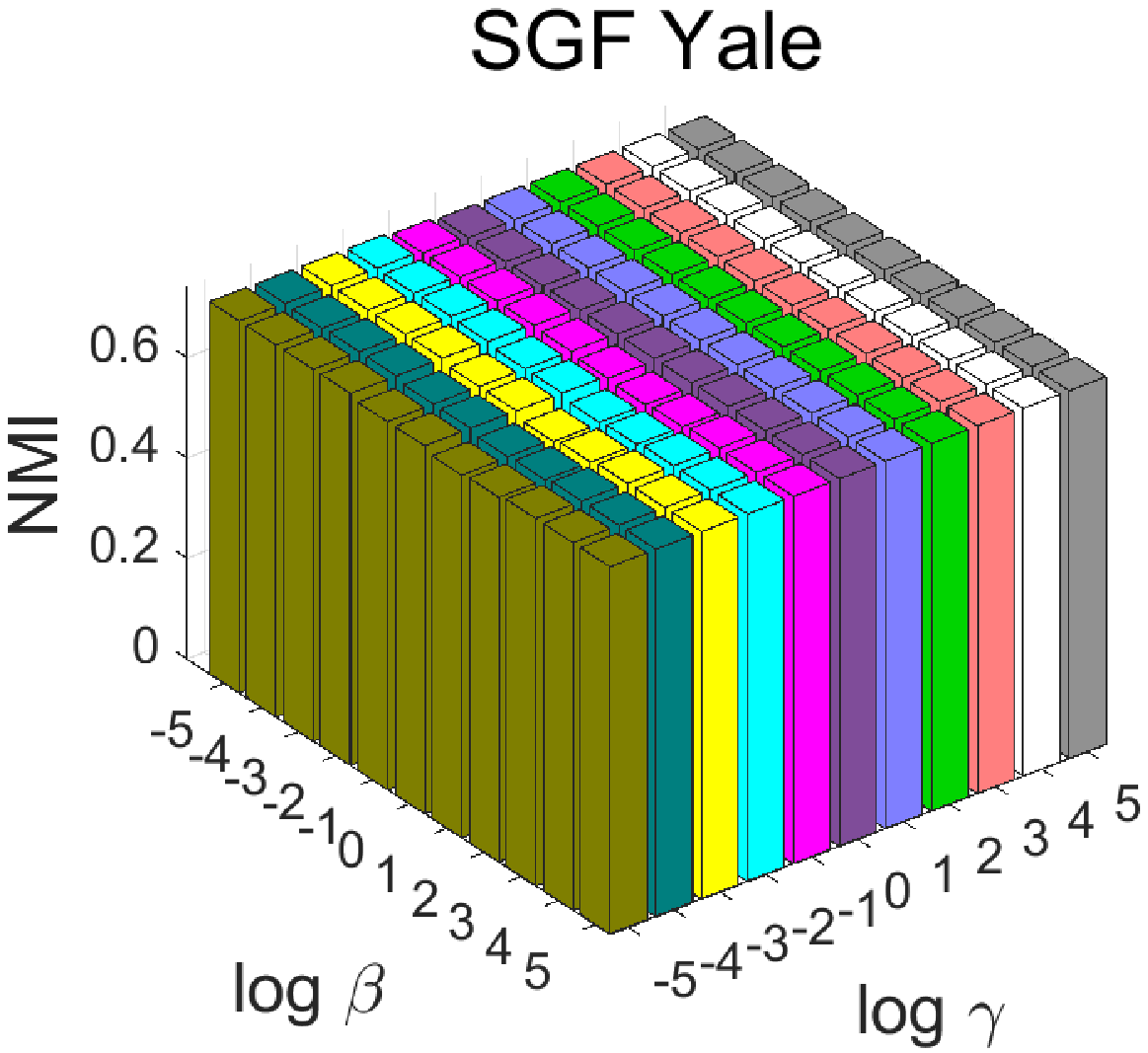}}\hfil
	\subfloat{\includegraphics[width=.166\linewidth]{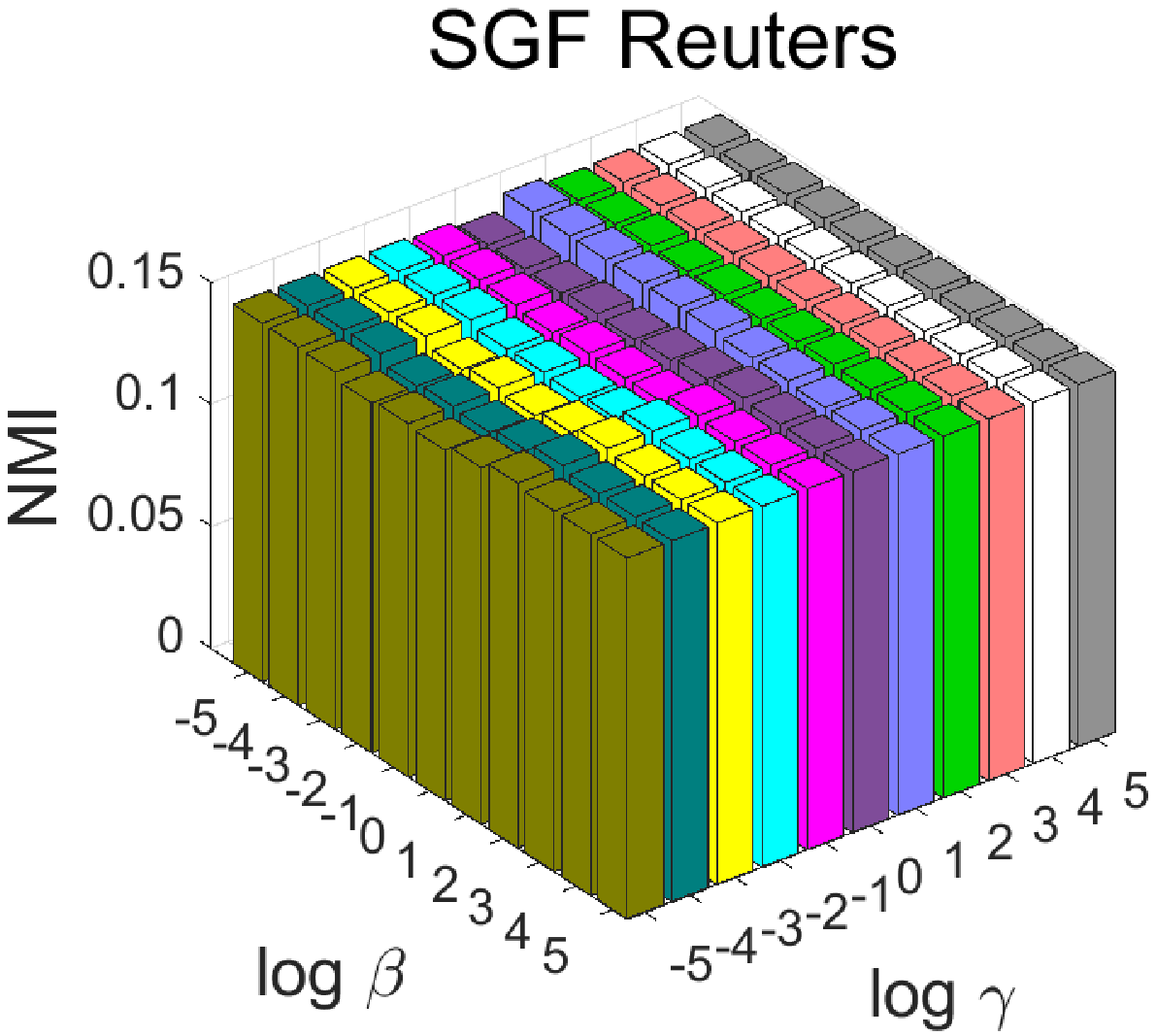}}\hfil
	\subfloat{\includegraphics[width=.166\linewidth]{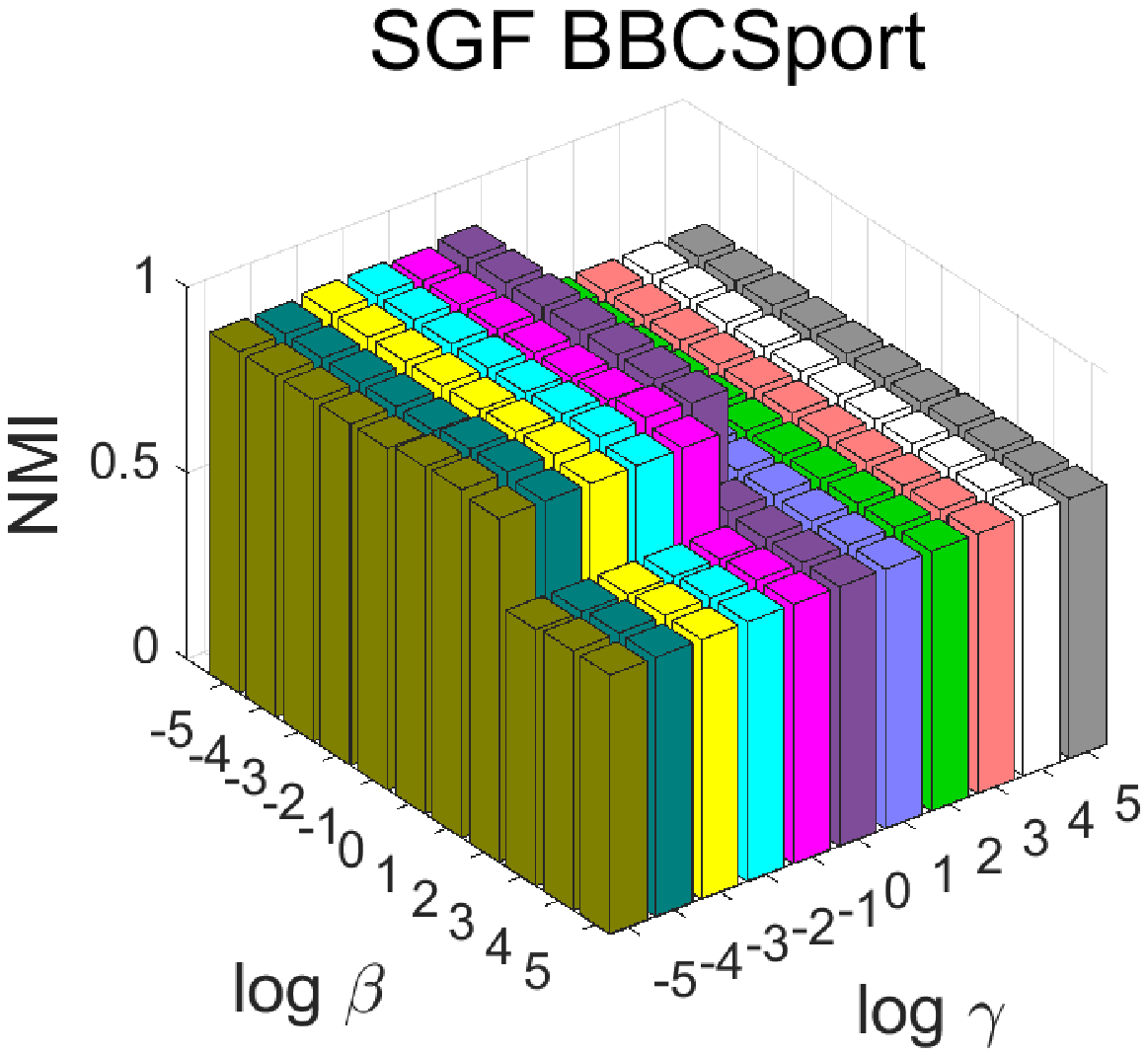}}\hfil
	\subfloat{\includegraphics[width=.166\linewidth]{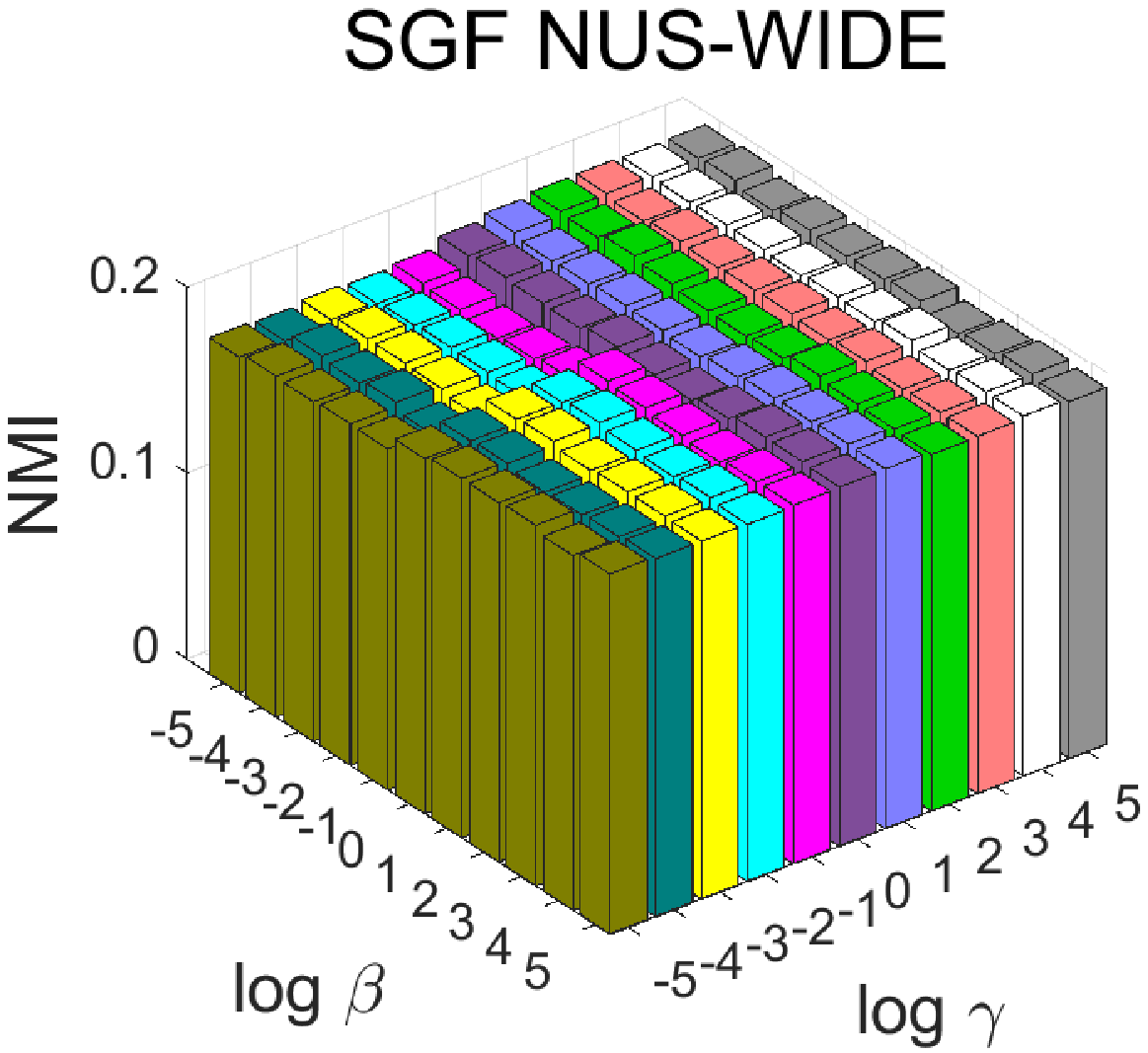}}\hfil
	\subfloat{\includegraphics[width=.166\linewidth]{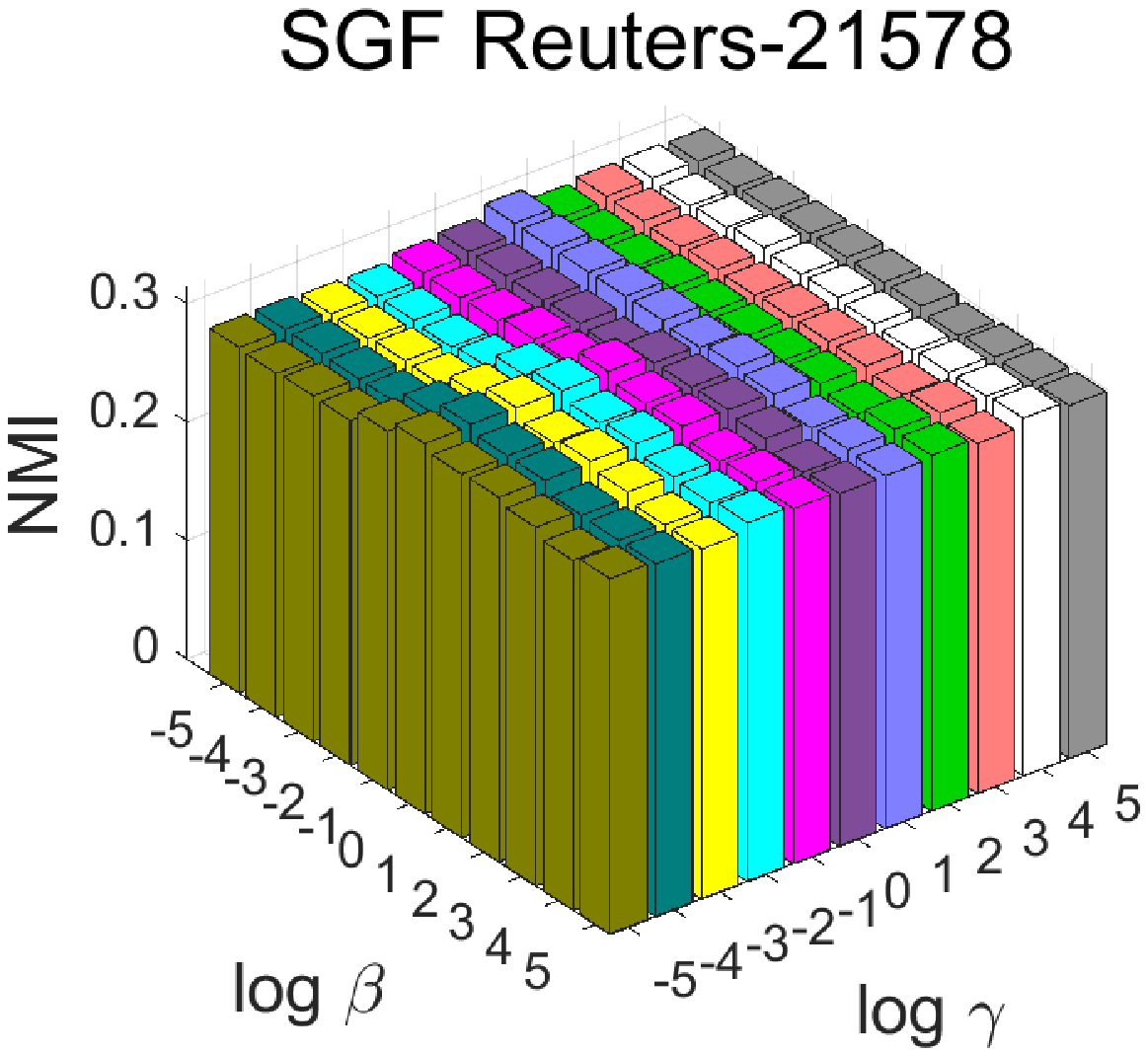}}\\
	\subfloat{\includegraphics[width=.166\linewidth]{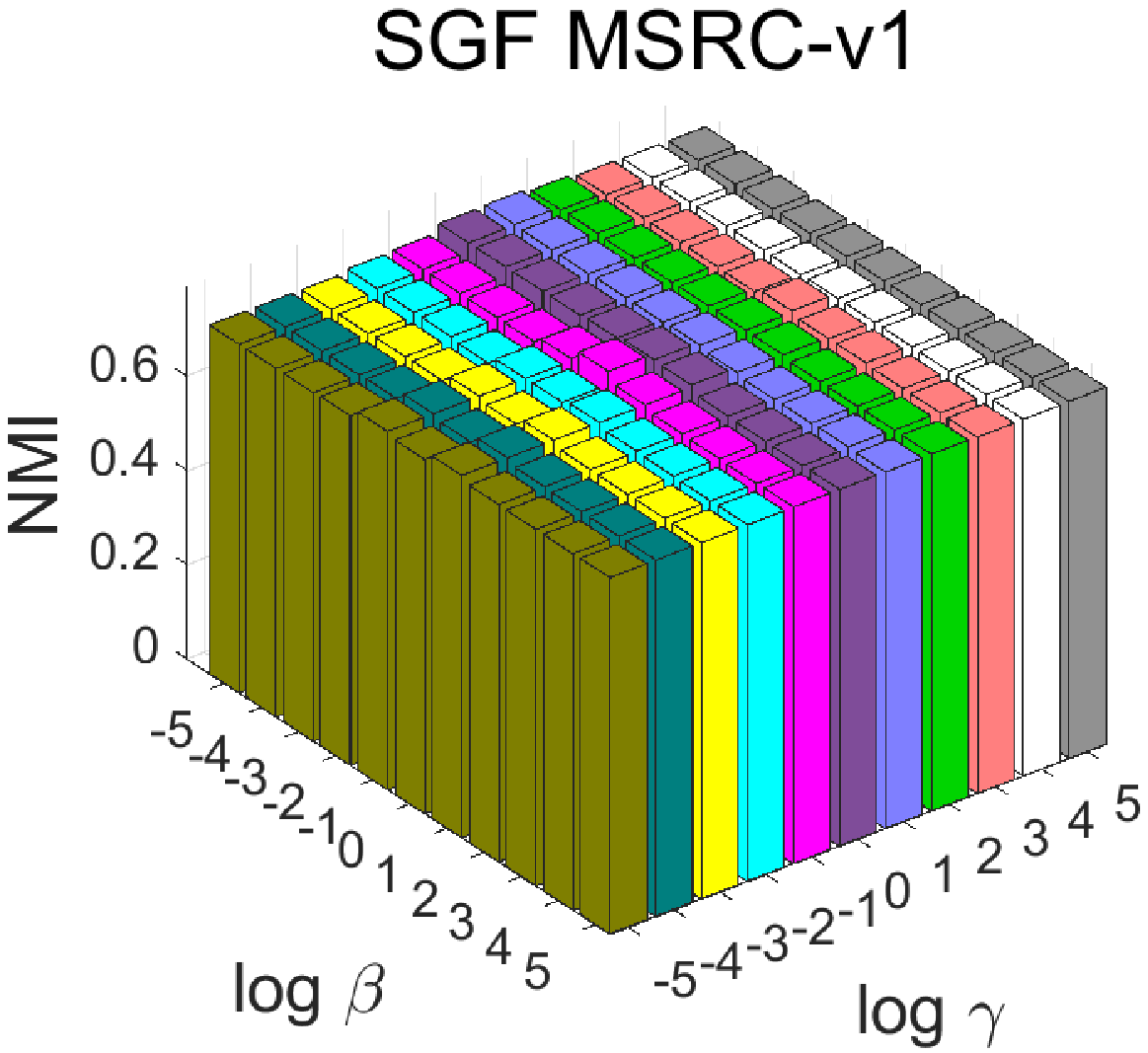}}\hfil
	\subfloat{\includegraphics[width=.166\linewidth]{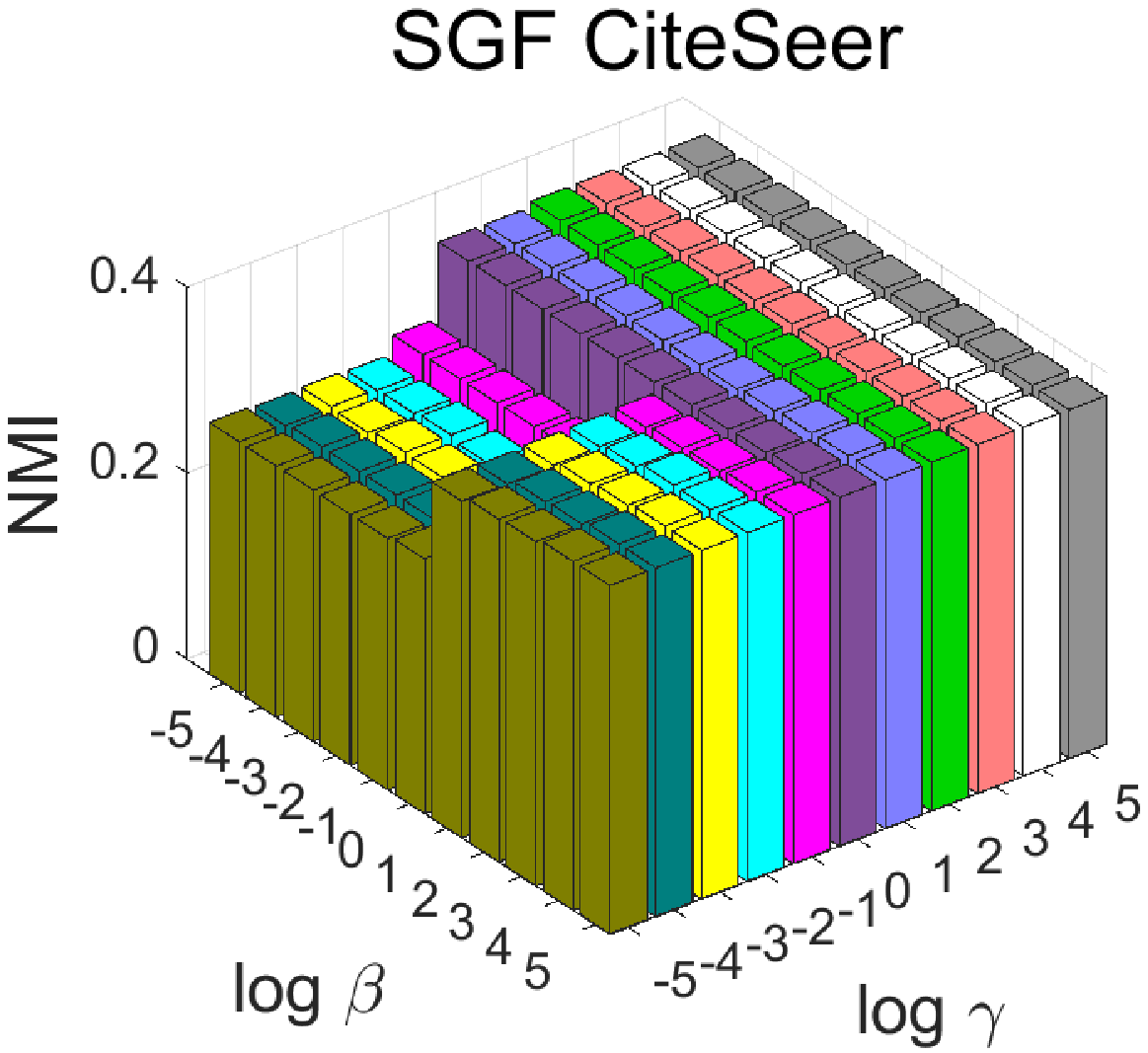}}\hfil
	\subfloat{\includegraphics[width=.166\linewidth]{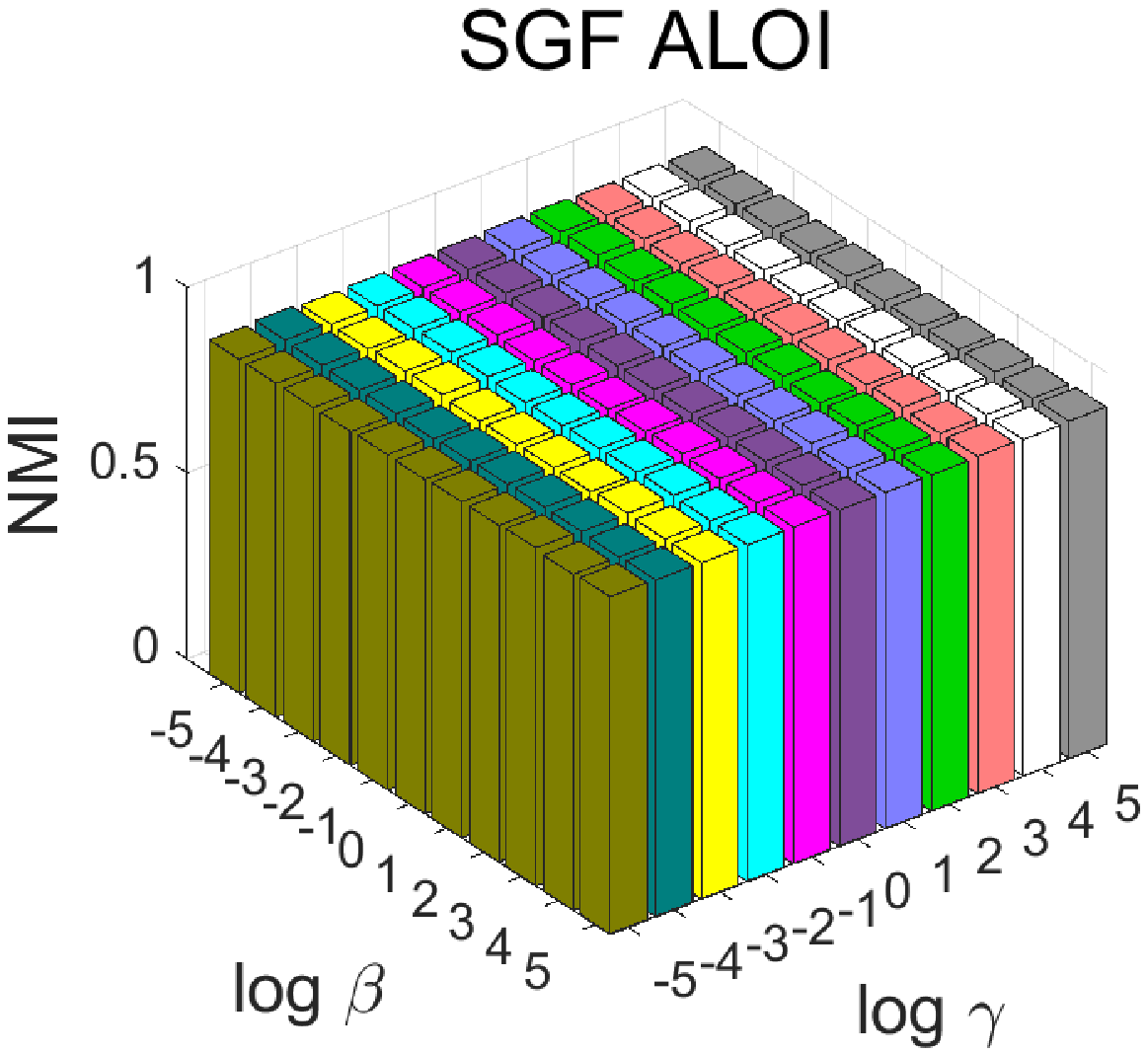}}\hfil
	\subfloat{\includegraphics[width=.166\linewidth]{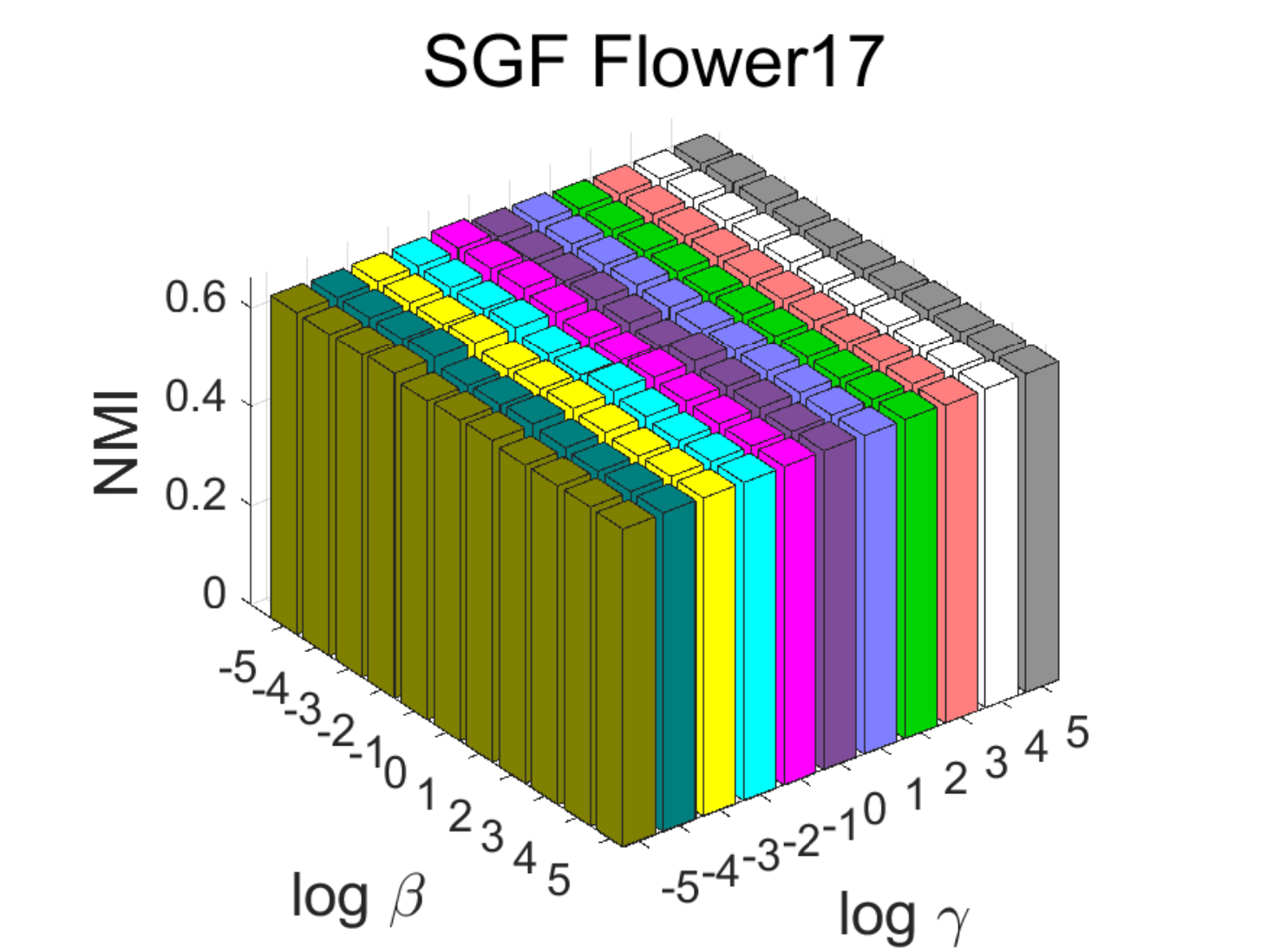}}\hfil
	\subfloat{\includegraphics[width=.166\linewidth]{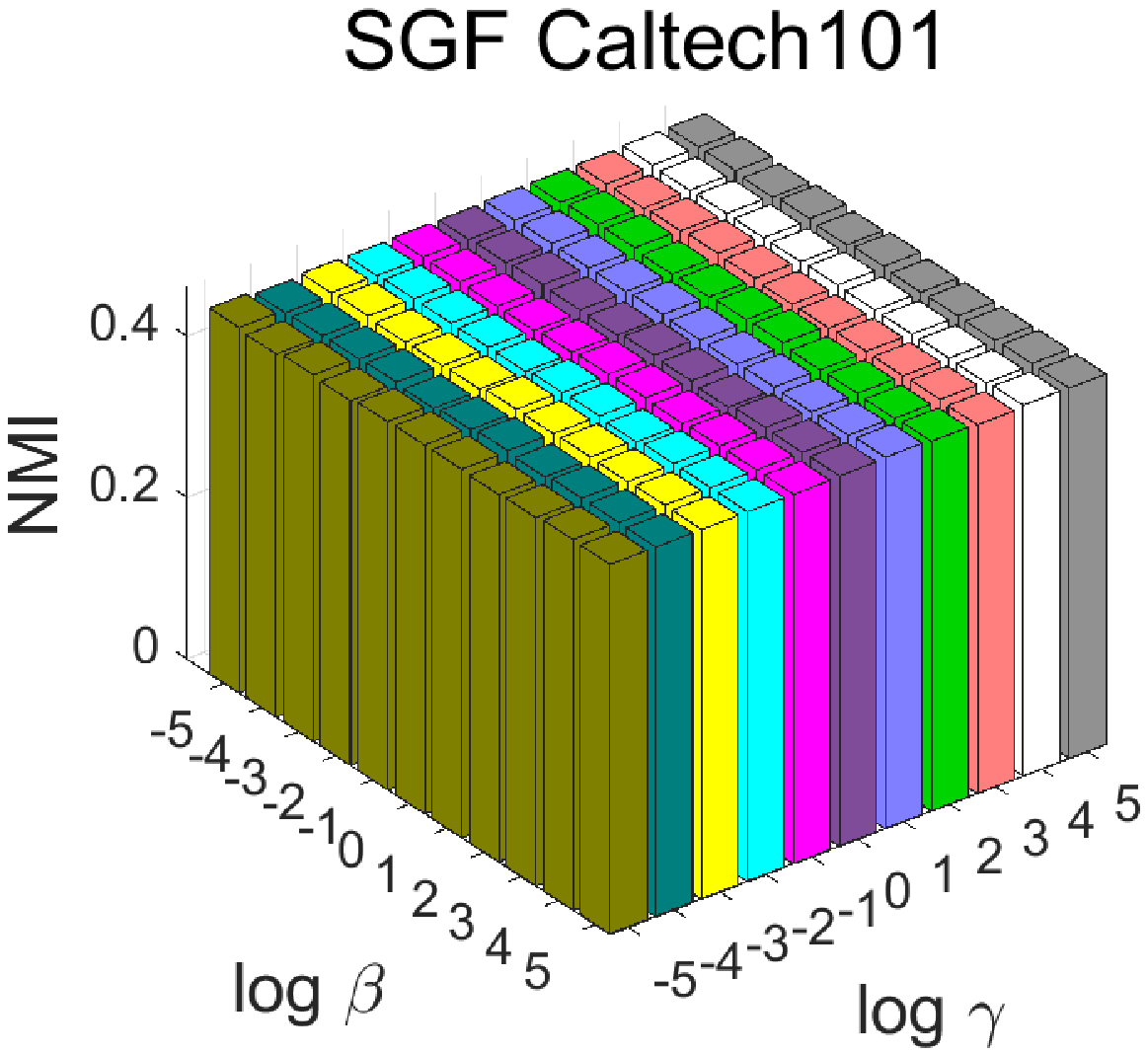}}\hfil
	\subfloat{\includegraphics[width=.166\linewidth]{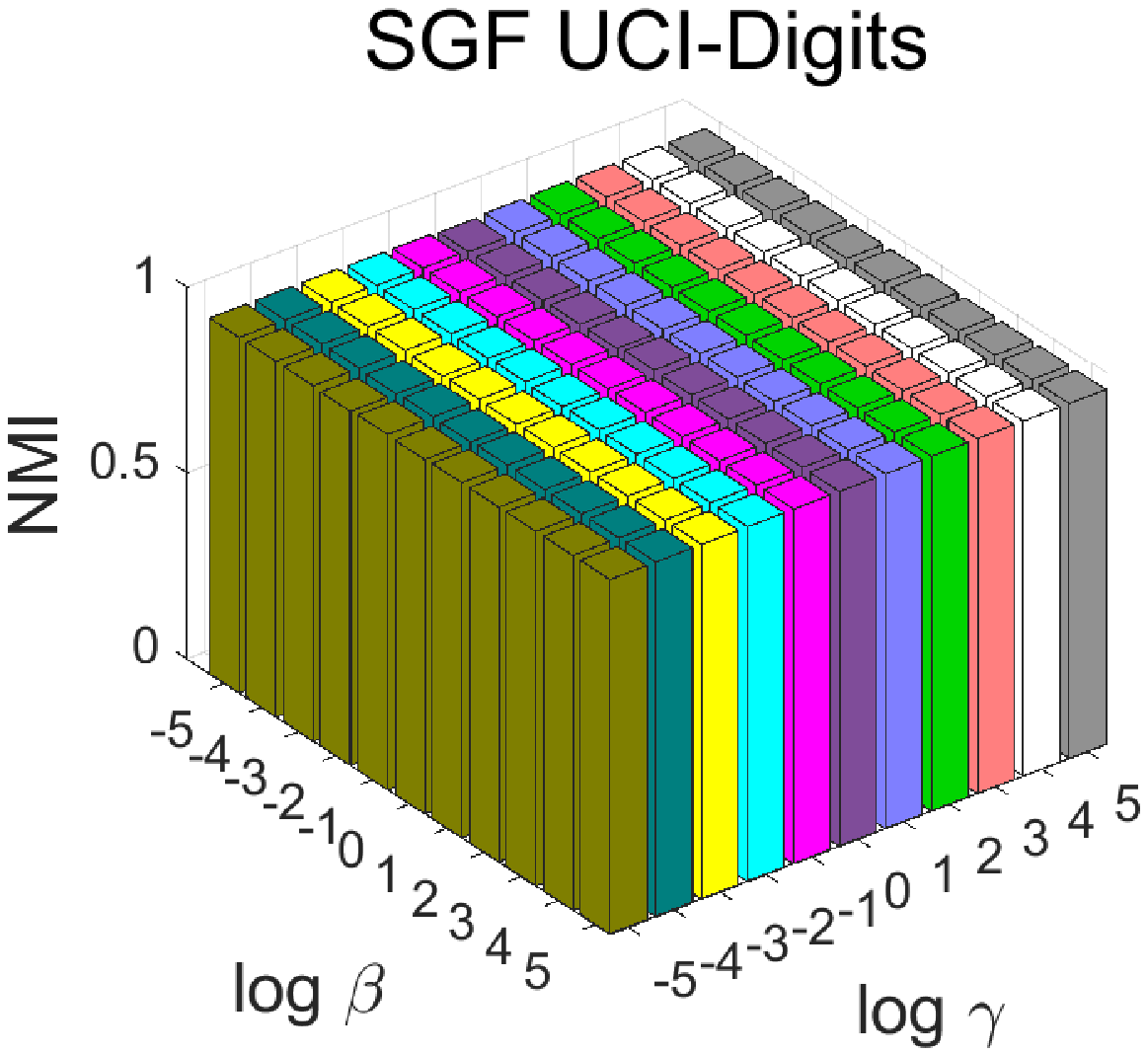}}
	\caption{(Best viewed/printed in color.) NMI against parameters $\beta$ (axis on the left) and $\gamma$  (axis on the right) of DGF and SGF on the 12 datasets. }%
	\label{fig:param}
\end{figure*}

\blue{We also compare the clustering results of the revised DGF and SGF algorithms against the DGF and SGF in the preliminary versions (denoted by DGF-0 and SGF-0) \cite{liang2019consistency}. As shown in Section~A.4 and Table~\tabresult, the revised algorithms obtain higher clustering scores than the preliminary ones on most tested datasets, sometimes surpassing the preliminary ones by a large margin. Moreover, the revised algorithms are more stable than the preliminary ones, performing reasonably well across a wide range of hyper-parameters (further details are provided in Fig~\ref{fig:param}), while the preliminary algorithms performs less robustly on some datasets or for certain hyper-parameters.}

\subsection{Empirical Convergence Analysis} 
We propose an alternating minimization scheme to solve the optimization problem in the proposed graph learning framework by dividing it into 3 subproblems. 
\blue{As we have analyzed in Section~\ref{sec:convergence}, alternatively solving the 3 subproblems can converge to a local minimizer of the overall optimization problem~\eqref{func:obj3} in most cases. Empirically, we find that the proposed optimization approach always converges in all experiments with different hyper-parameters and on various datasets, mostly within a few iterations, which demonstrates its reliability and efficiency.} The convergence curves are shown in Figure~\figconv.

\subsection{Empirical Complexity Analysis}
\blue{
In Section~\ref{sec:complexity}, we give a theoretical result that the time complexity of the graph learning algorithm can be reduced to $O(n_e v^2)$ or $O(k n v^2)$, where $n_e$, $n$, $v$, and $k$ are respectively the number of edges, nodes, views, and nearest neighbors in $k$NN. 
For each of the three factors $k, n, v$ in the complexity, we fix the other two factors and vary only the one being investigated in the experiments. We plot the results in Figures~\figcomplviews,~\figcomplpoints, and~\figcompledges. 
As shown in the figures, there is a clear linear trend of the running time of the proposed algorithm w.r.t. $k$ and $n$, respectively, and a quadratic trend w.r.t. $v$, in both DGF and SGF. Since the number of edges $n_e = k n$ and the empirical running time is linear in both $k$ and $n$, it can be concluded that the running time is linear in $n_e$, which aligns well with the theoretical result under our assumptions in Section~\ref{sec:complexity}. 
Due to these linear trends, our algorithm is among the most efficient multi-view clustering algorithms, as shown in Table~\ref{tab:time}.
}

\subsection{Parameters Sensitivity} \label{sec:param}
\blue{In this section, we conduct experiments to test the influence of the two hyper-parameters $\beta$ and $\gamma$  in the proposed multi-view graph learning algorithm. Since $\bs \lambda$ is meant to provide the users with an opportunity to incorporate prior knowledge through it, we use $\lambda_i=1$ in this paper for all views as no prior knowledge is involved. }
We test $(\beta,\gamma)$ on the grid $\{10^{-5},10^{-4},\dots,10^5\} \times \{10^{-5},10^{-4},\dots,10^5\}$. The results w.r.t. varying parameters $\beta$ and $\gamma$ are shown in Figure~\ref{fig:param}, which demonstrates that the performance of the proposed framework is stable across a wide range of parameters. We emphasize that even without parameter tuning, the framework still achieves generally superior performance against the state of the art. Please see the \app for the experiments. 

\begin{table*}[htbp]%
	\centering
	\caption{Average running time and standard deviation (in seconds) over 10 runs by different methods, where the time is recorded from constructing distance graphs to obtaining clustering result (i.e., the entire process); {DGF-0 and SGF-0 stand for the running time of the DGF and SGF algorithms, respectively, in the preliminary version of the paper \cite{liang2019consistency};} the running time of the fastest two multi-view clustering methods in each row (excluding single-view SC) is highlighted in bold.}
	\setlength{\tabcolsep}{2.8pt}
	\begin{tabular}{lcccccccccccc}
		\toprule
		{Dataset} & AASC  & AWP   & CoReg & MCGC  & MVGL  & RMSC  & WMSC  & SC (best) & DGF-0 & SGF-0 & DGF & SGF \\
		\midrule
		ORL & 0.36$_{\pm.00}$ & \textbbf{0.26}$_{\pm.00}$ & 7.11$_{\pm.19}$ & 1.05$_{\pm.02}$ & 3.97$_{\pm.02}$ & 3.00$_{\pm.02}$ & \textbbf{0.32}$_{\pm.00}$ & 0.26$_{\pm.00}$ & {0.39}$_{\pm.00}$ & {0.45}$_{\pm.00}$ & {0.39}$_{\pm.01}$ & {0.40}$_{\pm.00}$ \\
		Yale & 0.14$_{\pm.00}$ & \textbbf{0.07}$_{\pm.00}$ & 0.41$_{\pm.03}$ & 0.87$_{\pm.01}$ & 1.00$_{\pm.02}$ & 0.11$_{\pm.00}$ & \textbbf{0.09}$_{\pm.00}$ & 0.07$_{\pm.00}$ & {0.12}$_{\pm.00}$ & {0.12}$_{\pm.00}$ & {0.12}$_{\pm.00}$ & {0.12}$_{\pm.00}$ \\
		Reuters & 0.66$_{\pm.00}$ & \textbbf{0.62}$_{\pm.01}$ & 1.97$_{\pm.04}$ & 7.77$_{\pm.02}$ & 36.8$_{\pm.02}$ & 12.9$_{\pm.02}$ & 0.66$_{\pm.00}$ & 0.50$_{\pm.00}$ & {1.01}$_{\pm.01}$ & {0.87}$_{\pm.01}$ & \textbbf{0.62}$_{\pm.01}$ & \textbbf{0.63}$_{\pm.01}$ \\
		BBCSport & 0.13$_{\pm.01}$ & \textbbf{0.10}$_{\pm.00}$ & 3.75$_{\pm.02}$ & 0.95$_{\pm.02}$ & 7.62$_{\pm.02}$ & 0.50$_{\pm.01}$ & {0.12}$_{\pm.00}$ & 0.10$_{\pm.00}$ & {0.29}$_{\pm.00}$ & {0.16}$_{\pm.00}$ & \textbbf{0.11}$_{\pm.00}$ & \textbbf{0.11}$_{\pm.00}$ \\
		NUS-WIDE & 2.47$_{\pm.03}$ & \textbbf{0.90}$_{\pm.02}$ & 14.8$_{\pm.06}$ & 55.8$_{\pm.11}$ & 54.3$_{\pm.05}$ & 36.8$_{\pm.11}$ & 1.24$_{\pm.02}$ & 0.60$_{\pm.01}$ & {2.94}$_{\pm.02}$ & {6.37}$_{\pm.02}$ & \textbbf{1.12}$_{\pm.03}$ & {1.13}$_{\pm.02}$ \\
		Reuters-21578 & 1.70$_{\pm.01}$ & {1.64}$_{\pm.02}$ & 5.75$_{\pm.05}$ & 10.00$_{\pm.02}$ & 61.8$_{\pm.04}$ & 5.31$_{\pm.02}$ & 1.69$_{\pm.01}$ & 1.46$_{\pm.00}$ & {2.14}$_{\pm.01}$ & {2.33}$_{\pm.01}$ & \textbbf{1.56}$_{\pm.01}$ & \textbbf{1.54}$_{\pm.01}$ \\
		MSRC-v1 & 0.22$_{\pm.01}$ & \textbbf{0.04}$_{\pm.00}$ & 2.17$_{\pm.02}$ & 0.40$_{\pm.01}$ & 1.36$_{\pm.02}$ & 0.37$_{\pm.01}$ & \textbbf{0.06}$_{\pm.00}$ & 0.04$_{\pm.00}$ & {0.07}$_{\pm.00}$ & {0.09}$_{\pm.00}$ & {0.08}$_{\pm.01}$ & {0.07}$_{\pm.00}$ \\
		CiteSeer & 1.44$_{\pm.01}$ & \textbbf{1.18}$_{\pm.01}$ & 118$_{\pm.09}$ & 36.6$_{\pm.02}$ & 203$_{\pm.07}$ & 419$_{\pm1.6}$ & {1.32}$_{\pm.01}$ & 1.01$_{\pm.01}$ & {2.87}$_{\pm.03}$ & {2.41}$_{\pm.03}$ & {1.20}$_{\pm.02}$ & \textbbf{1.16}$_{\pm.02}$ \\
		ALOI & 39.4$_{\pm.41}$ & 36.1$_{\pm.18}$ & 2545$_{\pm7.0}$ & 713$_{\pm.21}$ & 5330$_{\pm6.1}$ & 388$_{\pm1.4}$ & 71.7$_{\pm.08}$ & 18.7$_{\pm.06}$ & {27.2}$_{\pm.07}$ & {25.9}$_{\pm.02}$ & \textbbf{9.53}$_{\pm.03}$ & \textbbf{9.67}$_{\pm.03}$ \\
		Flower17 & 0.70$_{\pm.01}$ & \textbbf{0.38}$_{\pm.01}$ & 50.0$_{\pm.19}$ & 24.3$_{\pm.04}$ & 69.3$_{\pm.04}$ & 9.21$_{\pm.21}$ & \textbbf{0.49}$_{\pm.00}$ & 0.10$_{\pm.01}$ & {1.33}$_{\pm.01}$ & {1.16}$_{\pm.00}$ & {0.62}$_{\pm.01}$ & {0.60}$_{\pm.01}$ \\
		Caltech101 & 42.5$_{\pm.14}$ & {38.8}$_{\pm.03}$ & 294$_{\pm.60}$ & 1129$_{\pm.49}$ & 7274$_{\pm8.5}$ & 2062$_{\pm8.0}$ & 41.2$_{\pm.07}$ & 30.2$_{\pm.07}$ & {54.8}$_{\pm.25}$ & {78.1}$_{\pm.12}$ & \textbbf{34.1}$_{\pm.32}$ & \textbbf{34.5}$_{\pm.29}$ \\
		UCI Digits & 1.02$_{\pm.01}$ & 1.33$_{\pm.03}$ & 352$_{\pm.68}$ & 15.7$_{\pm.03}$ & 79.5$_{\pm.05}$ & 22.5$_{\pm1.6}$ & 3.06$_{\pm.01}$ & 0.60$_{\pm.01}$ & {1.79}$_{\pm.00}$ & {6.61}$_{\pm.00}$ & \textbbf{0.81}$_{\pm.01}$ & \textbbf{0.79}$_{\pm.02}$ \\
		\bottomrule
	\end{tabular}
	\label{tab:time}%
\end{table*}

\subsection{Algorithmic Efficiency Comparison} \label{sec:empirical complexity}
We record the running time of each algorithm in our experiments and report the results in Table~\ref{tab:time}. We can see that DGF and SGF are among the two fastest multi-view algorithms on some datasets and run comparably fast on the other datasets against other algorithms. %
\blue{Besides, we also make an efficiency comparison with the DGF and SGF algorithms in the preliminary version of the paper \cite{liang2019consistency}. As shown in Table~\ref{tab:time}, the revised algorithms run clearly faster than the preliminary DGF and SGF algorithms~\cite{liang2019consistency} (denoted by DGF-0 and SGF-0) on the 12 datasets. }

DGF and SGF can be divided into two parts, which correspond to consistent graph learning (Algorithm~\ref{alg:graph}) and spectral clustering on the learned unified graph, respectively. \blue{Experiments show that the proposed graph learning algorithm is quite efficient comparing to the running time of the entire process of multi-view clustering (see Table~\tabtime). More interestingly, DGF and SGF run faster than single-view SC on the ALOI dataset (in Table~\ref{tab:time}). This may be because the learned similarity graphs in DGF and SGF have larger eigengap~\cite{eigengap-ijcai-2011}, which makes the eigen-decomposition easier and faster (intuitively, this may result from that the nodes on the learned graph are well clustered). }

Moreover, all comparing multi-view algorithms in this paper require eigen-decomposition (ED) or singular value decomposition (SVD) of an $n$-by-$n$ matrix at least two times (some require at least $v$ times), while the proposed DGF and SGF perform ED only once (which is in the spectral clustering on the unified graph). Since ED and SVD have at least $\Omega(n^2)$ time complexity \cite{saad2011numerical}, which is the complexity bottleneck in many multi-view clustering algorithms, our multi-view clustering algorithms could run several times faster than the comparing multi-view algorithms on very large datasets. 

\subsection{Consistency and Inconsistency of Multi-view Learning} 
In Figure~\ref{fig:visual}, we see that the learned unified graph by SGF is ``cleaner'' than any single view and clearly contains the consistent parts of all views. \blue{By the theoretical motivation of the proposed framework in Section~\ref{sec:motivation}, it could learn a better graph if the consistent part of multi-view graphs is the dominant part in most views, i.e., the inconsistent parts are sparse across views. As shown in Figure~\ref{fig:visual}, although the 4th single view is very noisy and very inconsistent with other views, the first three views exhibit consistent cluster patterns and thus the inconsistency is still \emph{sparse across views}.} 
\blue{
In the experiments, we do see, on the Caltech101 dataset, that the single-view SC on the \emph{best} view is slightly better than our algorithms (and is clearly better than other multi-view clustering methods). %
However, this does not undermine the advantages of the proposed framework where multi-view consistency and inconsistency are simultaneously exploited to learn a better unified graph, which can be verified by the significant performance improvements of the proposed approach on many datasets compared to the best single-view SC. }

\section{Conclusions} \label{conclusion}
This paper presents a novel multi-view graph learning approach, which for the first time simultaneously and explicitly models multi-view consistency as well as multi-view inconsistency in a unified optimization model, where multi-view consistency can be iteratively learned and fused into a unified graph as the multi-view inconsistency is automatically identified and removed. \blue{To optimize the objective, we design an efficient algorithm by exploiting the structures of the quadratic programs in the problem. The proposed algorithm has linear time complexity in the number of edges on the learned graph, even though exactly solving the problem is NP-hard. We further extend the proposed framework to two graph fusion versions, which correspond to distance (dissimilarity) graph fusion and similarity graph fusion, respectively.} %
Experimental results demonstrate the superiority, efficiency and robustness of the proposed algorithms against several state-of-the-art multi-view spectral clustering algorithms on a variety of real-world datasets. \blue{Remarkably, it maintains its good performance even without dataset-specific hyper-parameter tuning. }

\bibliographystyle{IEEEtran}
\bibliography{graphlearning}

\begin{thebibliography}{10}
\providecommand{\url}[1]{#1}
\csname url@samestyle\endcsname
\providecommand{\newblock}{\relax}
\providecommand{\bibinfo}[2]{#2}
\providecommand{\BIBentrySTDinterwordspacing}{\spaceskip=0pt\relax}
\providecommand{\BIBentryALTinterwordstretchfactor}{4}
\providecommand{\BIBentryALTinterwordspacing}{\spaceskip=\fontdimen2\font plus
\BIBentryALTinterwordstretchfactor\fontdimen3\font minus
  \fontdimen4\font\relax}
\providecommand{\BIBforeignlanguage}[2]{{%
\expandafter\ifx\csname l@#1\endcsname\relax
\typeout{** WARNING: IEEEtran.bst: No hyphenation pattern has been}%
\typeout{** loaded for the language `#1'. Using the pattern for}%
\typeout{** the default language instead.}%
\else
\language=\csname l@#1\endcsname
\fi
#2}}
\providecommand{\BIBdecl}{\relax}
\BIBdecl

\bibitem{jain10_survey}
A.~K. Jain, ``Data clustering: 50 years beyond $k$-means,'' \emph{Pattern
  Recognition Letters}, vol.~31, no.~8, pp. 651--666, 2010.

\bibitem{shi2000normalized}
J.~Shi and J.~Malik, ``Normalized cuts and image segmentation,'' \emph{IEEE
  Transactions on Pattern Analysis and Machine Intelligence}, vol.~22, no.~8,
  pp. 888--905, 2000.

\bibitem{ng2002spectral}
A.~Y. Ng, M.~I. Jordan, and Y.~Weiss, ``On spectral clustering: Analysis and an
  algorithm,'' in \emph{Advances in neural information processing systems},
  2002, pp. 849--856.

\bibitem{von2007tutorial}
U.~Von~Luxburg, ``A tutorial on spectral clustering,'' \emph{Statistics and
  computing}, vol.~17, no.~4, pp. 395--416, 2007.

\bibitem{Nie2014_kdd}
F.~Nie, X.~Wang, and H.~Huang, ``Clustering and projected clustering with
  adaptive neighbors,'' in \emph{Proceedings of the 20th ACM SIGKDD
  International Conference on Knowledge Discovery and Data Mining}, 2014, pp.
  977--986.

\bibitem{nie2016constrained}
F.~Nie, X.~Wang, M.~I. Jordan, and H.~Huang, ``The constrained laplacian rank
  algorithm for graph-based clustering,'' in \emph{Thirtieth AAAI Conference on
  Artificial Intelligence}, 2016.

\bibitem{zhan2017graph}
K.~{Zhan}, C.~{Zhang}, J.~{Guan}, and J.~{Wang}, ``Graph learning for multiview
  clustering,'' \emph{IEEE Transactions on Cybernetics}, vol.~48, no.~10, pp.
  2887--2895, 2018.

\bibitem{ZhanFusion}
K.~Zhan, C.~Niu, C.~Chen, F.~Nie, C.~Zhang, and Y.~Yang, ``Graph structure
  fusion for multiview clustering,'' \emph{IEEE Transactions on Knowledge and
  Data Engineering}, 2018.

\bibitem{Zhan2018}
K.~Zhan, F.~Nie, J.~Wang, and Y.~Yang, ``Multiview consensus graph
  clustering,'' \emph{IEEE Transactions on Image Processing}, vol.~28, no.~3,
  pp. 1261--1270, March 2019.

\bibitem{nie2017self}
F.~Nie, J.~Li, and X.~Li, ``Self-weighted multiview clustering with multiple
  graphs.'' in \emph{IJCAI}, 2017, pp. 2564--2570.

\bibitem{kdd2017robust}
A.~Bojchevski, Y.~Matkovic, and S.~G{\"u}nnemann, ``Robust spectral clustering
  for noisy data: Modeling sparse corruptions improves latent embeddings,'' in
  \emph{Proceedings of the 23rd ACM SIGKDD International Conference on
  Knowledge Discovery and Data Mining}.\hskip 1em plus 0.5em minus 0.4em\relax
  ACM, 2017, pp. 737--746.

\bibitem{liang2019consistency}
Y.~Liang, D.~Huang, and C.-D. Wang, ``Consistency meets inconsistency: A
  unified graph learning framework for multi-view clustering,'' in
  \emph{Proceedings of the IEEE International Conference on Data Mining}, 2019.

\bibitem{tao1998branch}
L.~T.~H. An and P.~D. Tao, ``A branch and bound method via d.c. optimization
  algorithms and ellipsoidal technique for box constrained nonconvex quadratic
  problems,'' \emph{Journal of Global Optimization}, vol.~13, no.~2, pp.
  171--206, 1998.

\bibitem{bickel2004multi}
S.~Bickel and T.~Scheffer, ``Multi-view clustering.'' in \emph{ICDM}, vol.~4,
  2004, pp. 19--26.

\bibitem{blum1998combining}
A.~Blum and T.~Mitchell, ``Combining labeled and unlabeled data with
  co-training,'' in \emph{Proceedings of the eleventh annual conference on
  Computational learning theory}.\hskip 1em plus 0.5em minus 0.4em\relax ACM,
  1998, pp. 92--100.

\bibitem{kumar2011co}
A.~Kumar, P.~Rai, and H.~Daum{\'e}, ``Co-regularized multi-view spectral
  clustering,'' in \emph{Advances in Neural Information Processing Systems},
  2011, pp. 1413--1421.

\bibitem{wang2016iterative}
Y.~Wang, W.~Zhang, L.~Wu, X.~Lin, M.~Fang, and S.~Pan, ``Iterative views
  agreement: An iterative low-rank based structured optimization method to
  multi-view spectral clustering,'' \emph{arXiv preprint arXiv:1608.05560},
  2016.

\bibitem{zong2018weighted}
L.~Zong, X.~Zhang, X.~Liu, and H.~Yu, ``Weighted multi-view spectral clustering
  based on spectral perturbation,'' in \emph{Thirty-Second AAAI Conference on
  Artificial Intelligence}, 2018.

\bibitem{wang2018multiview}
Y.~Wang, L.~Wu, X.~Lin, and J.~Gao, ``Multiview spectral clustering via
  structured low-rank matrix factorization,'' \emph{IEEE transactions on neural
  networks and learning systems}, no.~99, pp. 1--11, 2018.

\bibitem{xia2014robust}
R.~Xia, Y.~Pan, L.~Du, and J.~Yin, ``Robust multi-view spectral clustering via
  low-rank and sparse decomposition,'' in \emph{Twenty-Eighth AAAI Conference
  on Artificial Intelligence}, 2014.

\bibitem{nie2018multiview}
F.~Nie, L.~Tian, and X.~Li, ``Multiview clustering via adaptively weighted
  procrustes,'' in \emph{Proceedings of the 24th ACM SIGKDD International
  Conference on Knowledge Discovery \& Data Mining}.\hskip 1em plus 0.5em minus
  0.4em\relax ACM, 2018, pp. 2022--2030.

\bibitem{huang2012affinity}
H.-C. Huang, Y.-Y. Chuang, and C.-S. Chen, ``Affinity aggregation for spectral
  clustering,'' in \emph{2012 IEEE Conference on Computer Vision and Pattern
  Recognition}.\hskip 1em plus 0.5em minus 0.4em\relax IEEE, 2012, pp.
  773--780.

\bibitem{li2015convergence}
R.-C. Li and L.-H. Zhang, ``Convergence of the block lanczos method for
  eigenvalue clusters,'' \emph{Numerische Mathematik}, vol. 131, no.~1, pp.
  83--113, 2015.

\bibitem{burer2009nonconvex}
S.~Burer and A.~N. Letchford, ``On nonconvex quadratic programming with box
  constraints,'' \emph{SIAM Journal on Optimization}, vol.~20, no.~2, pp.
  1073--1089, 2009.

\bibitem{bomze1998standard}
I.~M. Bomze, ``On standard quadratic optimization problems,'' \emph{Journal of
  Global Optimization}, vol.~13, no.~4, pp. 369--387, 1998.

\bibitem{nips15-frankwolfe}
S.~Lacoste-Julien and M.~Jaggi, ``On the global linear convergence of
  frank-wolfe optimization variants,'' in \emph{Proceedings of the 28th
  International Conference on Neural Information Processing Systems - Volume
  1}, ser. NIPS'15, 2015, p. 496–504.

\bibitem{siam20-afw}
I.~M. Bomze, F.~Rinaldi, and D.~Zeffiro, ``Active set complexity of the
  away-step frank--wolfe algorithm,'' \emph{SIAM Journal on Optimization},
  vol.~30, no.~3, pp. 2470--2500, 2020.

\bibitem{siam19-active-set}
I.~M. Bomze, F.~Rinaldi, and S.~R. Bul\`{o}, ``First-order methods for the
  impatient: Support identification in finite time with convergent frank--wolfe
  variants,'' \emph{SIAM Journal on Optimization}, vol.~29, no.~3, pp.
  2211--2226, 2019.

\bibitem{sorensen1997implicitly}
D.~C. Sorensen, ``Implicitly restarted arnoldi/lanczos methods for large scale
  eigenvalue calculations,'' in \emph{Parallel Numerical Algorithms}.\hskip 1em
  plus 0.5em minus 0.4em\relax Springer, 1997, pp. 119--165.

\bibitem{hou2019eigen}
T.~Y. Hou, D.~Huang, K.~C. Lam, and Z.~Zhang, ``A fast hierarchically
  preconditioned eigensolver based on multiresolution matrix decomposition,''
  \emph{Multiscale Modeling \& Simulation}, vol.~17, no.~1, pp. 260--306, 2019.

\bibitem{nus-wide-civr09}
T.-S. Chua, J.~Tang, R.~Hong, H.~Li, Z.~Luo, and Y.-T. Zheng, ``Nus-wide: A
  real-world web image database from national university of singapore,'' in
  \emph{Proc. of ACM Conf. on Image and Video Retrieval (CIVR'09)}, Santorini,
  Greece., July 8-10, 2009.

\bibitem{winn2005locus}
J.~Winn and N.~Jojic, ``Locus: Learning object classes with unsupervised
  segmentation,'' in \emph{Tenth IEEE International Conference on Computer
  Vision (ICCV'05) Volume 1}, vol.~1.\hskip 1em plus 0.5em minus 0.4em\relax
  IEEE, 2005, pp. 756--763.

\bibitem{geusebroek2005amsterdam}
J.-M. Geusebroek, G.~J. Burghouts, and A.~W.-M. Smeulders, ``The amsterdam
  library of object images,'' \emph{International Journal of Computer Vision},
  vol.~61, no.~1, pp. 103--112, 2005.

\bibitem{houle2011knowledge}
M.~E. Houle, V.~Oria, S.~Satoh, and J.~Sun, ``Knowledge propagation in large
  image databases using neighborhood information,'' in \emph{Proceedings of the
  19th ACM international conference on Multimedia}, 2011, pp. 1033--1036.

\bibitem{nilsback2006visual}
M.-E. Nilsback and A.~Zisserman, ``A visual vocabulary for flower
  classification,'' in \emph{2006 IEEE Computer Society Conference on Computer
  Vision and Pattern Recognition (CVPR'06)}, vol.~2.\hskip 1em plus 0.5em minus
  0.4em\relax IEEE, 2006, pp. 1447--1454.

\bibitem{nilsback2008automated}
------, ``Automated flower classification over a large number of classes,'' in
  \emph{2008 Sixth Indian Conference on Computer Vision, Graphics \& Image
  Processing}.\hskip 1em plus 0.5em minus 0.4em\relax IEEE, 2008, pp. 722--729.

\bibitem{strehl2002cluster}
A.~Strehl and J.~Ghosh, ``Cluster ensembles---a knowledge reuse framework for
  combining multiple partitions,'' \emph{Journal of machine learning research},
  vol.~3, no. Dec, pp. 583--617, 2002.

\bibitem{vinh2010information}
N.~X. Vinh, J.~Epps, and J.~Bailey, ``Information theoretic measures for
  clusterings comparison: Variants, properties, normalization and correction
  for chance,'' \emph{Journal of Machine Learning Research}, vol.~11, no. Oct,
  pp. 2837--2854, 2010.

\bibitem{eigengap-ijcai-2011}
D.~Mavroeidis, ``Mind the eigen-gap, or how to accelerate semi-supervised
  spectral learning algorithms,'' in \emph{Proceedings of the Twenty-Second
  International Joint Conference on Artificial Intelligence}, ser. IJCAI'11,
  2011, p. 2692–2697.

\bibitem{saad2011numerical}
Y.~Saad, \emph{Numerical methods for large eigenvalue problems: revised
  edition}.\hskip 1em plus 0.5em minus 0.4em\relax SIAM, 2011.

\end{thebibliography}

\clearpage


\section*{Supplementary material}

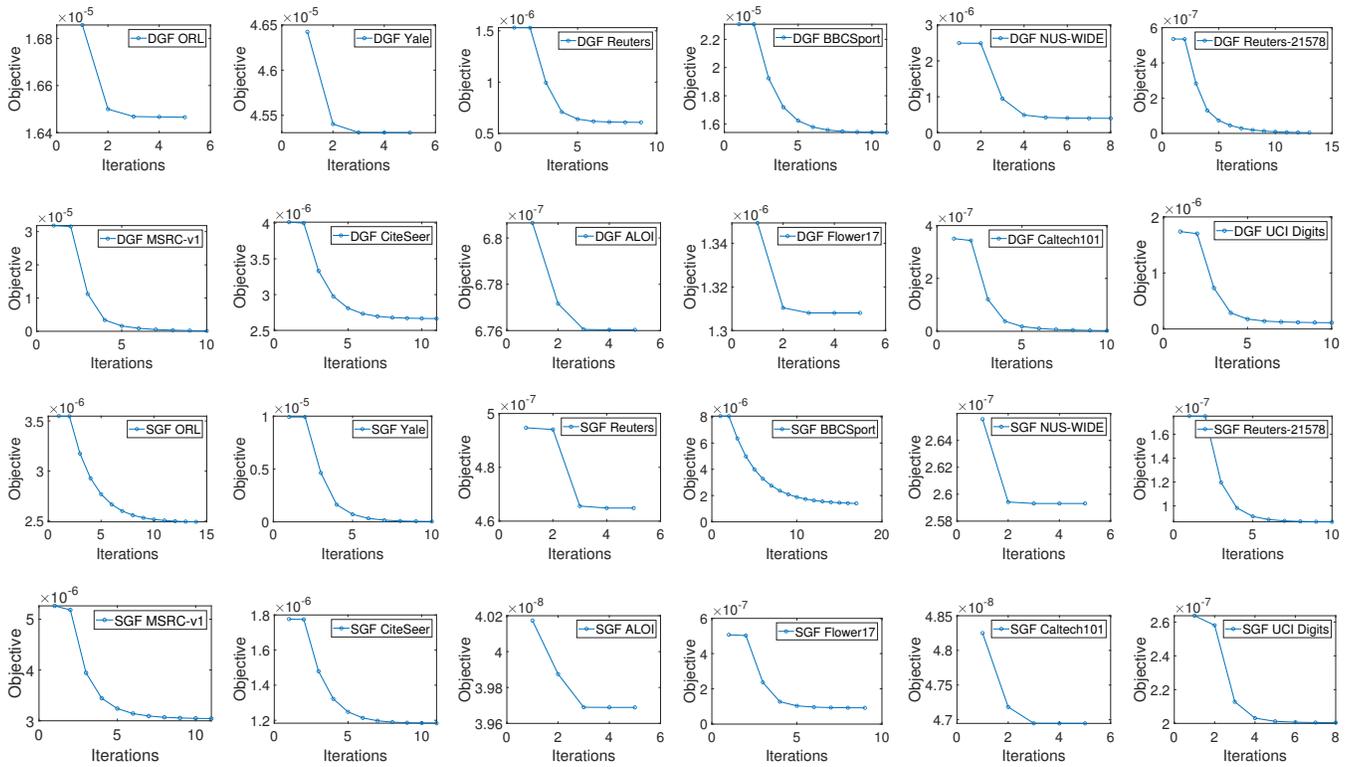

Fig. 1. Convergence curves of the proposed algorithms DGF and SGF on the 12 datasets. Note that the objective values are very small because we have normalized all adjacency matrices before performing consistent graph learning to improve convergence (see Section 4.3 in the main manuscript).

TABLE 3
Statistics and multi-view features (# dimensions) of the real-world datasets.

| Dataset | # views | # classes | # instances | View 1 | View 2 | View 3 | View 4 | View 5 | View 6 | View 7 |
|---|---|---|---|---|---|---|---|---|---|---|
| ORL | 3 | 40 | 400 | Intensity(4096) | LBP(3304) | Gabor(6750) | | | | |
| Yale | 3 | 15 | 165 | Intensity(4096) | LBP(3304) | Gabor(6750) | | | | |
| Reuters | 5 | 6 | 1200 | English(2000) | French(2000) | German(2000) | Italian(2000) | Spanish(2000) | | |
| BBCSport | 2 | 5 | 544 | seg1(3183) | seg2(3203) | | | | | |
| NUS-WIDE | 5 | 31 | 2000 | CH(65) | CM(226) | CORR(145) | EDH(74) | WT(129) | | |
| Reuters-21578 | 5 | 6 | 1500 | English(21531) | France(24892) | German(34251) | Italian(15506) | Spanish(11547) | | |
| MSRC-v1 | 5 | 7 | 210 | CM(24) | HOG(576) | GIST(512) | LBP(256) | CENT(254) | | |
| CiteSeer | 2 | 6 | 3312 | citations(3312) | content(3703) | | | | | |
| ALOI | 4 | 100 | 10800 | CS(77) | HAR(13) | HSB(64) | RGB(125) | | | |
| Flower17[a] | 7 | 17 | 1360 | Color | Texture | Shape | HOG | HSV | SIFT bdy | SIFT int |
| Caltech101 | 6 | 102 | 9144 | Gabor(48) | WM(40) | CENT(254) | HOG(1984) | GIST(512) | LBP(928) | |
| UCI Digits | 6 | 10 | 2000 | PIX(240) | FOU(76) | FAC(216) | ZER(47) | KAR(64) | MOR(6) | |

[a] Only distance matrices are available.

only need to solve Subproblems (1) and (2) in the main manuscript, where $\mathbf{A}$ is always the same as $\mathbf{W}$ (since no decomposition of $\mathbf{W}$ is required). Besides, the hyper-parameters $\beta$ and $\gamma$ in Algorithm 3 (Consistent Graph Learning) are set to 0 in order to optimize Eq. (1). We report the clustering performance of using Eq. (1) in Table 5, which shows that the proposed graph learning-based multi-view clustering framework achieve better or significantly better performance than using Eq. (1) as the objective on most of the benchmark datasets. Specifically, in terms of SGF, the average NMI(%), ACC(%), ARI(%), and purity(%) scores

(across the twelve datasets) have seen improvements from 56.91 to 61.62, from 59.18 to 63.85, from 45.10 to 51.01, and from 64.35 to 68.53, respectively, while in terms of DGF, the average NMI(%), ACC(%), ARI(%), and purity(%) scores (across the twelve datasets) have also gained improvements from 59.39 to 61.82, from 62.27 to 64.24, from 48.94 to 51.13, and from 66.83 to 68.81, respectively.

# REFERENCES

[1] J. Wu and J. M. Rehg, "Centrist: A visual descriptor for scene categorization," *IEEE Transactions on Pattern Analysis and Machine*

TABLE 4
Average clustering scores and standard deviation (%) over 10 runs by the revised DGF and SGF algorithms and the preliminary algorithms [7] (denoted by DGF-0 and SGF-0); the best score of the DGF-like (DGF and DGF-0) and SGF-like (SGF and SGF-0) algorithms are respectively highlighted in bold.

| Metric | Dataset | Fixed parameters $\beta$ and $\gamma$ | | | | Best $\beta$ and $\gamma$ | | | |
|---|---|---|---|---|---|---|---|---|---|
| | | SGF-0 | DGF-0 | SGF | DGF | SGF-0 | DGF-0 | SGF | DGF |
| NMI | ORL | $90.22_{\pm.48}$ | $80.65_{\pm.55}$ | $91.25_{\pm.24}$ | $91.22_{\pm.22}$ | $90.58_{\pm.46}$ | $88.49_{\pm.26}$ | $\mathbf{91.81_{\pm.47}}$ | $\mathbf{91.82_{\pm.30}}$ |
| | Yale | $71.12_{\pm.81}$ | $64.85_{\pm1.0}$ | $73.20_{\pm.00}$ | $73.90_{\pm.00}$ | $71.02_{\pm1.0}$ | $68.17_{\pm.31}$ | $\mathbf{74.04_{\pm.00}}$ | $\mathbf{73.90_{\pm.00}}$ |
| | Reuters | $9.15_{\pm.00}$ | $18.93_{\pm.58}$ | $14.81_{\pm.00}$ | $14.52_{\pm.00}$ | $14.67_{\pm2.1}$ | $\mathbf{19.56_{\pm.16}}$ | $\mathbf{14.99_{\pm.00}}$ | $16.31_{\pm.08}$ |
| | BBCSport | $69.59_{\pm.00}$ | $88.55_{\pm.00}$ | $69.82_{\pm.00}$ | $92.68_{\pm.00}$ | $70.72_{\pm.00}$ | $90.86_{\pm.00}$ | $\mathbf{93.96_{\pm.00}}$ | $\mathbf{92.68_{\pm.00}}$ |
| | NUS-WIDE | $19.13_{\pm.12}$ | $15.38_{\pm.32}$ | $19.41_{\pm.26}$ | $18.79_{\pm.27}$ | $\mathbf{19.73_{\pm.14}}$ | $16.35_{\pm.37}$ | $19.52_{\pm.21}$ | $\mathbf{19.96_{\pm.22}}$ |
| | Reuters-21578 | $25.59_{\pm.00}$ | $30.32_{\pm.00}$ | $30.03_{\pm.49}$ | $29.46_{\pm.08}$ | $28.30_{\pm.00}$ | $30.59_{\pm.04}$ | $\mathbf{31.40_{\pm.11}}$ | $\mathbf{31.03_{\pm.01}}$ |
| | MSRC-v1 | $70.28_{\pm.02}$ | $65.71_{\pm.31}$ | $75.66_{\pm.12}$ | $77.67_{\pm.00}$ | $74.18_{\pm.71}$ | $70.48_{\pm.17}$ | $\mathbf{78.91_{\pm.00}}$ | $\mathbf{80.98_{\pm.00}}$ |
| | CiteSeer | $0.25_{\pm.00}$ | $27.26_{\pm.03}$ | $37.50_{\pm.02}$ | $37.09_{\pm.04}$ | $0.28_{\pm.03}$ | $27.64_{\pm.04}$ | $\mathbf{37.55_{\pm.00}}$ | $\mathbf{38.47_{\pm.05}}$ |
| | ALOI | $90.61_{\pm.32}$ | $79.50_{\pm.20}$ | $90.44_{\pm.42}$ | $90.55_{\pm.48}$ | $91.03_{\pm.20}$ | $81.80_{\pm.14}$ | $\mathbf{91.08_{\pm.32}}$ | $\mathbf{90.98_{\pm.28}}$ |
| | Flower17 | $56.58_{\pm.35}$ | $43.26_{\pm.52}$ | $64.25_{\pm.32}$ | $64.56_{\pm.22}$ | $59.76_{\pm.23}$ | $45.87_{\pm.19}$ | $\mathbf{66.00_{\pm.70}}$ | $\mathbf{65.34_{\pm.78}}$ |
| | Caltech101 | $43.10_{\pm.20}$ | $38.56_{\pm.14}$ | $45.62_{\pm.22}$ | $45.07_{\pm.22}$ | $\mathbf{49.31_{\pm.21}}$ | $39.70_{\pm.28}$ | $46.09_{\pm.36}$ | $\mathbf{46.44_{\pm.21}}$ |
| | UCI Digits | $94.46_{\pm.00}$ | $79.19_{\pm.00}$ | $95.42_{\pm.00}$ | $95.15_{\pm.00}$ | $95.25_{\pm.00}$ | $83.31_{\pm.00}$ | $\mathbf{95.63_{\pm.00}}$ | $\mathbf{95.77_{\pm.00}}$ |
| ACC | ORL | $83.55_{\pm.65}$ | $66.30_{\pm1.1}$ | $84.35_{\pm.38}$ | $83.90_{\pm.55}$ | $84.20_{\pm1.5}$ | $81.20_{\pm.45}$ | $\mathbf{85.05_{\pm1.0}}$ | $\mathbf{84.50_{\pm.53}}$ |
| | Yale | $68.48_{\pm.00}$ | $65.33_{\pm.51}$ | $70.91_{\pm.00}$ | $70.91_{\pm.00}$ | $68.48_{\pm.00}$ | $65.70_{\pm.54}$ | $\mathbf{70.91_{\pm.00}}$ | $\mathbf{70.91_{\pm.00}}$ |
| | Reuters | $20.67_{\pm.00}$ | $38.12_{\pm.43}$ | $29.92_{\pm.00}$ | $29.83_{\pm.00}$ | $27.48_{\pm1.7}$ | $\mathbf{39.13_{\pm.11}}$ | $29.97_{\pm.07}$ | $\mathbf{31.90_{\pm.14}}$ |
| | BBCSport | $71.88_{\pm.00}$ | $96.14_{\pm.00}$ | $71.88_{\pm.00}$ | $97.98_{\pm.00}$ | $72.24_{\pm.00}$ | $97.24_{\pm.00}$ | $\mathbf{98.35_{\pm.00}}$ | $\mathbf{97.98_{\pm.00}}$ |
| | NUS-WIDE | $15.86_{\pm.52}$ | $13.41_{\pm.49}$ | $16.31_{\pm.73}$ | $15.33_{\pm.58}$ | $\mathbf{16.88_{\pm.34}}$ | $14.50_{\pm.52}$ | $15.99_{\pm.64}$ | $\mathbf{16.39_{\pm.51}}$ |
| | Reuters-21578 | $47.27_{\pm.00}$ | $48.47_{\pm.00}$ | $50.80_{\pm.48}$ | $49.85_{\pm.30}$ | $49.67_{\pm.00}$ | $48.59_{\pm.03}$ | $\mathbf{51.64_{\pm.28}}$ | $\mathbf{50.59_{\pm.03}}$ |
| | MSRC-v1 | $76.19_{\pm.00}$ | $80.67_{\pm.26}$ | $\mathbf{81.14_{\pm.26}}$ | $79.05_{\pm.00}$ | $78.38_{\pm.26}$ | $82.00_{\pm.21}$ | $80.48_{\pm.00}$ | $\mathbf{87.14_{\pm.00}}$ |
| | CiteSeer | $21.14_{\pm.00}$ | $52.78_{\pm.03}$ | $\mathbf{63.44_{\pm.00}}$ | $\mathbf{63.67_{\pm.04}}$ | $21.20_{\pm.00}$ | $52.34_{\pm.07}$ | $63.40_{\pm.01}$ | $63.64_{\pm.09}$ |
| | ALOI | $84.44_{\pm1.4}$ | $72.34_{\pm.83}$ | $82.15_{\pm1.4}$ | $82.66_{\pm.97}$ | $\mathbf{84.93_{\pm.79}}$ | $74.76_{\pm.46}$ | $84.17_{\pm1.6}$ | $\mathbf{84.18_{\pm1.3}}$ |
| | Flower17 | $61.59_{\pm.65}$ | $48.51_{\pm.58}$ | $66.38_{\pm.72}$ | $66.50_{\pm.24}$ | $64.51_{\pm.52}$ | $51.29_{\pm.26}$ | $\mathbf{68.06_{\pm1.4}}$ | $\mathbf{67.88_{\pm1.6}}$ |
| | Caltech101 | $19.91_{\pm.30}$ | $19.77_{\pm.24}$ | $23.06_{\pm.48}$ | $22.06_{\pm.48}$ | $\mathbf{26.04_{\pm.94}}$ | $21.93_{\pm.63}$ | $\mathbf{23.45_{\pm.44}}$ | $23.34_{\pm.89}$ |
| | UCI Digits | $97.55_{\pm.00}$ | $79.55_{\pm.00}$ | $98.00_{\pm.00}$ | $97.90_{\pm.00}$ | $97.90_{\pm.00}$ | $82.60_{\pm.00}$ | $\mathbf{98.10_{\pm.00}}$ | $\mathbf{98.20_{\pm.00}}$ |
| ARI | ORL | $75.13_{\pm1.5}$ | $53.61_{\pm1.2}$ | $77.11_{\pm.91}$ | $77.21_{\pm.72}$ | $75.98_{\pm1.2}$ | $72.18_{\pm.83}$ | $\mathbf{78.80_{\pm1.0}}$ | $\mathbf{77.95_{\pm.71}}$ |
| | Yale | $50.74_{\pm1.5}$ | $44.29_{\pm1.3}$ | $54.53_{\pm.00}$ | $54.83_{\pm.00}$ | $50.67_{\pm1.6}$ | $47.31_{\pm.66}$ | $\mathbf{55.07_{\pm.00}}$ | $\mathbf{54.83_{\pm.00}}$ |
| | Reuters | $1.62_{\pm.00}$ | $11.75_{\pm1.8}$ | $6.46_{\pm.00}$ | $6.37_{\pm.00}$ | $3.58_{\pm.94}$ | $\mathbf{13.35_{\pm.20}}$ | $\mathbf{6.52_{\pm.02}}$ | $8.90_{\pm.16}$ |
| | BBCSport | $59.97_{\pm.00}$ | $90.56_{\pm.00}$ | $60.49_{\pm.00}$ | $94.76_{\pm.00}$ | $60.56_{\pm.00}$ | $93.26_{\pm.00}$ | $\mathbf{95.53_{\pm.00}}$ | $\mathbf{94.76_{\pm.00}}$ |
| | NUS-WIDE | $4.70_{\pm.25}$ | $3.54_{\pm.27}$ | $4.96_{\pm.36}$ | $4.76_{\pm.24}$ | $\mathbf{5.21_{\pm.27}}$ | $4.13_{\pm.29}$ | $4.92_{\pm.33}$ | $\mathbf{5.78_{\pm.27}}$ |
| | Reuters-21578 | $18.84_{\pm.00}$ | $24.53_{\pm.00}$ | $\mathbf{23.69_{\pm.37}}$ | $22.45_{\pm.04}$ | $22.18_{\pm.00}$ | $\mathbf{24.66_{\pm.02}}$ | $23.54_{\pm.05}$ | $21.31_{\pm.00}$ |
| | MSRC-v1 | $62.27_{\pm.01}$ | $59.93_{\pm.49}$ | $69.45_{\pm.14}$ | $70.59_{\pm.00}$ | $68.17_{\pm.71}$ | $64.58_{\pm.21}$ | $\mathbf{72.31_{\pm.00}}$ | $\mathbf{75.35_{\pm.00}}$ |
| | CiteSeer | $-0.02_{\pm.00}$ | $20.09_{\pm.04}$ | $37.63_{\pm.00}$ | $37.41_{\pm.05}$ | $-0.02_{\pm.00}$ | $22.11_{\pm.02}$ | $\mathbf{37.63_{\pm.01}}$ | $\mathbf{38.15_{\pm.08}}$ |
| | ALOI | $78.15_{\pm1.3}$ | $58.58_{\pm.79}$ | $76.79_{\pm1.6}$ | $77.01_{\pm1.4}$ | $78.81_{\pm.67}$ | $62.68_{\pm.32}$ | $\mathbf{79.15_{\pm1.3}}$ | $\mathbf{78.27_{\pm.40}}$ |
| | Flower17 | $42.43_{\pm.35}$ | $28.58_{\pm.53}$ | $50.47_{\pm.58}$ | $50.36_{\pm.32}$ | $46.01_{\pm.34}$ | $30.99_{\pm.32}$ | $\mathbf{52.27_{\pm1.2}}$ | $\mathbf{51.60_{\pm1.3}}$ |
| | Caltech101 | $12.35_{\pm.64}$ | $12.33_{\pm.26}$ | $14.19_{\pm.67}$ | $13.45_{\pm.46}$ | $\mathbf{16.07_{\pm.96}}$ | $12.88_{\pm.29}$ | $\mathbf{14.60_{\pm.46}}$ | $14.07_{\pm.84}$ |
| | UCI Digits | $94.64_{\pm.00}$ | $57.30_{\pm.00}$ | $95.60_{\pm.00}$ | $95.38_{\pm.00}$ | $95.37_{\pm.00}$ | $78.12_{\pm.00}$ | $\mathbf{95.82_{\pm.00}}$ | $\mathbf{96.04_{\pm.00}}$ |
| purity | ORL | $85.15_{\pm.72}$ | $69.25_{\pm.68}$ | $86.80_{\pm.37}$ | $86.40_{\pm.22}$ | $85.70_{\pm.78}$ | $83.30_{\pm.48}$ | $\mathbf{87.65_{\pm.68}}$ | $\mathbf{86.85_{\pm.68}}$ |
| | Yale | $68.97_{\pm.27}$ | $65.58_{\pm.66}$ | $70.91_{\pm.00}$ | $70.91_{\pm.00}$ | $68.97_{\pm.27}$ | $65.70_{\pm.54}$ | $\mathbf{70.91_{\pm.00}}$ | $\mathbf{70.91_{\pm.00}}$ |
| | Reuters | $25.08_{\pm.00}$ | $40.20_{\pm1.3}$ | $34.33_{\pm.00}$ | $34.42_{\pm.00}$ | $31.57_{\pm2.0}$ | $\mathbf{42.32_{\pm.15}}$ | $\mathbf{34.42_{\pm.00}}$ | $35.97_{\pm.05}$ |
| | BBCSport | $55.74_{\pm.00}$ | $96.14_{\pm.00}$ | $55.74_{\pm.00}$ | $97.98_{\pm.00}$ | $75.92_{\pm.00}$ | $97.24_{\pm.00}$ | $\mathbf{98.35_{\pm.00}}$ | $\mathbf{97.98_{\pm.00}}$ |
| | NUS-WIDE | $25.97_{\pm.47}$ | $23.59_{\pm.30}$ | $25.65_{\pm.34}$ | $25.61_{\pm.54}$ | $\mathbf{26.83_{\pm.39}}$ | $24.44_{\pm.51}$ | $25.79_{\pm.37}$ | $\mathbf{26.92_{\pm.54}}$ |
| | Reuters-21578 | $54.20_{\pm.00}$ | $58.00_{\pm.00}$ | $57.12_{\pm.45}$ | $56.19_{\pm.30}$ | $56.07_{\pm.00}$ | $\mathbf{58.19_{\pm.03}}$ | $\mathbf{57.97_{\pm.28}}$ | $56.99_{\pm.03}$ |
| | MSRC-v1 | $80.00_{\pm.00}$ | $80.67_{\pm.26}$ | $82.86_{\pm.00}$ | $83.33_{\pm.00}$ | $82.00_{\pm.40}$ | $82.00_{\pm.21}$ | $\mathbf{83.81_{\pm.00}}$ | $\mathbf{87.14_{\pm.00}}$ |
| | CiteSeer | $21.29_{\pm.00}$ | $54.55_{\pm.01}$ | $65.94_{\pm.00}$ | $66.11_{\pm.04}$ | $21.30_{\pm.02}$ | $54.29_{\pm.05}$ | $\mathbf{65.95_{\pm.01}}$ | $\mathbf{66.58_{\pm.04}}$ |
| | ALOI | $86.54_{\pm1.0}$ | $84.62_{\pm1.1}$ | $84.96_{\pm1.1}$ | $\mathbf{86.54_{\pm.77}}$ | $76.80_{\pm.42}$ | $\mathbf{85.94_{\pm1.1}}$ | $\mathbf{86.21_{\pm.00}}$ | |
| | Flower17 | $62.31_{\pm.39}$ | $51.18_{\pm.76}$ | $67.84_{\pm.75}$ | $68.54_{\pm.23}$ | $65.44_{\pm.28}$ | $52.54_{\pm.34}$ | $\mathbf{69.88_{\pm1.4}}$ | $\mathbf{69.81_{\pm1.5}}$ |
| | Caltech101 | $42.71_{\pm.29}$ | $41.17_{\pm.06}$ | $45.44_{\pm.34}$ | $44.66_{\pm.40}$ | $\mathbf{48.28_{\pm.38}}$ | $41.84_{\pm.72}$ | $45.97_{\pm.47}$ | $\mathbf{45.96_{\pm.39}}$ |
| | UCI Digits | $97.55_{\pm.00}$ | $82.05_{\pm.00}$ | $98.00_{\pm.00}$ | $97.90_{\pm.00}$ | $97.90_{\pm.00}$ | $85.25_{\pm.00}$ | $\mathbf{98.10_{\pm.00}}$ | $\mathbf{98.20_{\pm.00}}$ |

*Intelligence*, vol. 33, no. 8, pp. 1489–1501, 2010.

[2] A. Oliva and A. Torralba, "Modeling the shape of the scene: A holistic representation of the spatial envelope," *International journal of computer vision*, vol. 42, no. 3, pp. 145–175, 2001.

[3] R. M. Haralick, K. Shanmugam, and I. H. Dinstein, "Textural features for image classification," *IEEE Transactions on systems, man, and cybernetics*, no. 6, pp. 610–621, 1973.

[4] N. Dalal and B. Triggs, "Histograms of oriented gradients for human detection," in *2005 IEEE computer society conference on computer vision and pattern recognition (CVPR'05)*, vol. 1. IEEE, 2005, pp. 886–893.

[5] D. G. Lowe, "Distinctive image features from scale-invariant keypoints," *International journal of computer vision*, vol. 60, no. 2, pp. 91–110, 2004.

[6] C.-H. Teh and R. T. Chin, "On image analysis by the methods of moments," *IEEE Transactions on pattern analysis and machine intelligence*, vol. 10, no. 4, pp. 496–513, 1988.

[7] Y. Liang, D. Huang, and C.-D. Wang, "Consistency meets inconsistency: A unified graph learning framework for multi-view clustering," in *Proceedings of the IEEE International Conference on Data Mining*, 2019.

TABLE 5
Comparison between the proposed framework and the variant using Eq. (1) under different metrics (%).

| Metric | Dataset | SGF | | DGF | |
|---|---|---|---|---|---|
| | | − | Using Eq. (1) | − | Using Eq. (1) |
| NMI | ORL | $91.45_{\pm.32}$ | $90.48_{\pm.46}$ | $91.63_{\pm.30}$ | $90.25_{\pm.55}$ |
| | Yale | $74.04_{\pm.00}$ | $71.31_{\pm.47}$ | $73.90_{\pm.00}$ | $70.87_{\pm.29}$ |
| | Reuters | $14.99_{\pm.00}$ | $14.36_{\pm.00}$ | $16.30_{\pm.09}$ | $14.19_{\pm.00}$ |
| | BBCSport | $93.96_{\pm.00}$ | $69.08_{\pm.00}$ | $92.68_{\pm.00}$ | $91.55_{\pm.00}$ |
| | NUS-WIDE | $19.41_{\pm.24}$ | $19.21_{\pm.51}$ | $19.83_{\pm.20}$ | $18.56_{\pm.17}$ |
| | Reuters-21578 | $31.40_{\pm.02}$ | $27.40_{\pm.31}$ | $31.03_{\pm.01}$ | $27.18_{\pm.02}$ |
| | MSRC-v1 | $78.86_{\pm.10}$ | $70.54_{\pm.00}$ | $80.98_{\pm.00}$ | $75.29_{\pm.00}$ |
| | CiteSeer | $37.54_{\pm.02}$ | $37.24_{\pm.00}$ | $38.46_{\pm.07}$ | $37.88_{\pm.12}$ |
| | ALOI | $90.74_{\pm.33}$ | $89.47_{\pm.32}$ | $90.43_{\pm.38}$ | $90.52_{\pm.51}$ |
| | Flower17 | $65.61_{\pm.11}$ | $55.69_{\pm.54}$ | $64.65_{\pm.21}$ | $56.87_{\pm.44}$ |
| | Caltech101 | $45.78_{\pm.18}$ | $42.65_{\pm.23}$ | $46.18_{\pm.49}$ | $44.58_{\pm.37}$ |
| | UCI-Digits | $95.63_{\pm.00}$ | $95.44_{\pm.00}$ | $95.77_{\pm.00}$ | $94.99_{\pm.00}$ |
| | Avg. Score | $61.62_{\pm.11}$ | $56.91_{\pm.24}$ | $61.82_{\pm.15}$ | $59.39_{\pm.21}$ |
| ACC | ORL | $84.15_{\pm.88}$ | $83.25_{\pm1.4}$ | $83.65_{\pm1.2}$ | $82.95_{\pm.78}$ |
| | Yale | $70.91_{\pm.00}$ | $69.09_{\pm.00}$ | $70.91_{\pm.00}$ | $69.09_{\pm.00}$ |
| | Reuters | $30.00_{\pm.00}$ | $28.83_{\pm.00}$ | $31.92_{\pm.12}$ | $29.57_{\pm.04}$ |
| | BBCSport | $98.35_{\pm.00}$ | $71.69_{\pm.00}$ | $97.98_{\pm.00}$ | $97.61_{\pm.00}$ |
| | NUS-WIDE | $15.60_{\pm.38}$ | $15.37_{\pm.15}$ | $15.96_{\pm.27}$ | $15.18_{\pm.39}$ |
| | Reuters-21578 | $51.64_{\pm.24}$ | $47.56_{\pm.33}$ | $50.60_{\pm.00}$ | $47.39_{\pm.07}$ |
| | MSRC-v1 | $80.38_{\pm.21}$ | $76.19_{\pm.00}$ | $87.14_{\pm.00}$ | $79.05_{\pm.00}$ |
| | CiteSeer | $63.39_{\pm.03}$ | $62.95_{\pm.00}$ | $63.56_{\pm.15}$ | $63.86_{\pm.25}$ |
| | ALOI | $83.54_{\pm1.1}$ | $81.90_{\pm1.7}$ | $81.61_{\pm1.5}$ | $82.39_{\pm2.4}$ |
| | Flower17 | $67.32_{\pm.15}$ | $54.26_{\pm1.4}$ | $66.59_{\pm.25}$ | $60.32_{\pm.93}$ |
| | Caltech101 | $22.76_{\pm.69}$ | $20.99_{\pm.70}$ | $22.82_{\pm.86}$ | $21.94_{\pm.64}$ |
| | UCI-Digits | $98.10_{\pm.00}$ | $98.05_{\pm.00}$ | $98.20_{\pm.00}$ | $97.85_{\pm.00}$ |
| | Avg. Score | $63.85_{\pm.30}$ | $59.18_{\pm.47}$ | $64.24_{\pm.37}$ | $62.27_{\pm.46}$ |
| ARI | ORL | $77.54_{\pm.83}$ | $75.97_{\pm1.2}$ | $77.80_{\pm.89}$ | $75.44_{\pm1.3}$ |
| | Yale | $55.07_{\pm.00}$ | $50.53_{\pm.94}$ | $54.83_{\pm.00}$ | $49.99_{\pm.22}$ |
| | Reuters | $6.53_{\pm.00}$ | $5.63_{\pm.00}$ | $8.89_{\pm.17}$ | $6.09_{\pm.01}$ |
| | BBCSport | $95.53_{\pm.00}$ | $59.75_{\pm.00}$ | $94.76_{\pm.00}$ | $93.96_{\pm.00}$ |
| | NUS-WIDE | $4.81_{\pm.19}$ | $4.82_{\pm.09}$ | $5.54_{\pm.09}$ | $4.60_{\pm.27}$ |
| | Reuters-21578 | $23.54_{\pm.02}$ | $20.58_{\pm.05}$ | $21.32_{\pm.01}$ | $20.22_{\pm.01}$ |
| | MSRC-v1 | $72.25_{\pm.14}$ | $63.32_{\pm.00}$ | $75.35_{\pm.00}$ | $69.27_{\pm.00}$ |
| | CiteSeer | $37.62_{\pm.02}$ | $37.47_{\pm.00}$ | $38.10_{\pm.14}$ | $39.24_{\pm.17}$ |
| | ALOI | $78.01_{\pm.68}$ | $74.24_{\pm1.4}$ | $76.58_{\pm1.5}$ | $76.84_{\pm2.1}$ |
| | Flower17 | $51.61_{\pm.20}$ | $39.67_{\pm.87}$ | $50.48_{\pm.28}$ | $42.31_{\pm.43}$ |
| | Caltech101 | $13.83_{\pm.60}$ | $13.56_{\pm.52}$ | $13.87_{\pm1.2}$ | $14.05_{\pm.60}$ |
| | UCI-Digits | $95.82_{\pm.00}$ | $95.72_{\pm.00}$ | $96.04_{\pm.00}$ | $95.27_{\pm.00}$ |
| | Avg. Score | $51.01_{\pm.23}$ | $45.10_{\pm.42}$ | $51.13_{\pm.36}$ | $48.94_{\pm.42}$ |
| purity | ORL | $86.90_{\pm.52}$ | $85.10_{\pm1.1}$ | $86.45_{\pm.87}$ | $84.70_{\pm.84}$ |
| | Yale | $70.91_{\pm.00}$ | $69.09_{\pm.00}$ | $70.91_{\pm.00}$ | $69.09_{\pm.00}$ |
| | Reuters | $34.42_{\pm.00}$ | $34.17_{\pm.00}$ | $35.93_{\pm.11}$ | $34.17_{\pm.00}$ |
| | BBCSport | $98.35_{\pm.00}$ | $75.55_{\pm.00}$ | $97.98_{\pm.00}$ | $97.61_{\pm.00}$ |
| | NUS-WIDE | $25.61_{\pm.30}$ | $25.52_{\pm.42}$ | $26.92_{\pm.18}$ | $25.43_{\pm.36}$ |
| | Reuters-21578 | $57.97_{\pm.24}$ | $54.88_{\pm.03}$ | $57.00_{\pm.00}$ | $54.60_{\pm.00}$ |
| | MSRC-v1 | $83.14_{\pm.00}$ | $80.48_{\pm.00}$ | $87.14_{\pm.00}$ | $82.86_{\pm.00}$ |
| | CiteSeer | $65.94_{\pm.00}$ | $65.07_{\pm.00}$ | $66.58_{\pm.05}$ | $64.81_{\pm.18}$ |
| | ALOI | $85.38_{\pm1.1}$ | $84.17_{\pm.96}$ | $84.29_{\pm1.2}$ | $84.58_{\pm1.7}$ |
| | Flower17 | $69.15_{\pm.14}$ | $57.35_{\pm.85}$ | $68.59_{\pm.14}$ | $62.07_{\pm.77}$ |
| | Caltech101 | $45.82_{\pm.30}$ | $42.78_{\pm.31}$ | $45.68_{\pm.52}$ | $44.19_{\pm.67}$ |
| | UCI-Digits | $98.10_{\pm.00}$ | $98.05_{\pm.00}$ | $98.20_{\pm.00}$ | $97.85_{\pm.00}$ |
| | Avg. Score | $68.53_{\pm.21}$ | $64.35_{\pm.30}$ | $68.81_{\pm.26}$ | $66.83_{\pm.37}$ |

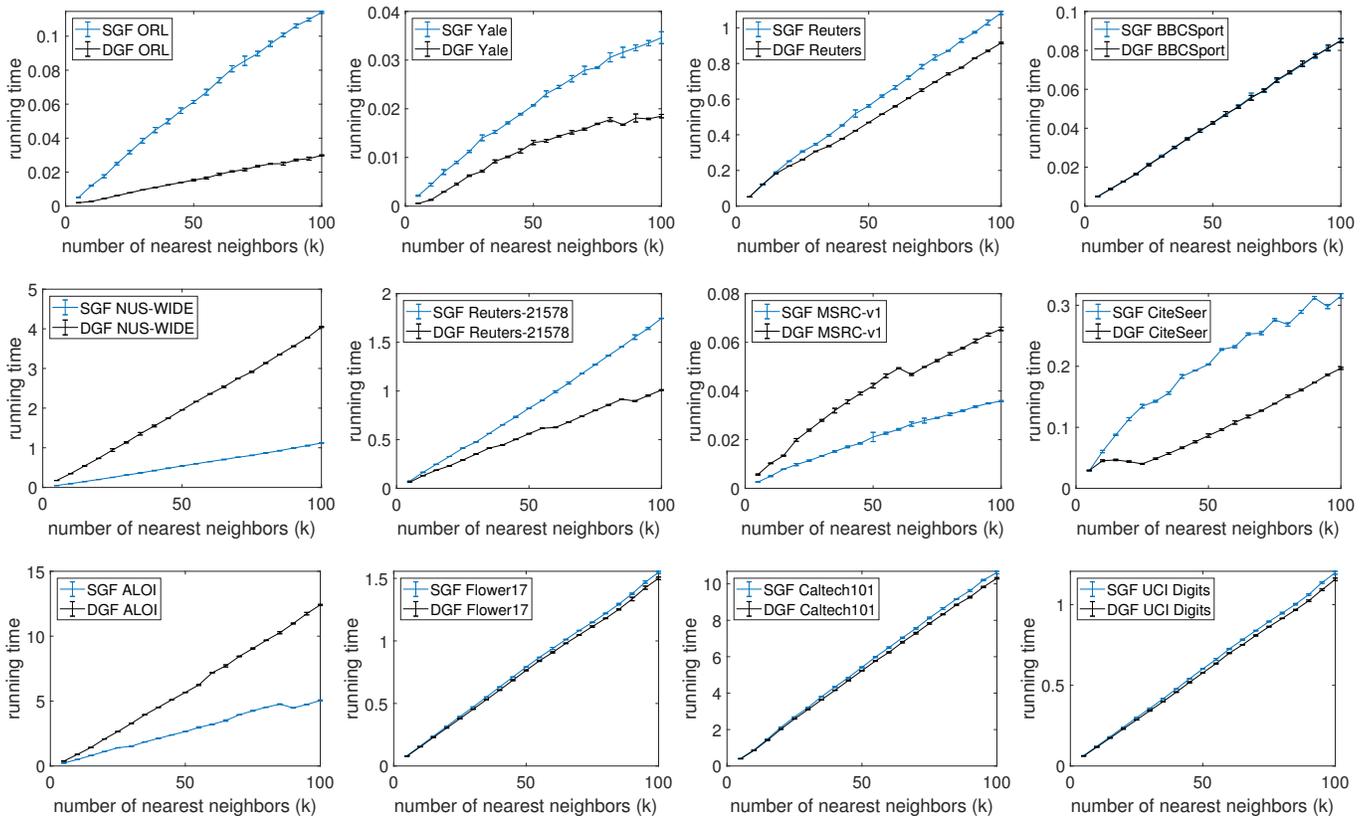

Fig. 2. Running time (in seconds) of the proposed algorithms DGF and SGF against the number of nearest neighbors (in $k$NN) for the 12 datasets. We fix the number of samples (using all samples from the datasets) and vary the number of nearest neighbors (in $\{5, 10, \ldots, 100\}$) to perform the experiments. For each point on the plots, we repeat the experiment for 5 times and plot the mean and standard deviation of the results.

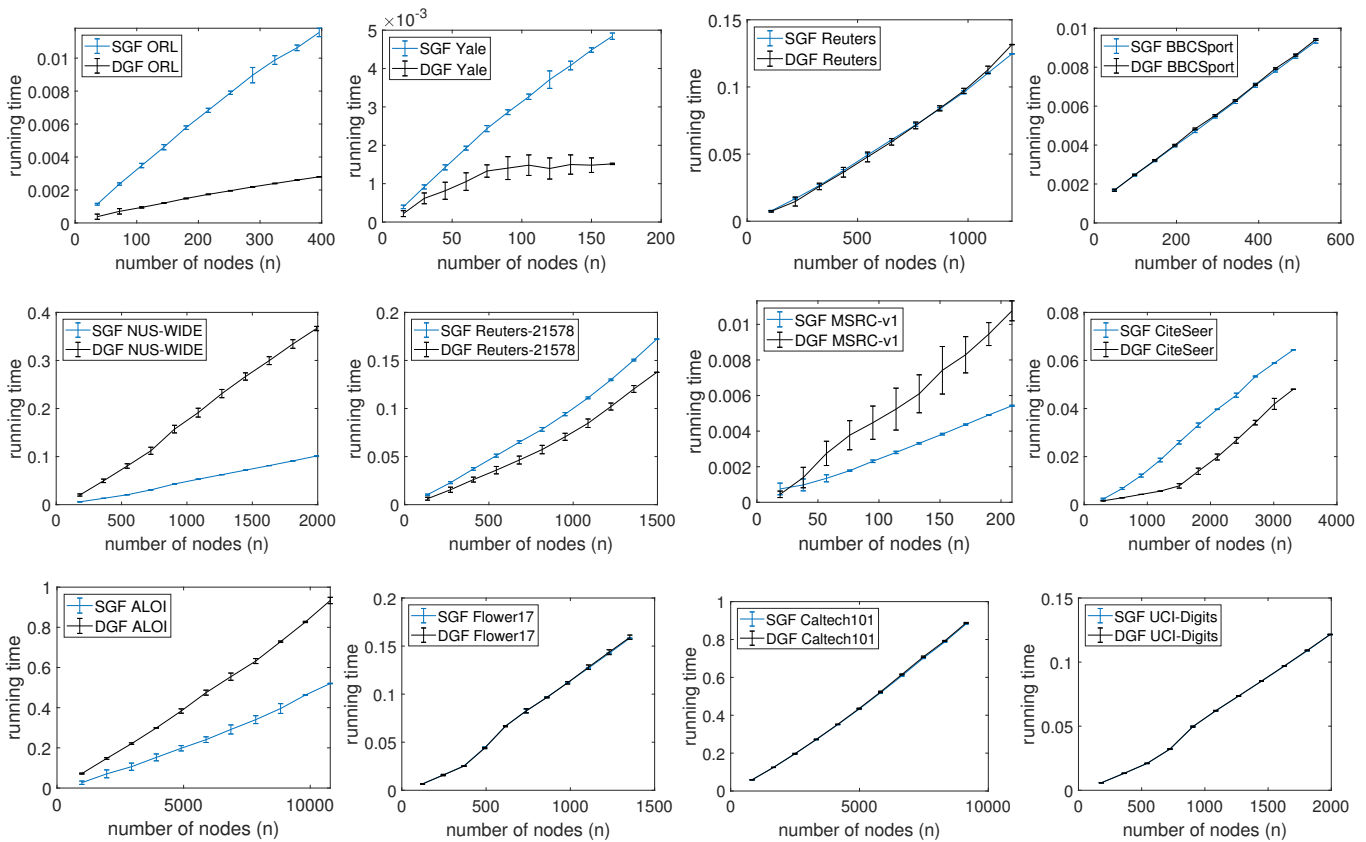

Fig. 3. Running time (in seconds) of the proposed algorithms DGF and SGF against the number of nodes on a single-view graph (i.e., number of data points) for the 12 datasets. We fix the number of nearest neighbors to 10 and vary the number of data points to perform the experiments. For each point on the plots, we randomly sample the required number of data points (without replacement) from the entire dataset, repeat the sampling for 5 times, and plot the mean and standard deviation of the results. Note that on each of the last three figures there are indeed 2 curves, which are highly overlapping.

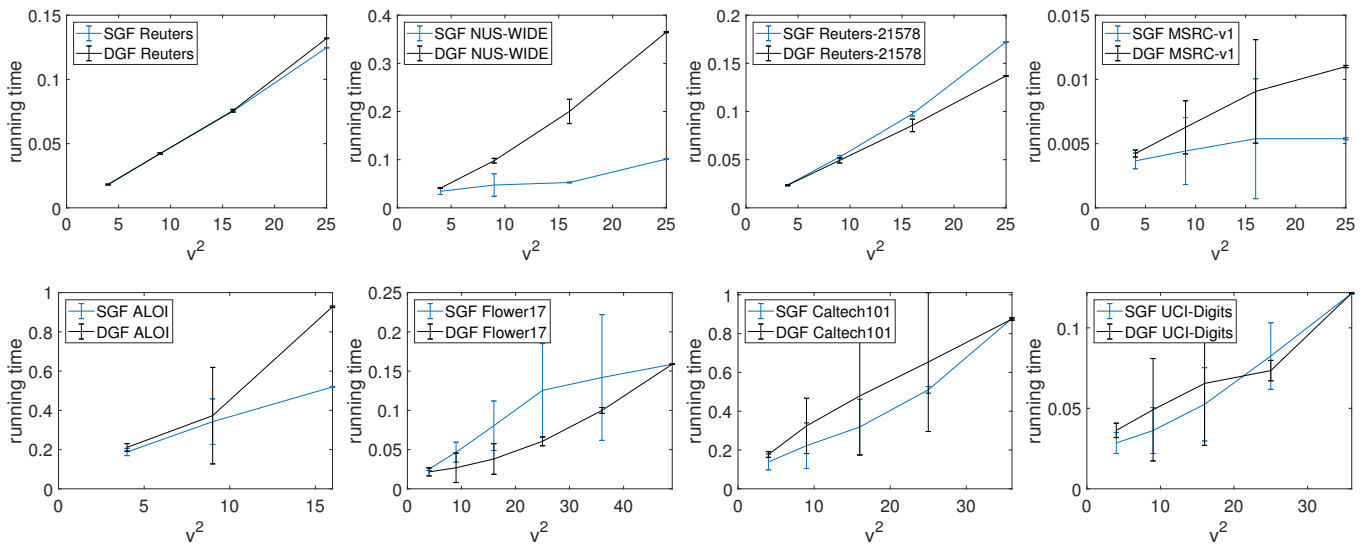

Fig. 4. Running time (in seconds) of the proposed algorithms DGF and SGF against the square of number of views ($v^2$) on the graphs for the 8 datasets with at least 4 views (because at least 4 views are required to draw 3 points on the plot (corresponding to $v = 2, 3, 4$) to see the trend of running time against $v^2$). For each point on the plots (corresponding to a specific number of views $v$), we fix the number of nearest neighbors to 10 and use all possible combinations of the $v$ views from the original dataset to perform the experiments, and we plot the mean and standard deviation of the results. For example, for a total number of 4 views, we choose *all* 2-view combinations from the 4 views to plot the result for $v = 2$.



\end{document}